\documentclass{article}
\usepackage{comment}
\usepackage[export]{adjustbox}
\usepackage[english]{babel}
\usepackage[utf8]{inputenc}
\usepackage{johd}
\usepackage{rotating}
\usepackage{capt-of}
\usepackage{amsmath}
\usepackage{mathtools}
\usepackage{amssymb,amsmath,amsthm}

\usepackage{amsmath}
\usepackage{mathtools}
\DeclarePairedDelimiter\abs{\lvert}{\rvert}%
\usepackage{adjustbox}
\usepackage{array}
\usepackage{booktabs}
\usepackage{dsfont}
\usepackage{multirow}
\usepackage{hhline}
\usepackage[utf8]{inputenc} 
\usepackage[T1]{fontenc}    
\usepackage{hyperref}
\usepackage{url}            
\usepackage{booktabs}       
\usepackage{amsfonts}       
\usepackage{nicefrac}       
\usepackage{microtype}      
\usepackage{amsthm}
\theoremstyle{definition}
\newtheorem{definition}{Definition}[section]

\newtheorem{prop}{Proposition}
\usepackage{float}
\usepackage[caption = false]{subfig}
\usepackage{graphicx}
\DeclareMathOperator*{\argmin}{arg\,min}
\DeclareMathOperator*{\argmax}{arg\,max}

\usepackage[toc,page]{appendix}
\usepackage{algorithm,algorithmic}
\usepackage{xcolor}
\usepackage{amsthm,mathtools}

\newcommand{\COMMENT}[1]{\textcolor{red}{#1}}
\newcommand{\indep}{\perp \!\!\! \perp}

\title{A Fair Pricing Model via Adversarial Learning}

\author{Vincent Grari$^{a,c}$ $^{*}$, Arthur Charpentier$^{b}$, 
Marcin Detyniecki$^{c}$ \\
        \small $^{a}$ Sorbonne Université, ISIR/CNRS, Paris, France \\
        \small $^{b}$ Université du Québec à Montréal (UQAM), Montréal (Québec), Canada \\
        \small $^{c}$ AXA Group, Paris, France \\\\
        \small $^{*}$Corresponding author\footnote{A link to our python code is available here: \url{https://github.com/fairml-research/unbiased_pricingmodel}.}: Vincent Grari; \tt{vincent.grari@isir.upmc.fr}
}

\date{} 

\begin{document}


\maketitle
\begin{abstract} 
\noindent 

At the core of insurance business lies classification between risky and non-risky insureds, actuarial fairness meaning that risky insureds should contribute more and pay a higher premium than non-risky or less-risky ones. Actuaries, therefore, use econometric or machine learning techniques to classify, but the distinction between a fair actuarial classification and `discrimination' is subtle. For this reason, there is a growing interest about fairness and discrimination in the actuarial community \cite{lindholm2022discrimination}.
Presumably, non-sensitive characteristics can serve as substitutes or proxies for protected attributes. For example, the color and model of a car, combined with the driver’s occupation, may lead to an undesirable gender bias in the prediction of car insurance prices. Fairness in insurance pricing is a relatively new and much-requested topic, especially in light of new laws and regulations and past issues encountered in practice \citep{gao2018feature,frees2021discriminating, embrechts2022recent}. 
Consequently, companies/regulators are looking for new methodologies to ensure a sufficient level of fairness while maintaining an adequate accuracy of predictive models. This paper discusses the importance of adapting the traditional fairness algorithms to specific real-life applications and, in particular, to insurance pricing. We claim that mitigating undesired biases with a generic fair algorithm can be counterproductive insurance.  
We will show that traditional Fair-ML as adversarial methods are not currently adequate for insurance pricing. Therefore, for these purposes, we have developed a more suitable and effective framework to satisfy a fairness objective while maintaining a sufficient level of predictor accuracy. 
Inspired by the recent approaches \cite{blier2021geographic,wuthrich2021statistical} that have shown the value of autoencoders in pricing, we will show that (2) it can be generalized to multiple pricing factors (geographic, car type), (3) it is more adapted 
for a fairness context (since it allows to debias the set of pricing components): We extend this main idea to a general framework in which a single whole pricing model is trained by generating the geographic and car pricing components needed to predict the pure premium while mitigating the unwanted bias according to the desired metric.

\end{abstract}


\noindent\\


\section{Introduction}
\subsection{General Context}
Insurance is usually described as the contribution of many to the misfortune of the few by pooling risks together. A fair contribution that should be asked to policyholders, who purchased an insurance policy, is its expected loss over the coverage period of the contract (usually one year), the so-called pure premium. Insurance pricing relies essentially on the law of large numbers, but since risks have to be homogeneous, it is important to classify the risks properly, as explained for instance in \cite{thomas2012non}. For this risk classification, insurers try to split policyholders into different risk groups, and risks are pooled within each group. Those groups are supposed to be as homogeneous as possible. They are usually based on observable factors, such as in motor insurance, the age of the (main) driver, the power of the car, some information about the spatial location, possibly the value of the car, and maybe the gender of the driver. This classification is never perfect, but it should be accepted by policyholders and as valid as possible 
to ask for competitive premiums. 
Heterogeneity within groups means that policyholders cross-subsidy and that could yield adverse selection, where lower risks might be attracted by a competitor, able to capture that residual heterogeneity by offering some more personalized premiums (possibly cheaper for some of them), while the more risky will remain in the portfolio, and cross-subsidy will not work anymore. Therefore, the goal of risk classification is to ask for a risk-based fair actuarial premium and to avoid unnecessary cross-subsidy. As explained in \cite{Paefgen}, ``{\em In order to differentiate the risk of insurance policies, actuaries use a set of rate factors to separate policies into groups (i.e., tariff classes). The construction of tariff classes is ultimately a clustering task. Each tariff class corresponds to a certain combination of rate factor categories or intervals in the case of continuous rate factors. For each tariff class, actuaries analyze historical claims data to arrive at a reliable estimate of the corresponding pure premium, that is, the minimum required payment per policy to cover the expected losses from its class.}''

Using observable 
features ${x}=(x_1,\cdots,x_d)$ to categorize can be explained by its simplicity and reliability. The number of kilometers driven per year is a very interesting information about the risk exposure, unfortunately, 
such information cannot be observed when signing the contract. 
On the other hand, the power of the car, the address, the age, and the gender of the driver are easy to obtain and can be used as {\em predictive} variables simply because they 
might be 
correlated with the associated risk, denoted $y$ (either the claims frequency, individual cost, or total yearly loss). In life insurance, the two most important variables to predict the risk of dying (say within the next year) are respectively the age and the gender of the insured, see for instance \cite{McCarthyTurner}. If we exclude infant 
and child 
mortality, the probability to die within the year is, almost in all countries, increasing with the age 
and there are physiological explanations. Women also tend to live longer than men, at least over the past 100 years. And there are strong biological explanations, with influence blood pressure, and many factors known for causing death. And besides those biological explanations, some behavioral patterns might also be used to explain this gender gap. 


Somehow, having simply a correlation between a variable and the risk might not be sufficient. Actuary, and policyholders, wish to have an explanation about the possible association, not to say that they wish to establish a causal relationship between the classification factor and the risk, as mentioned in \cite{Thiery2006fairness}. Or to be more specific, correlations (or statistical associations) ``{\em should only be seen as a somewhat necessary, but not a sufficient criterion for permissible use of classification}'', as stated in \cite{Wortham}. An important assumption is related to prevention, and control: if I do not want to pay a high premium for my car insurance, I can purchase a car with less power, but I can change neither my age nor my gender. Regarding age, \cite{macnicol2006age} claimed that it is less seen as discrimination since age is not a club into which we are born (as the race or the gender could be).
The use of gender as a classification variable is on the other hand seen as strongly discriminatory and is now forbidden in Europe. 
A major problem for actuaries is that sensitive variable can actually be highly correlated  with many other features. For instance, the color and the model of a car combined with the driver’s occupation can lead to unwanted gender bias in the prediction of car insurance prices. 
Such a situation will be called ``{\em proxy discrimination}'' in \cite{prince2019proxy}. 
For this reason, next to optimizing the performance of a machine learning model, the new challenge for data scientists/actuaries is to determine whether the model output predictions are discriminatory, and how they can mitigate such unwanted bias as much as possible.

\subsection{Motivation}
As ignoring sensitive attributes as input of predictive machines is not sufficient, since complex correlations in the data may provide unexpected links to sensitive information, many bias mitigation strategies for machine learning have been proposed in recent years. However, as we will show most of them are not well adapted for insurance pricing model purposes. 
The mitigation of unwanted biases in the traditional two-step pricing model shows its limitations; some essential information in these components can be not considered in the model while improving for fairness even if it is not biased. For this reason, we have developed a novel pricing model approach where a single whole pricing model is trained to predict the pure premium while mitigating the unwanted biases on all the essential components. For this purpose, we present our approach 
on private car insurance but it can be generalized for commercial, health, and household insurance products. In this context of big data, where insurance companies collect more and more data, aggregating those features by embedding techniques seems to be a judicious choice to preserve explainability, computational traceability, and also fairness.

\subsection{Notations}

Throughout this document, we consider a supervised machine learning algorithm $h_{w_h}$ with parameters $w_h$ for regression or classification problems, 
uses as a pricing model. 
We consider here a variable $y$ that we want to predict, that is either quantitative or categorical and a collection of possible features that were collected. Among the features, $s$ will denote sensitive attributes, that cannot be used to predict, but, 
is a variable that we must observe to ensure fairness of the model, regarding those attributed. The training data consists of $n$ examples ${(x_{i},s_{i},y_{i})}_{i=1}^{n}$, where $x_{i} \in \mathbb{R}^{d}$ is the feature vector with $d$ predictors of the $i$-th policyholder
, $s_i \in \Omega_{S}$ the value of its sensitive attribute 
and $y_{i} \in \Omega_{Y}$ its label to be predicted. According to the setting, the domain $\Omega_{S}$ of the sensitive attribute $S$ can be either a discrete  or a continuous set.


\subsection{Agenda}

In this article, we will consider some fair pricing models in the sense that we do not want to simply remove the sensitive variable $s$ but we want to provide an algorithm that ensures that the pricing will be completely free of any possible discrimination related to $s$ according to a desired specific definition. In Section \ref{sec:fair}, we will discuss a bit more fairness and discrimination with the focus on how to measure fairness for different tasks of a predictive model. Then, in section \ref{sec:notations:method} we focus on how do we correct 
to guarantee fairness in particular via adversarial learning. In section~\ref{chap7:AcruarialPricing}, we briefly describe the actuarial pricing literature with a traditional model with respect to the application of mitigation biases. 
In section~\ref{sec:method_auto_enc}, we present our contribution with a general extension model that can be more adapted for fairness context.
In section~\ref{sec:toyscenario}, we illustrate the fundamental functionality of our proposal
with a simple toy scenario.
Finally, the sections~\ref{sec:res_bin},\ref{sec:res_freq},\ref{sec:res_cont} present the experimental results for the binary, frequency and continuous cases respectively.

. 

\section{Fairness and Discrimination}\label{sec:fair}

\subsection{Discrimination and Unfairness}

There are various notions of fairness, that can be related to discrimination. For instance, direct discrimination happens when a person is treated less favorably than another person in a comparable situation, in the sense that the two persons are otherwise similar except on a sensitive attribute, such as gender, race, etc. This is also called systematic discrimination or disparate treatment. In contrast, indirect discrimination happens when an ``{\em apparently neutral practice put persons of a protected ground at a particular disadvantage compared with other persons}'', as explained in \cite{zliobaite2015survey}. Such discrimination is also known as structural discrimination or disparate outcome.

In many articles about discrimination, 
the sensitive attribute is the gender or the sex 
of a person (here, the policyholder). 
In 1978, the Supreme Court in the U.S. stated that ``{\em the differential was discriminatory in its "treatment of a person in a manner which, but for that person's sex, would be different." The statute, which focuses on fairness to individuals, rather than fairness to classes, precludes treating individuals as simply components of a group such as the sexual class here. Even though it is true that women as a class outlive men, that generalization cannot justify disqualifying an individual to whom it does not apply}''. Following that decision, theoretical discussions about fairness of gender-based pricing started. In Europe, On 13 December 2004, the so-called {\em gender directive} was introduced (Council Directive 2004/113/EC), even if it took almost ten years to provide legal guidelines on the application of the directive to insurance activities. The goal was to enforce the principle of equal treatment between men and women in the access to and supply of goods and services, including insurance. As a direct consequence, it prohibited the use of gender as a rating variable in insurance pricing. As discussed in \cite{Schmeiser2014Unisex}, gender equality in the European Union (EU) was supposed to be ensured from 21 December 2012.

\subsection{Fairness Objective and Discrimination}
\label{fairness_definitions}
In order to achieve fairness, it is essential to establish a clear understanding of its formal definition. In the following, we outline the most popular definitions used in recent research. First, there is information sanitization which limits the data that is used for training the classifier. Then, there is individual fairness, which binds at the individual level and suggests that fairness means that similar individuals should be treated similarly. Finally, there is statistical or group fairness. This kind of fairness partitions the world into groups defined by one or several sensitive attributes. It requires that a specific relevant statistic about the classifier is equal across those groups. In the following, we focus on this family of fairness measures and explain the two most popular definitions.


\paragraph{1) Demographic Parity}

The most common objective in fair machine learning is \emph{Demographic parity} \cite{Dwork2011}. Based on this definition, a model is considered fair if the output prediction $\widehat{Y}$ from features $X$ is independent of the sensitive attribute $S$: 
$\widehat{Y}\indep S$.





\begin{definition}
A machine learning algorithm achieves \emph{Demographic Parity} if the associated prediction $\widehat{Y}$
is independent of the sensitive attribute $S$~\footnotemark[1]: 
\begin{eqnarray*}
\mathbb{P}(\widehat{Y} \leq y |S=s)=\mathbb{P}(\widehat{Y} \leq y ),~\forall s.
\label{def:demographic_parity}
\end{eqnarray*}

\end{definition}
\footnotetext[1]{For the binary case, it is equivalent to $\mathbb{P}(\widehat{Y} = y |S)=\mathbb{P}(\widehat{Y} = y )$}

\paragraph{2) Equalized Odds}
\label{sec:equalized_odds}
The second most common objective in fair machine learning is \emph{Equalized Odds}. Based on this definition, a model is considered fair if the output prediction $\widehat{Y}$ from features $X$ is independent of the sensitive attribute $S$ given the outcome true value Y: $\widehat{Y}\indep S \lvert Y$.

\begin{definition}
A machine learning algorithm achieves Equalized Odds if the associated prediction $\widehat{Y}$
is conditionally independent of the sensitive attribute $S$ given $Y$: 
\begin{eqnarray*}
\mathbb{P}(\widehat{Y} \leq y|S=s,Y=\gamma)=\mathbb{P}(\widehat{Y} \leq y|Y=\gamma),~\forall s,\gamma.
\end{eqnarray*}
\end{definition}

To illustrate Equalized Odds, let's imagine a car insurance pricing scenario where young people have higher claims than older people. A classic pricing model would charge young people a higher premium. However, in the case of demographic parity, the average price would be the same across  all ages. This means that older people would generally pay more than their real cost, and younger people less. In contrast, for Equalized Odds the dependence between the predictions and the real claim cost is preserved, independently of the sensitive variable age.



\subsection{Measuring Fairness in a Binary Setting}
\label{sec:Binary_Setting}
In this context, we consider a binary scenario where the targeted sensitive attribute and the actual value of the outcome are both binaries ($S,Y \in \{0,1\})$. 
\paragraph{1) Demographic Parity}
The use of \emph{Demographic Parity} was originally introduced in this context of binary scenarios ~\cite{Dwork2011}, 
where the underlying idea is that each demographic group has the same chance for a positive outcome.

\begin{definition}
A classifier is considered fair according to the demographic parity principle if $$\mathbb{P}(\widehat{Y}=1|S=0)=\mathbb{P}(\widehat{Y}=1|S=1).$$
\end{definition}



There are multiple ways to assess this objective.
The $p$-rule assessment ensures the ratio of the positive rate for the unprivileged group is no less than  a fixed threshold $\displaystyle{p}$.
The classifier is considered totally fair when this ratio satisfies a 100\%-rule. Conversely, a 0\%-rule indicates a completely unfair model.
\begin{equation}
\emph{p-rule: }
\min\left\lbrace\frac{\mathbb{P}(\widehat{Y}=1|S=1)}{\mathbb{P}(\widehat{Y}=1|S=0)},\frac{\mathbb{P}(\widehat{Y}=1|S=0)}{\mathbb{P}(\widehat{Y}=1|S=1)}\right\rbrace.
\end{equation}
The second metric available for Demographic Parity is the disparate impact ($DI$) assessment~\citep{Feldman2014}. It considers the absolute difference of outcome distributions for subpopulations with different sensitive attribute values. The smaller the difference, the fairer the model.
\begin{equation}
\emph{DI: }
|\mathbb{P}(\widehat{Y}=1|S=1)-\mathbb{P}(\widehat{Y}=1|S=0)|.
\end{equation}

Note that potential differences between demographic groups are not taken into account in this notion. Indeed, in this binary context, only the \emph{weak independence} is required: $\mathbb{E}(\widehat{Y}|S=s)=\mathbb{E}(\widehat{Y}),~\forall s$, and demographic disparity in terms of probability can still be present.
 \\

\paragraph{2) Equalized Odds} 
Equalized Odds can be measured with the disparate mistreatment ($DM$)~\citep{Zafar2017mechanisms}. It computes the absolute difference between the false positive rate (FPR) and the false-negative rate (FNR) for both demographics.
{\small
\begin{eqnarray}
D_{FPR}:
|\mathbb{P}(\widehat{Y}=1|Y=0,S=1)-\mathbb{P}(\widehat{Y}=1|Y=0,S=0)|\\
D_{FNR}:
|\mathbb{P}(\widehat{Y}=0|Y=1,S=1)-\mathbb{P}(\widehat{Y}=0|Y=1,S=0)|
\end{eqnarray}
}
The closer the values of $D_{FPR}$ and $D_{FNR}$ to 0, the lower the degree of disparate mistreatment of the classifier. Therefore the classifier is considered fair if across both demographics $S=0$ and $S=1$, for the outcome $Y=1$ the predictor $\widehat{Y}$ has equal \textit{true} positive rates, and for $Y=0$ the predictor $\widehat{Y}$ has equal \textit{false}-positive rates~\citep{hardt2016equality}. This constraint enforces that accuracy is equally high in all demographics since the rate of positive and negative classification is equal across the groups. The notion of fairness here is that chances of being correctly (or incorrectly) classified as positive should be the same across groups.

\subsection{Measuring Fairness in Continuous Statistical Dependence}
\label{sec:CONTINUOUS}
In this context, we consider a continuous scenario where the targeted sensitive attribute and the actual value of the outcome are both continuous ($S,Y \in \mathbb{R})$ 
In order to assess these fairness definitions in the continuous case, it is essential to look at the concepts and measures of statistical dependence. 
Simple ways of measuring dependence are Pearson's correlation, Kendall's tau, or Spearman's rank correlation. Those types of measures have already been used in fairness, with the example of mitigating the conditional covariance for categorical variables~\cite{Zafar2017mechanisms}. However, the major problem with these measures is that they only capture a limited class of association patterns, like linear or monotonically increasing functions. For example, a random variable with standard normal distribution and its cosine (non-linear) transformation are not correlated in the sense of Pearson. 

To overcome those linearity limitations, \cite{Scarsini1984} first introduced a series of axioms that a {\em measure of concordance} $\delta$ between two random variables should satisfy. Among them, $\delta(U,V)=0$ if and only if $U$ and $V$ are independent, and $\delta(U,V)=1$ if and only if $U$ and $V$ are co-monotonic, meaning that there is some deterministic monotone relationship between $U$ and $V$ (there are $f$ and $g$ such that $V=f(U)$ and $U=g(V)$). \cite{renyi1959measures} suggested to consider the supremum of $r(f(U),g(V))$, where $r$ denotes Pearson's correlation, for all functions $f$ and $g$ such that the correlation can be computed. Such measure was considered earlier in \cite{hirschfeld_1935} and \cite{Gebelein},
$$
HGR(U,V)=\underset{f,g}{\max}\left\lbrace r(f(U),g(V))\right\rbrace,
$$
provided what such a correlation exists. 
An alternative expression is obtained by considering $\mathcal{S}_U=\{f:\mathcal{U}\rightarrow\mathbb{R}:\mathbb{E}[f(U)]=0\text{ and }\mathbb{E}[f(U)^2]=1\}$ and similarly $\mathcal{S}_V$, and then
$$
HGR(U,V)=\underset{f\in\mathcal{S}_U,g\in\mathcal{S}_V}{\max}\left\lbrace \mathbb{E}[f(U)g(V)]\right\rbrace.
$$
And that measure also appeared earlier in \cite{BarrettLampard} and \cite{Lancaster}, while introducing what is called {\em nonlinear canonical analysis}, where we want to write the joint density of the pair $(U,V)$
$$
f_{UV}(u,v) = f_U(u)f_V(v)\left[1+\sum_{i=1}^\infty \alpha_i h_{U,i}(u)h_{V,i}(v) \right]
$$
for some decreasing $\alpha_i$'s in $[0,1]$, for some series of orthonormal centered functions $h_{U,i}$'s and $h_{V,i}$'s, called canonical components. Then one can prove that under mild technical conditions,
$$
HGR(U,V)=\underset{f\in\mathcal{S}_U,g\in\mathcal{S}_V}{\max}\left\lbrace \mathbb{E}[f(U)g(V]\right\rbrace = \mathbb{E}[h_{U,1}(U)h_{V,1}(V)]
$$
For instance, if $(U,V)$ is a Gaussian vector with correlation $r$, then $h_{i}$'s are Hermite's polynomial functions, $h_1(u)=u$, and $HGR(U,V)=|r|$ (the value of the maximal correlation in the Gaussian case was actually established in \cite{Gebelein}). \cite{JensenMayer} extended \cite{renyi1959measures}'s approach by considering some association measure that depend non only on the first canonical correlation, but all of them. Several papers, such as \cite{Buja}, discussed the estimation of maximal correlation, or such as 
kernel based techniques in \cite{DauxoisNkiet}, where $U$ and $V$ are no longer univariate random variables but can take values in more general Hilbert spaces, and more recently \cite{tjostheim2022statistical}.
\footnotetext[2]{$r(U, V)$ := $\frac{Cov(U;V)}{\sigma_{U}\sigma_{V}}$, where $Cov(U;V)$, $\sigma_{U}$ and $\sigma_{V}$ are respectively the covariance between $U$ and $V$, the standard deviation of $U$ and the standard deviation of $V$, respectively.}
The $HGR$ coefficient is equal to 0 if, and only if, the two random variables are independent. If they are strictly dependent the value is 1. 
The spaces for the functions $f$ and $g$ are  infinite-dimensional. This property is the reason why the $HGR$ coefficient proved difficult to compute. 

Note that correlation measure $HGR$ can be extended to a conditional version~\footnotemark[3],
\footnotetext[3]{
Where $\mathcal{S}_{U|Z}=\{f:\mathcal{U}\rightarrow\mathbb{R}:\mathbb{E}[f(U)|Z]=0\text{ and }\mathbb{E}[f(U)^2|Z]=1\}$}
$$
HGR(U,V|Z)=\underset{f\in\mathcal{S}_{U|Z},g\in\mathcal{S}_{V|Z}}{\max}\Big\lbrace \mathbb{E}[f(U)g(V)|Z]\Big\rbrace 
$$
Several approaches rely on Witsenhausen's linear algebra characterization (see ~\cite{witsenhausen1975sequences} to compute the $HGR$ coefficient. For discrete features, this characterization can be combined with Monte-Carlo estimation of probabilities ~\cite{baharlouei2019r}, or with kernel density estimation ($KDE$) \cite{mary2019fairness_full} to compute the $HGR$ coefficient. 
We will refer to this second metric, 
as $HGR_{KDE}$. Note that this latter approach can be extended to the continuous case by discretizing the density support. 
Another way to approximate this coefficient, Randomized Dependence Coefficient ($RDC$) \cite{lopez2013randomized}, is to require that $f$ and $g$ belong to reproducing kernel Hilbert spaces ($RKHS$) and take the largest canonical correlation between two sets of copula random projections. We will make use of this approximated metric 
as $HGR_{RDC}$. Recently a new approach \cite{grari2019fairness} proposes to estimate the $HGR$ by deep neural network. The main idea is to use two inter-connected neural networks to approximate the optimal transformation functions $f$ and $g$. 
The $ HGR_{NN}(U,V)$ estimator is computed by considering the expectation of the products of standardized outputs of both networks ($\hat{f}_{w_{f}}$ and $\hat{g}_{w_{g}}$). The respective parameters $w_{f}$ and $w_{g}$ are updated by gradient ascent on the objective function to maximize: $J(w_f,w_g)=\mathbb{E}[\hat{f}_{w_{f}}(U)\hat{g}_{w_{g}}(V)]$. This estimation has the advantage of being estimated by backpropagation and can handle multi-dimensional random variables~\cite{GrariLD20}.





\paragraph{1) Demographic Parity:}
Compared to the most common discrete binary setting, where the demographic parity can be reduced to ensure \emph{weak independance:} $\mathbb{E}[\widehat{Y}|S=s]=\mathbb{E}[\widehat{Y}],~\forall s$ \cite{pmlr-v80-agarwal18a} and implies $\displaystyle{\sup_{f} r(f(X),S)} = 0$, it does \textbf{not generally imply demographic parity} when $S$ is continuous.
On the other hand, 
the minimization of the $HGR$ dependence ensures \emph{strong Independence} on distribution: $\mathbb{P}[\widehat{Y}\leq y|S]=\mathbb{P}[\widehat{Y}\leq y],~\forall y$ \cite{grari2019fairness} and therefore satisfies the demographic parity objective as below:
\begin{definition}
A machine learning algorithm achieves \emph{Demographic Parity} if the associated prediction $\widehat{Y}$ and the sensitive attribute $S$ satisfies: 
\begin{align}
HGR(\widehat{Y},S)=0.
\label{hgr_DP}
\end{align}
\end{definition}

Compared to the binary case where the fairness measures are fully reliable as the $p$-rule or the $DI$ metrics, they are only estimations in the continuous case as the $HGR_{NN}$ estimation\cite{grari2019fairness} or $HGR_{KDE}$ \cite{mary2019fairness_full}. 
For this reason, we also describe in this paper a metric based on discretization of the sensitive attribute. This $FairQuant$ metric\citep{grari2019fairness} splits the set samples $X$ in $K$ quantiles denoted as $X_k$ (50 in our experiments) with regards to the sensitive attribute, in order to obtain sample groups of the same size. 
It compute the mean absolute difference between the global average and the means computed in each quantile: 

\begin{definition}
We define $K$ as the number of quantiles,  $m_k$ as the mean of the predictions $h_w(X_k)$ in the $k$-th quantile set $X_k$, and $m$ its mean on the full sample $X$. The $FairQuant$ is defined as below:
\begin{align}
FairQuant=\frac{1}{K}\sum_{k=1}^{K} |m_k - m|
\label{Fairquant}
\end{align}
\end{definition}

\paragraph{2) Equalized Odds}
As \emph{Demographic Parity}, the \emph{equalized odds} definition in the binary setting can be reduced to ensure \emph{weak independence} $\mathbb{E}[\widehat{Y}|S,Y]=\mathbb{E}[\widehat{Y}|Y]$ \cite{pmlr-v80-agarwal18a}. It does \textbf{not generally imply equalized odds} when $S$ is continuous. On the other hand, 
the minimization of the $HGR$ dependence ensure the equalized odds objective with \emph{strong independence}:  $\mathbb{P}[\widehat{Y}\leq y|S,Y]=\mathbb{P}[\widehat{Y}\leq y|Y]$ for all $y$. Weak and strong Independence are equivalent when $Y$ is binary.


\begin{prop}
A machine learning algorithm achieves equalized odds if the associated prediction $\widehat{Y}$ and the sensitive attribute $S$ satisfies:
\begin{align}
HGR(\widehat{Y},S\vert Y)=0.
\end{align}
\end{prop}


\subsection{Measuring Fairness in Frequency Statistical Dependence}
In this setting, we assume that the outcome target $Y$ can be represented as a number of events occurring in a fixed interval of time. In this particular frequency setting, we notice a lack of work for assessing the level of fairness. 

\paragraph{1) Demographic Parity}
For the demographic parity objective, the weak independence by expectation is not sufficient in this context to ensure the definition of \ref{def:demographic_parity}. Indeed, by the nature of the continuous output of a number of events, it requires the notion of strong independence on the distribution. We propose in this context to apply the same notion of continuous proposes~\ref{sec:CONTINUOUS} as seen above by assessing the $HGR$ neural network estimator and the Fairquant Def.~\ref{Fairquant} between the prediction and the sensitive attribute.

\paragraph{2) Equalized Odds}
For the equalized odds objective, 
we propose to assess the level of independence on each number of events $Y$. For this purpose, the notion of demographic parity is required for each subset of value $Y$.


%
 $$HGR_{EO}=\frac{1}{\#\Omega_{Y}}\sum_{y \in \Omega_{Y}}HGR(h_{w_h}(X),S) $$

$$FairQuant_{EO}=\frac{1}{\#\Omega_{Y}}\sum_{y \in \Omega_{Y}}\frac{1}{K}\sum_{i=1}^{K} |m_{i,y} - m|$$
with $m_{i,y}$ as the mean of the predictions $h_w(X_k)$ in the $k$-th quantile set $X_k$ conditioned on $Y=y$, and $m$ its mean on the full sample $X$.
For computational reasons, $\Omega_Y$ denotes the set of `valid' entries. In the context of counts, the standard goodness of fit test is Pearson's chi-squared test, unfortunately, the latter is not valid if the expected frequencies are too small (as in \cite{bol1979chi} and there is no general agreement on the minimum expected frequency allowed). In practice, if $y\in\{0,1,2,3,4,5,6\}$ with (say) less than 1\% for counts (strictly) exceeding 2, it is rather common to consider  $\Omega_Y=\{0,1,2^+\}$, where $2^+$ denotes the case where counts exceed (strictly) 1. 

\section{Using Adversarial Learning to Ensure Fairness}\label{sec:notations:method}

Three main families  of fairness approaches exist in the literature. While   
pre-processing \citep{kamiran2012data,bellamy2018ai,calmon2017optimized} and post-processing  \citep{hardt2016equality,chen2019fairness} approaches respectively  act on the input or the output of a classically trained predictor,  
pre-processing \citep{kamiran2012data,bellamy2018ai,calmon2017optimized} and post-processing  \citep{hardt2016equality,chen2019fairness} approaches respectively act on the input or the output of a classically trained predictor,  
in-processing approaches mitigate the undesired bias directly during the
training phase \citep{Zafar2017mechanisms, zhang2018mitigating,wadsworth2018achieving,louppe2017learning}. In this paper, we focus on in-processing fairness and in particular with adversarial learning, which reveals as the most powerful framework for settings where acting on the training process is an option. 
Since this method relies on penalization during training, we will first describe the general literature on this subject. Then, we will show how penalization is achieved in the fairness framework for demographic parity and equalized odds and this for three following prediction tasks: binary task, frequency task, and average costs task.

\subsection{Penalization via Adversarial Learning}

In this section, we will describe briefly certain types of penalization during the training of the predictor model. One of the most known penalization procedure for a classical machine learning model where we want to learn a model $h_{w_h}(X)$ that should be close to $Y$ (measured via a loss function $\mathcal{L}$ ) is as follow:
$$
\underset{w_h}{\text{argmin}}\big\lbrace\mathcal{L}(h_{w_h}(X),Y)+\lambda p(w_h)\big\rbrace
$$
for some penalty function $p$, usually taking into account the complexity of the model, that will help us avoid overfitting.  

There are other kinds of learning penalization named adversarial machine learning, as explained in \cite{Goodfellow2018},
that are machine learning techniques that tend to robustify the predictive mode, by attempts to fool models by supplying deceptive input. In order to improve the robustness of the model, it can be natural to consider an adversarial approach such as:
$$
\underset{w_h}{\text{argmin}}\big\lbrace\max_{m,|m|<\epsilon}\big\lbrace\mathcal{L}(h_{w_h}(X),Y)+\lambda \ell(h_{w_h}(X),h_{w_h}(X+m))\big\rbrace\big\rbrace
$$ 
Where we consider the worst-case impact of a small perturbation of the data. The first term represents the classical loss function $\mathcal{L}$ 
which is minimized in order to improve the accuracy of the predictions. The second term, represented as the penalization term, is computed in order to compare the prediction with an altered version (the highest perturbation $m$ on $X$). A classical example in the context of pictures labeling is the {\em ostrich} example in \cite{szegedy2014intriguing}, where all pictures (yellow bus, dog, pyramid, insect, etc), slightly perturbed with some noisy picture, still, look as before for a human eye, but are all labeled as an ostrich.

Some other types of adversarial learning which are more related to our work rely on two separate machine learning models trained simultaneously.
The most popular 
one is the  Generative Adversarial Networks (GANs) \cite{Goodfellow2014}.
The objective is, given a training set, to generate new data similar to the training set. The two models are defined as the generator $G$ that captures the data distribution $X \in \mathbb{R}^{d}$ from a 
$d_{z}$-dimensional random noise $Z \in \mathbb{R}^{d_z}$
, and a discriminative model $D$ that estimates the probability that a sample came from the real training data $X$ rather than $G$.
$$
\underset{G}{\text{argmin}}\big\lbrace\max_{D}\big\lbrace\mathbb{E}_{X\sim p_{data}}[\log(D(X)]+\mathbb{E}_{Z\sim p_{z}}[\log(1-D(G(Z))]\big\rbrace\big\rbrace
$$
This framework corresponds to a minimax two-player game. 
The first term represents the expectation over the real data $X$ from the training distribution $p_{data}$. This term is only used for the discriminator and allows to assign the larger label to the real data. The second term is expectation over noise. It inputs the noise $Z$ with distribution $p_{z}$ (for e.g. a multivariate normal distribution) to the generator to obtain $G(Z)$. This latter generator's output represents the generated observations and is fed as input to the discriminator for obtaining $D(G(Z))$. The discriminator attempts to minimize $D(G(Z))$ since the smaller label is assigned to the generated data. 
Therefore the generator attempts to fool the discriminator to label the generated data as real.

\subsection{Improving Demographic Parity:}
The fair state-of-the-art algorithms for achieving the demographic parity objective %
are generally constructed 
with a penalization term that can be plugged in the following generic optimization problem as below: 
\begin{eqnarray}
\begin{aligned}
    \argmin_{{w_h}} & \ \Big\lbrace \mathcal{L}(h_{w_h}(X),Y)+
    \lambda p(h_{w_h}(X),S)\Big\rbrace
    \label{genericfunction_DP},
\end{aligned}
\end{eqnarray}
where  $\mathcal{L}$ is the predictor loss function (the mean squared error for regression or log-loss for the binary classification task for example) between the output $h_{w_h}(X) \in \mathbb{R}$ and the corresponding target $Y$, with $h_{w_h}$ the prediction model which can be for example be GLM or a deep neural network with parameters $w_h$, and $p(h_{w_h}(X),S)$ the penalization term which evaluates the correlation loss between two variables.
The aim is thus to find a mapping $h_{w_h}(X)$ which both minimizes the  deviation with the expected target $Y$ and does not imply too much dependency with the sensitive $S$.
The hyperparameter $\lambda$ controls 
for impact of the correlation loss in the optimization. 
The correlation loss $p$ can correspond to a Pearson coefficient or a Mutual Information Neural Estimation (MINE, \cite{belghazi2018mutual}), or a value that reflects the inability to reconstruct the sensitive attribute~\cite{zhang2018mitigating,kim2017auditing} 
or $HGR$ neural estimators that we will discuss below. 


\subsubsection{Adversarial Simple Architecture:} 
\begin{figure*}[h]
  \centering
  \includegraphics[scale=0.95]{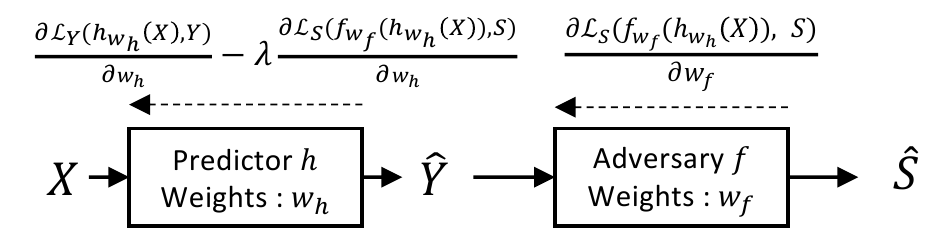}
  \caption{The Adversarial simple architecture} 
  \label{fig:ADV_SIMPLE}
\end{figure*}

Also, some approaches \cite{zhang2018mitigating} assess the level of dependency by considering how it may be able to reconstruct the sensitive attribute $S$ from the output prediction $h_{w_h}(X)$. 
By feeding the output prediction as input to an adversary $f_{w_f}$ that takes the form of a GLM or deep neural network with the objective to predict 
$S$ it allows to measure the level of dependence during the training. 
The goal is to obtain a predictor model $h_{w_h}$ whose outputs do not allow the adversarial function to reconstruct the value of the sensitive attribute.
If this objective is achieved, the data bias in favor of some demographics disappeared from the output prediction. The predictor with weights $w_h$ has fooled the adversary. The optimization problem is as below:

\begin{eqnarray}
\underset{w_h}{\text{argmin}}\Big\lbrace\max_{w_f}\Big\lbrace\mathcal{L_Y}(h_{w_h}(X),Y)-\lambda \mathcal{L_S}(f_{w_f}(h_{w_h}(X)),S)\Big\rbrace\Big\rbrace
\label{eq:zhang}
\end{eqnarray}
where  $\mathcal{L_Y}$ is the predictor loss function 
between the output $h_{w_h}(X) \in \mathbb{R}$ and the corresponding target $Y$ and  $\mathcal{L_S}$ is the adversary loss function 
between the adversary output $f_{w_f}(h_{w_h}(X)) \in \mathbb{R}$ and the corresponding sensitive attribute $S$.
The hyperparameter $\lambda$ controls the impact of the dependence loss in the optimization. 
The prediction $h_{w_h}(X)$ is the input given to the adversarial $f_{w_f}$. 
%
The backpropagation of the adversary with parameters $f_{w_f}$ is performed by multiple steps of gradient ascent. This allows us to optimize a more accurate estimation of the reconstruction of the sensitive, leading to a greatly more stable learning process.  
Figure~\ref{fig:ADV_SIMPLE} gives the  architecture of this adversarial learning algorithm. It depicts the predictor function $h_{w_{h}}$, which outputs the prediction from $X$, the adversarial predictor $f_{w_f}$ which seek at defining the most accurate prediction to $S$ from the predictor function $h_{w_{h}}$. Left arrows represent gradients back-propagation. The learning is done via stochastic gradient, alternating steps of adversarial maximization, and global loss minimization. 
At the end of each iteration, the algorithm updates the parameters of the prediction parameters $h_{w_h}$  by one step of gradient descent. Concerning the  adversarial, the backpropagation of the parameters $w_f$ is carried by multiple steps of gradient ascent. This allows us to optimize a more accurate estimation of the sensitive attribute at each step, leading to a greatly more stable learning process. 

Note that for an infinite $\lambda$, the second term is the only one to be optimized. Maximizing the negative gradient on the parameter $w_f$ allows minimizing the loss function between the adversary's prediction and the sensitive attribute. 
Then, by minimizing this term in a second step via the $w_h$ parameters, it allows removing all the sensitive biases. 
Also, note that, if there exists $(w_h^*,w_{f}^*)$ such that $w_{f}^*=\argmax_{w_{f}}\lbrace \mathbb{P}_{w_h^*,w_{f}}(S|h_{w_h}^*(X))\rbrace$ on the training set, $\mathbb{P}_{w_h^*,w_{f}^*}(S\leq s|h_{w_h}^*(X))=\widehat{\mathbb{P}}(S\leq s)$ and $\mathbb{P}_{w_h^*}(Y\leq y|X)=\widehat{\mathbb{P}}(Y\leq y|X)$, with $\widehat{\mathbb{P}}(S\leq s)$ and $\widehat{\mathbb{P}}(Y\leq y|X)$,~$\forall s,y$ the corresponding distributions on the training set, $(w_h^*,w_f^*)$ is a global optimum of our min-max problem eq. (\ref{eq:zhang}). In that case, we have both a perfect classifier in training and a completely fair model since 
the best possible adversary is not able to predict $S$ more accurately  than the estimated prior distribution. While such a perfect setting
does not always exist in the data, it shows that the model is able to identify a solution when it reaches one. If a perfect solution does not exist in the data, the optimum of the minimax 
problem is a trade-off between prediction accuracy and fairness, controlled by the hyperparameter $\lambda$.



However, 
in a regression task, with the mean square loss for e.g.,
this type of approach 
achieves the global fairness optimum when $\mathbb{E}(S \vert \widehat{Y})=\mathbb{E}(S)$
. This does \textbf{not generally imply demographic parity} when $S$ is continuous. On the other hand, adversarial approaches based on the $HGR_{NN}$ \cite{grari2019fairness} that we describe bellows 
achieve the optimum when $HGR(\widehat{Y},S)=0$, which is equivalent to demographic parity: $\mathbb{P}(\widehat{Y}\leq y\vert S)=\mathbb{P}(\widehat{Y}\leq y)$, $\forall y$. 

\subsubsection{Adversarial $HGR$ Architecture:}
The $HGR$ approach proposed by \cite{grari2019fairness} uses an adversarial network that takes the form of two inter-connected neural networks for approximating the optimal transformations functions $f$ and $g$. 

\begin{eqnarray}
\begin{aligned}
\nonumber
    \argmin_{w_h}\Big\lbrace\max_{{w_{f},w_{g}}}\big\lbrace&\mathcal{\mathcal{L}}(h_{w_h}(X),Y)
    +
    \lambda\mathbb{E}_{(X,S)\sim\mathcal{D}}(\widehat{f}_{w_{f}}(h_{w_h}(X))\widehat{g}_{w_{g}}(S)\Big\rbrace\Big\rbrace
    \label{func:adv_HGR}
\end{aligned}
\end{eqnarray}
where  $\mathcal{L}$ is the predictor loss function 
between the output $h_{w_h}(X) \in \mathbb{R}$ and the corresponding target $Y$.
The hyperparameter $\lambda$ controls the impact of the dependence loss in the optimization. 
The prediction $h_{w_h}(X)$ is the input given to the adversarial $f_{w_f}$ and the sensitive $S$ is given as input to the adversarial $g_{w_g}$. In that case, we only capture for each gradient iteration the estimated $HGR$ between the prediction and the sensitive attribute. The algorithm takes as input a training set from which it samples batches of size $b$ at each iteration. At each iteration, it first standardizes the output scores of networks $f_{w_f}$ and $g_{w_g}$ to ensure $0$ mean and a variance of $1$ on the batch. Then it computes the objective function to maximize the estimated $HGR$ score and the global predictor objective.  Finally, at the end of each iteration, the algorithm updates the parameters of the $HGR$ adversary $w_f$ and $w_g$ by multiple steps of gradient ascent and the regression parameters $w_h$ by one step of gradient descent. This allows us to optimize a more accurate estimation of the $HGR$ at each step, leading to a greatly more stable learning process. 

\begin{figure}[h]
  \centering
  \includegraphics[scale=0.90,valign=t]{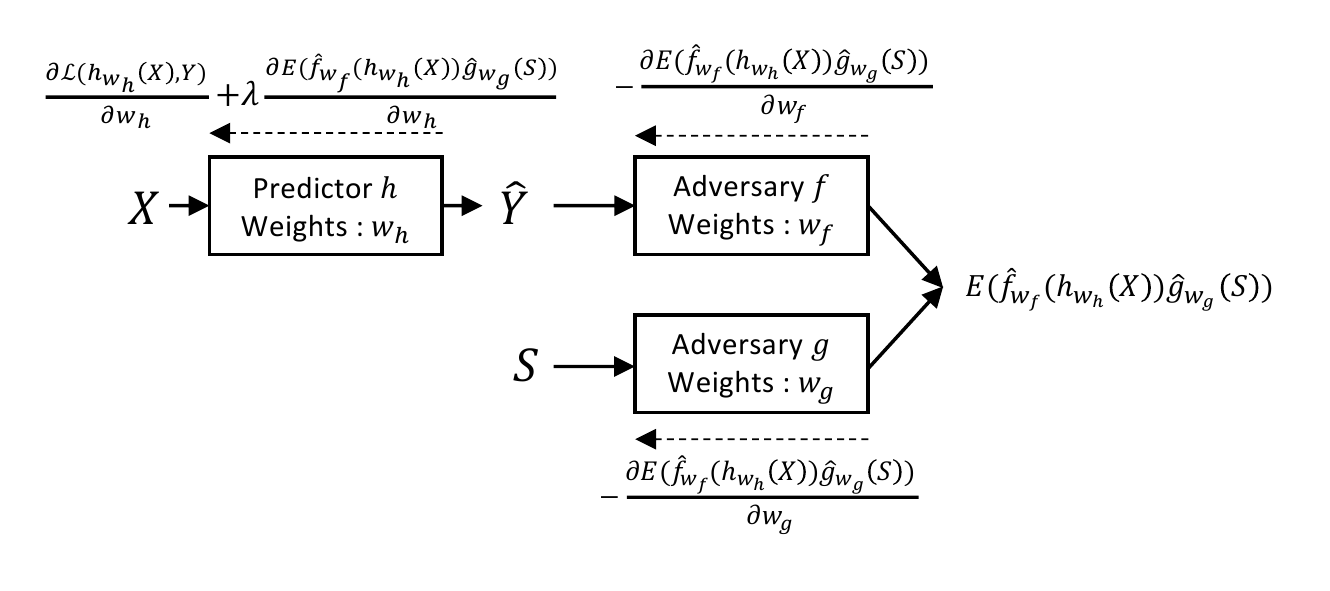}
  \caption{The Fair $HGR$ NN adversarial algorithm for demographic parity.} 
  \label{fig:hgr_fair}
\end{figure}

Figure \ref{fig:hgr_fair} gives the full architecture of the adversarial learning algorithm using the neural $HGR$ estimator for demographic parity. It depicts the prediction function $h_{w_h}$, which outputs $\widehat{Y}$ from $X$, and the two neural networks $f_{w_f}$ and $g_{w_g}$, which seek at defining the more strongly correlated transformations of  $\widehat{Y}$ and $S$. Left arrows represent gradient back-propagation. The training is done via stochastic gradient, alternating steps of adversarial maximization, and global loss minimization. 

\subsection{Improving Equalized Odds:}
The fair in-processing algorithms for achieving the equalized odds objective are generally constructed with a penalization term that can be plugged in the following generic optimization problem as below: 
\begin{eqnarray}
\begin{aligned}
    \argmin_{{w_h}} & \Big\lbrace\ \mathcal{L}(h_{w_h}(X),Y)+
    \lambda p(h_{w_h}(X),S,\textcolor{red}{Y})\Big\rbrace
    \label{genericfunction_imp_EO}
\end{aligned}
\end{eqnarray}
where  $\mathcal{L}$ is the predictor loss function 
between the output $h_{w_h}(X) \in \mathbb{R}$ and the corresponding target $Y$, with $h_{w_h}$ the prediction model with parameters $w_h$, and $p(h_{w_h},Y,S)$ the penalization term which evaluate the correlation loss between the output prediction and the sensitive attribute given the expected outcome $Y$. The aim is thus to find a mapping $h_{w_h}(X)$ which both minimizes the deviation with the expected target $Y$ and does not imply too much dependency with the sensitive $S$ given $Y$. 

\subsubsection{Adversarial Simple Architecture:} 
\begin{figure}[h]
  \centering
  \includegraphics[scale=0.90,valign=t]{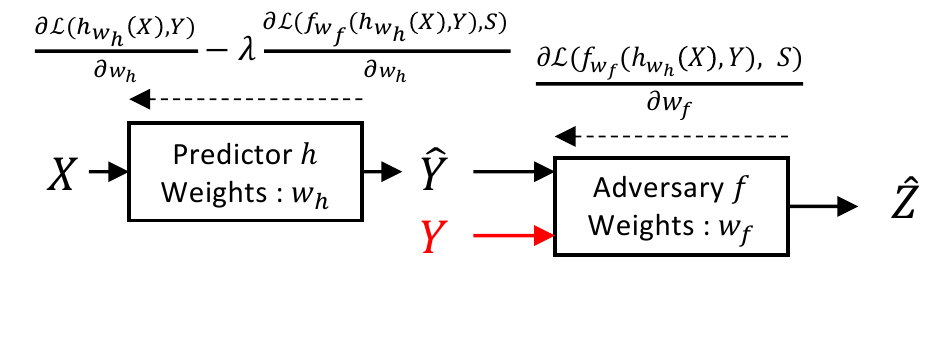}
  \caption{The Fair $HGR_{NN}$ NN adversarial algorithm for equalized odds.} 
  \label{fig:adv_simpleEO}
\end{figure}

Following the  idea of adversarial simple architecture for demographic parity, \cite{zhang2018mitigating} proposes to concatenate the label $Y$
to the output prediction 
to form the input vector of the
adversary ($h_{w_h}(X)$, $Y$), so that the adversary function $f_{w_f}$
could be able to output different conditional probabilities
depending on the label $Y_i$ of individual $i$.




$$
\underset{w_h}{\text{argmin}}\big\lbrace\max_{w_f}\big\lbrace\mathcal{L_Y}(h_{w_h}(X),Y)-\lambda \big\lbrace
\mathcal{L_S}(f_{w_f}(h_{w_h}(X,\textcolor{red}{Y})),S)
\big\rbrace\big\rbrace
$$
Figure \ref{fig:adv_simpleEO} gives the full architecture of this adversarial learning algorithm simple for equalized odds. It depicts the predictor function $h_{w_{h}}$, which outputs the prediction from $X$, the adversarial predictor $f_{w_f}$ which seek at defining the most accurate prediction to $S$ from the predictor function $h_{w_{h}}$ and the targeted variable $Y$. Left arrows represent gradients back-propagation. The learning is done via stochastic gradient, alternating steps of adversarial maximization, and global loss minimization. 

However, as demographic parity, this algorithm can only be considered in binary cases since it does not generally imply equalized odds when $S$ is continuous. 


\subsubsection{$HGR$ Adversarial Architecture:}
Whether the sensitive variable is continuous or binary, the 
approach 
of \cite{grari2022} proposes 
to extend the idea of the $HGR$ adversarial algorithm for binary equalized odds. For the decomposition of disparate mistreatment, 
it proposes to divide the mitigation based on the two different values of $Y$.
Identification and mitigation of the specific non-linear dependence for these two subgroups leads to 
the same false-positive and the same false-negative rates for each demographic.
The optimization is written as follows:
\begin{eqnarray}
\begin{aligned}
\nonumber
    \argmin_{w_h}\Big\lbrace\max_{{w_{f_{0}},w_{g_{0}},w_{f_{1}},w_{g_{1}}}}\Big\lbrace&\mathcal{L}(h_{w_h}(X),Y)
    \\
    &+
    \lambda_{0}\mathbb{E}_{(X,S)\sim\mathcal{D}_{0}}(\widehat{f}_{w_{f_{0}}}(h_{w_h}(X))\widehat{g}_{w_{g_{0}}}(S))\\
    &+
    \lambda_{1}\mathbb{E}_{(X,S)\sim\mathcal{D}_{1}}(\widehat{f}_{w_{f_{1}}}(h_{w_h}(X))\widehat{g}_{w_{g_{1}}}(S))\Big\rbrace\Big\rbrace
    \label{func:HGRARCH}
\end{aligned}
\end{eqnarray}
where $\mathcal{D}_{0}$ corresponds to the distribution of pair $(X,S)$ conditional on ${Y=0}$ and $\mathcal{D}_{1}$ to the distribution conditional on ${Y=1}$. 
The hyperparameters $\lambda_{0}$ and $\lambda_{1}$ control the impact of the dependence loss for the false positive and the false negative objective respectively. 

The first penalization (controlled by $\lambda_0$) enforces the independence between the output prediction $h_{w_h}(X)=p_{w_h}(Y=1|X)$ and the sensitive $S$ only for the cases where $Y=0$. It enforces naturally the mitigation of the difference of false positive rate between demographics since at optimum for $w_h^*$ with  $(X,S)\sim\mathcal{D}_{0}$, $HGR(h_{w_h^{*}}(X),S)=0$  and implies theoretically: $h_{w_h^{*}}(X,S) \indep S| Y=0$. 
We propose to extend and generalize this idea to binary and frequency tasks where outcome target $Y$ can be represented as a number of events occurring in a fixed interval of time. 
We propose to divide the mitigation based on all the different values of $Y$. 
The optimization is as follows:
\begin{eqnarray}
\begin{aligned}
\nonumber
    \argmin_{w_h}\Big\lbrace\max_{{w_{f_{0}},w_{g_{0}},...,w_{f_{K}},w_{g_{K}}}}\Big\lbrace&\mathcal{L}(h_{w_h}(X),Y)
    +
    \frac{1}{K+1}\sum_{y \in \Omega_{Y}}
    \lambda_{y}\mathbb{E}_{(X,S)\sim\mathcal{D}_{y}}(\widehat{f}_{w_{f_{y}}}(h_{w_h}(X))\widehat{g}_{w_{g_{y}}}(S))\Big\rbrace\Big\rbrace
    \label{genericfunctionEO}
\end{aligned}
\end{eqnarray}
where $\mathcal{D}_{y}$ corresponds to the distribution of pair $(X,S)$ conditional on ${Y=y}$ and $K=\#\Omega_{Y}-1$. 
The hyperparameters $\lambda_{y}$ control the impact of the dependence loss for the different number event objective. 
The penalization enforces the independence between the output prediction 
and the sensitive $S$ only for the cases where $Y=y$. It enforces naturally the mitigation of equalized odds since it enforces the mitigation of biases for demographic parity for each number of events. 

\section{Traditional Actuarial Pricing} 
\label{chap7:AcruarialPricing}

A classical technique in actuarial science is based on Generalized Linear Models, GLM, since they satisfy a ``{\em balance property}'', as called in  \citep{wuthrich2021statistical} for instance, stating that the sum of $y_i$'s (on the training dataset) should equal the sum of $\widehat{y}_i$'s. The economic interpretation is that the first sum of the sum of losses, while the second one is the sum of premiums. Hence, using GLMs, we ensure that, on average, the insurer will be able to repay policyholders claiming a loss. 
Unfortunately, GLMs experience difficulties when categorical rating factors have a large number of levels, not only because of the computational cost of dealing with high-dimensional design matrices, but also because of the implied statistical uncertainty, both in parameter estimation and prediction (even if regularization techniques can be used, as in \citep{frees2015rating}).

\subsection{Description of the Method}\label{sec:method}
In the insurance pricing context, the GLM preditor model is not fitted only once as in the traditional machine learning standard. Instead, as mentioned in \citep{taylor_1989}, \citep{boskov_verrall_1994}, or more recently \citep{Tufvesson} a standard approach in insurance is to consider a two-step procedure, where the first step initially considers the generation of geographic and automobile risk components. Then the second step is based on the predictive task. The insurer often has recourse to a large number of external variables, which can usually exceed a hundred~\citep{BOUCHER,Beraud}. These can be geographical (e.g., the total number of thefts or crime rate in the area of residence) 
and car-specific (e.g., airbag, emergency braking, etc.).
\citep{shi2021nonlife} recalls that it can be difficult to use, in classical actuarial pricing models, categorical variables with a large number of categories, such as a ZIP code (spatial information) or type/variant/version/model/make of cars (vehicle information).  In order to keep models under control (i.e., computational traceability and explainability), actuaries traditionally prefer to aggregate all this information into single variables. 
The generation of these components is performed by a first model based on the prediction of the target $Y$ (i.e., frequency and severity), excluding all external information (spatial and unstructured effects). 
The idea is that the observed residuals of this model correspond to the missing information that it was unable to capture (i.e., geographical, car information).
In this way, a second model is trained on these observed residues from these external data. This method allows assigning external risk prediction to new policyholders based on their external information. For example, suppose a policyholder comes from a residential area that the insurer has never had in its historical data. 
In this case, the insurer will still be able to generate a geographic risk level using only external information from the area.
Note that actuaries, notably by K-Nearest Neighbors (KNN) methods, reprocess this residual to smooth the predictions 
(e.g, KNN on the spatial neighborhood for the geographic component). As discussed in \citep{blier2021geographic}, early geographic models in actuarial science were models that smoothed the residuals of a regression model (also called ``{\em correction models}'') where geographic effects are captured after the main regression model, in a smoothing model, as in \citep{taylor_1989}. For example \citep{fahrmeir2003generalized}, and more recently \citep{wang2017geographical}, suggested using spatial interpolation, inspired by kriging techniques, to capture spatial heterogeneity. In addition, a quantile approach is 
applied to these components in order to divide these components into several risk levels. Finally, a final model is fitted to the objective task based on these generated risk components and policyholder information. 

In this particular context, the training data 
$x_{i} \in \mathbb{R}^{d}$ 
is decomposed into to subsample $x_{p_{i}} \in \mathbb{R}^{d_{p}}$ corresponding to the policyholder's information, $x_{g_{i}} \in \mathbb{R}^{d_{g}}$ corresponding to the geographical information with $d_{g}$ predictors and $x_{c_{i}} \in \mathbb{R}^{d_{c}}$ corresponding to the car information with $d_{c}$ predictors.

\begin{figure}[h]
\centering
\includegraphics[width=0.75\textwidth]{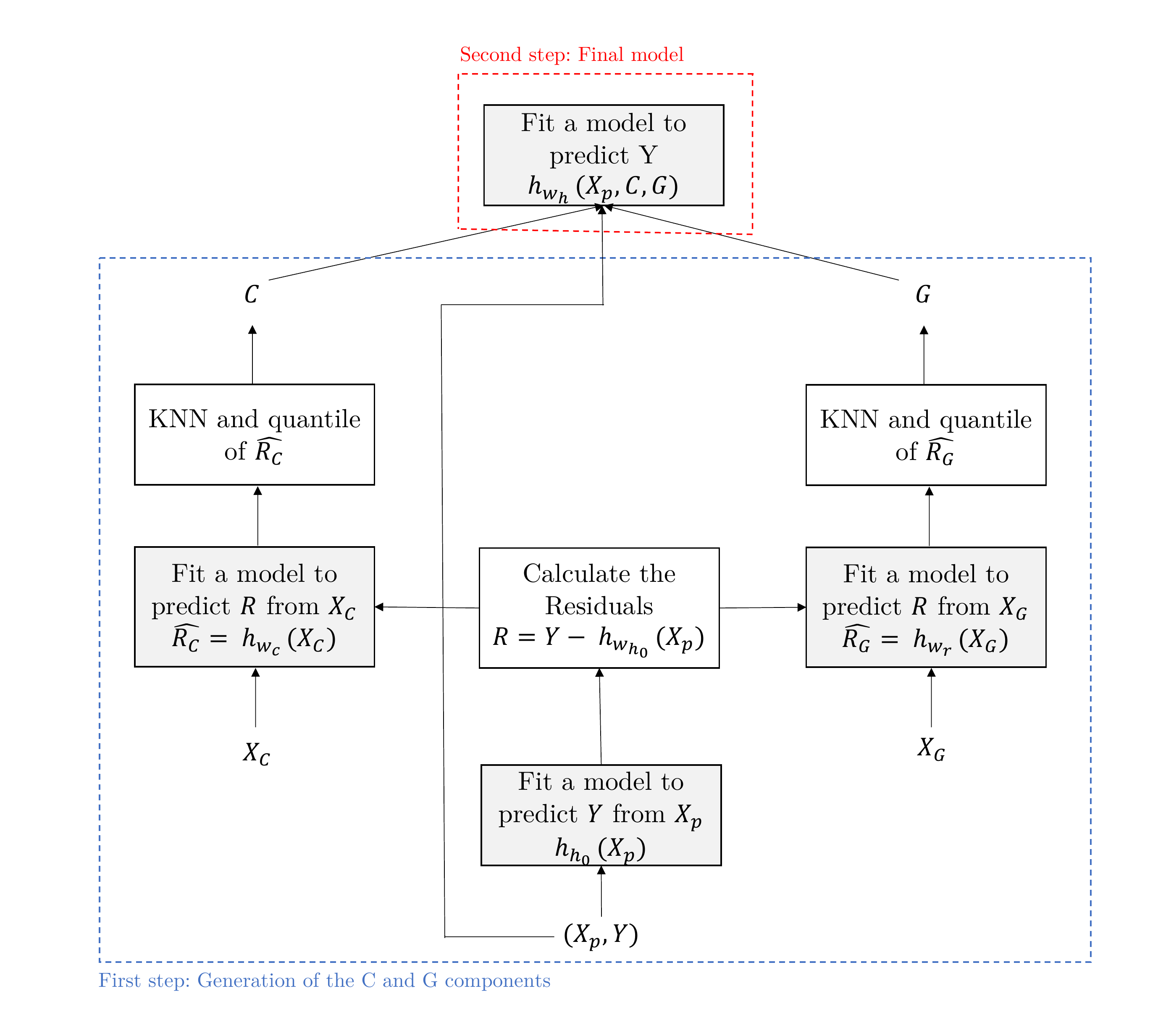}
\caption{\label{fig:two_step_trad} Two-Step Traditional Pricing model}
\label{Fig:Trad_PricingModel2}
\end{figure}


In Figure~\ref{Fig:Trad_PricingModel2}, we have represented the general process of the two-step traditional pricing. For the first step, a model $h_{w_{0}}$ with parameters $w_0$ is fitted to predict the target task $Y$ from the policyholder's information $X_p$ (i.e., without including the external information $X_{g}$ and $X_c$). The residual of this model $R$ is calculated (can be calculated on the relative difference in practice~\citep{SAID}).  
In second, some models $h_{w_g}$ and $h_{w_c}$ with parameters $w_g$ and $w_c$ are trained to predict the observed residuals $R$ from the corresponding set of variables $X_g$ and $X_c$ respectively. Some prereatreatments are realized on the output predictions of these models as KNN methods and/or quantiles. Finally, a last model $h_{w_h}$ with parameter $w_h$ is trained to predict the target task $Y$ from the policyholder's information $X_p$ and the two generated features $C$ and $G$. 

\subsection{Fairness on the Traditional Two-Step Pricing Model}
\label{TP_FAIR}


A classic methodology for enforcing fairness should be to act directly on the final predictor model $h_{w_h}$. An additional adversarial 
could mitigate the sensitive bias during the training process. By following the different optimizations seen in subsection~\ref{sec:notations:method}, the demographic parity objective can be improved on the traditional Two-Step Pricing Model as follow:

\paragraph{TP\_Corr} Traditional Pricing with a pearson coefficient penalization
\begin{eqnarray}
\begin{aligned}
    \argmin_{{w_h}} & \ \Big\lbrace \mathcal{L_Y}(h_{w_h}(X_p,C,G),Y)+
    \lambda \abs{r(h_{w_h}(X_p,C,G),S)}\Big\rbrace
    \label{genericfunction_TP_Cor},
\end{aligned}
\end{eqnarray}

\paragraph{TP\_Simple} Traditional Pricing with an adversarial simple penalization
\begin{eqnarray}
\begin{aligned}
    \underset{w_h}{\text{argmin}}\Big\lbrace\max_{w_f}\Big\lbrace\mathcal{L_Y}(h_{w_h}(X_p,C,G),Y)-\lambda \mathcal{L_S}(f_{w_f}(h_{w_h}(X_p,C,G)),S)\Big\rbrace\Big\rbrace
    \label{genericfunction_TP_Simple},
\end{aligned}
\end{eqnarray}

\paragraph{TP\_HGR} Traditional Pricing with an $HGR$ penalization
\begin{eqnarray}
\begin{aligned}
\nonumber
    \argmin_{w_h}\Big\lbrace\max_{{w_{f},w_{g}}}\big\lbrace&\mathcal{\mathcal{L_Y}}(h_{w_h}(X_p,C,G),Y)
    +
    \lambda\mathbb{E}_{(X_p,C,G,S)\sim\mathcal{D}}(\widehat{f}_{w_{f}}(h_{w_h}(X_p,C,G))\widehat{g}_{w_{g}}(S)\Big\rbrace\Big\rbrace
    \label{func:TP_HGR}
\end{aligned}
\end{eqnarray}

However, we argue that this is not an appropriate process. Note that the car risk $C$ and the geographic risk $G$ have already been trained in the first step and are not re-trained in this optimization. Therefore, the unwanted bias is only mitigated on the predictor $h_{w_h}$. Consequently, if the predictor $h_{w_h}$ is a GLM and if $G$ and $C$ are strongly dependent on the sensitive attribute $S$, the parameters $w_h$ corresponding to $G$ and $C$ may tend towards $0$ and thus nullify the effect of these factors. For improving fairness, the risk is to miss the relevant information about $G$ and $C$ for predicting $Y$. 

This motivates us for a more robust model that breaks away from this drawback. So the goal here will be to define a new 
actuarial framework that is able to generate fair geographical and car variables, 
that are strongly correlated with the risk $Y$, but fair with respect to $S$. 









\label{two_stage}

\section{Pricing with an Autoencoder Structure}
\label{sec:method_auto_enc}

We propose in this section to generalize an actuarial model that would be better fitted for fairness. The main idea would be to train the geographical and car risk at the same time of the model.
As discussed in \citep{wuthrich2021statistical}, a classical starting point is some initial feature engineering step, where some embedding of spatial components, and information relative to the vehicle, are considered, using principal component analysis (PCA). However, in this approach, the generated components are not trained with the predictor and cannot be specifically targeted to the objective task. Instead of considering several separate models stacked together as described above, it is possible to create a unique actuarial pricing model that can be trained as a whole. This has different advantages, which we will mention below, especially for fairness.  Different approaches have recently focused on spatial embedding \citep{blier2021rethinking, blier2021geographic} and have shown superior performance than traditional pricing strategies. These models propose to aggregate by deep neural network the geographic information into a unique/multidimensional representation by providing this information into the predictor model 
during the training. As said in \citep{blier2021geographic}, those ``{\em geographic embeddings are a fundamentally different approach to geographic models studied in actuarial science}'', since the different models are trained at the same time with the objective to predict the pure premium. It allows having aggregate geographic risks adapted for the targeted risks. 
We extend this main idea to a general framework where a unique model is trained by generating the car ($C$) and geographic pricing ($G$) components required to predict the pure premium.

\subsection{Description of the Method}\label{sec:method_AE}
\begin{figure}[h]
\centering
\includegraphics[width=0.75\textwidth]{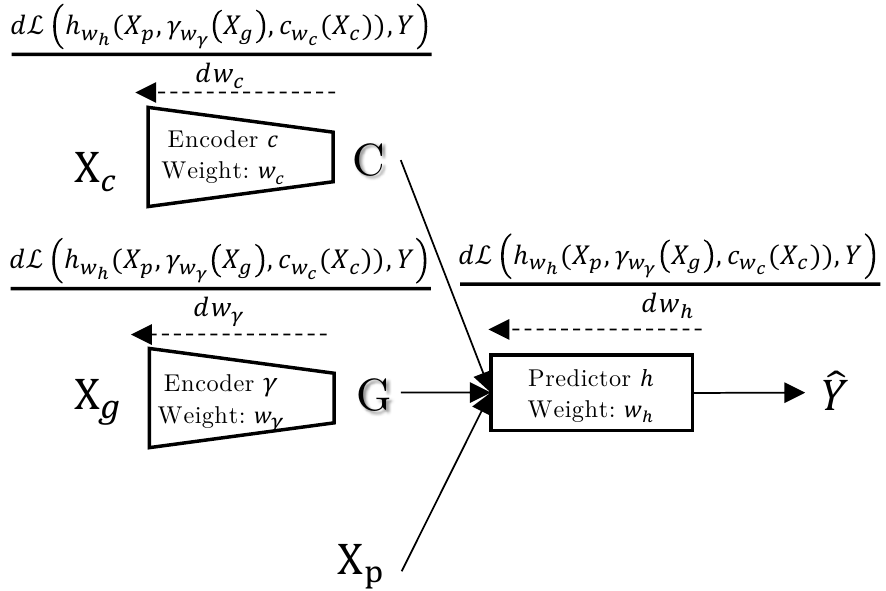}
\caption{\label{Fig:auto_enc} Pricing model via an autoencoder structure}
\end{figure}

First in a pure predictive task, we propose to find a car and a geographical aggregation via a latent representation $C$ and $G$ respectively, which both minimizes the deviation between the target $Y$ and the output prediction $\widehat{Y}$. In Figure~\ref{Fig:auto_enc}, we represent our model extension.  The output prediction is provided by a function $h_{w_{h}}(X_{p},C,G)$ where $h_{w_h}$ is a predictor 
with parameters $w_h$, which takes as input $X_{p}$ the policies information, $C$ and $G$. 
Let $c_{w_c}$ and $\gamma_{w_\gamma}$ be two neural networks with respective parameters $w_c$ and $w_\gamma$, the latent representation $C$ is generated as $c_{w_c}(X_c)$ (resp. $G$ as $\gamma_{w_\gamma}(X_g)$) with $X_c$ as the information about the car (resp. $X_g$ as the geographical information). Depending on the task or objective, we can consider the latent representation $C$ and $G$ as multi-dimensional. 
This can therefore provide a rich representation for the geographical and car ratemaking.
The mitigation procedure follows the optimization problem: 
\begin{eqnarray}
\begin{aligned}
    \argmin_{h_{w_{h}},g_{w_{g}},c_{w_{c}}} \Big\lbrace& \ \mathcal{L}(h_{w_{h}}(X_{p},\gamma_{w_{\gamma}}(X_{g}),c_{w_{c}}(X_{c})),Y) \Big\rbrace
    \label{genericfunctionAL}
\end{aligned}
\end{eqnarray}
\tolerance=1000
where \tolerance=1000  $\mathcal{L}$ is the predictor loss function 
between the 
prediction $h_{w_{h}}(X_{p},\gamma_{w_{\gamma}}(X_{g}),c_{w_{c}}(X_{c})) \in \mathbb{R}$ and the corresponding target $Y$. 
Note that smoothing can be performed on $G$ such as \citep{blier2021geographic} to avoid that some nearby regions have too high premium volatility, especially when the risk exposure is very low. 

\subsection{A Fair Pricing Model via an Autoencoder Structure}
\label{subsec:fairpriauto}
\tolerance=10000
In this section, we adapt our autoencoder pricing model by adding an adversarial structure for fairness purposes. The objective is to find a mapping of a prediction $h_{w_h(X)}$ that both minimizes the deviation from the expected target $Y$ and does not imply too much dependence on the sensitive attribute $S$, according to its definition for the desired fairness objective seen in section~\ref{sec:notations:method}.
This strategy is radically different from the previous two-stage traditional strategy seen in subsection~\ref{TP_FAIR} since the training of the car risk and the geographic risk is done simultaneously with the learning of the predictor. In this case, the objective is, unlike the previous model, to recover the essential information from $G$ and $C$ to predict $Y$ and solely neutralize the undesirable effects during the learning process. To achieve this, the back-propagation of the learning of the $c_{w_c}$ and $\gamma_{w_\gamma}$ encoders is performed at the same time as the penalization of the adversarial fairness component. It allows minimizing the deviation from the expected target $Y$ and does not imply too much dependence on the sensitivity $S$, as defined for the desired fairness objective in section~\ref{fairness_definitions}. The predictor $h_{w_h}$ is also back-propagated in the same way but takes as input the $G$ and $C$ attributes in addition to the policy contract information $X_p$. Due to $L$-Lipschitzness of GLM~\cite{van2008high,kakade2011efficient} and neural network architectures~\cite{912251}, we know that $HGR((X_g,X_c,X_p),S)\geq HGR((\gamma_{w_\gamma}(X_g),c_{w_c}(X_c),X_p),S)$. Acting on $G$ and $C$ lead to removing bias 
even for components ignored by the predictor $h_{w_{h}}$ in train. 

We present below our methodology with an $HGR$ penalization denoted as AE\_HGR algorithm. We extend this framework to other penalization such as the AE\_SIMPLE corresponding to an adversarial simple penalization and the AE\_corr to the Pearson correlation penalization.

\paragraph{Demographic Parity}

\begin{sidewaysfigure}
\includegraphics[width=1\textwidth]{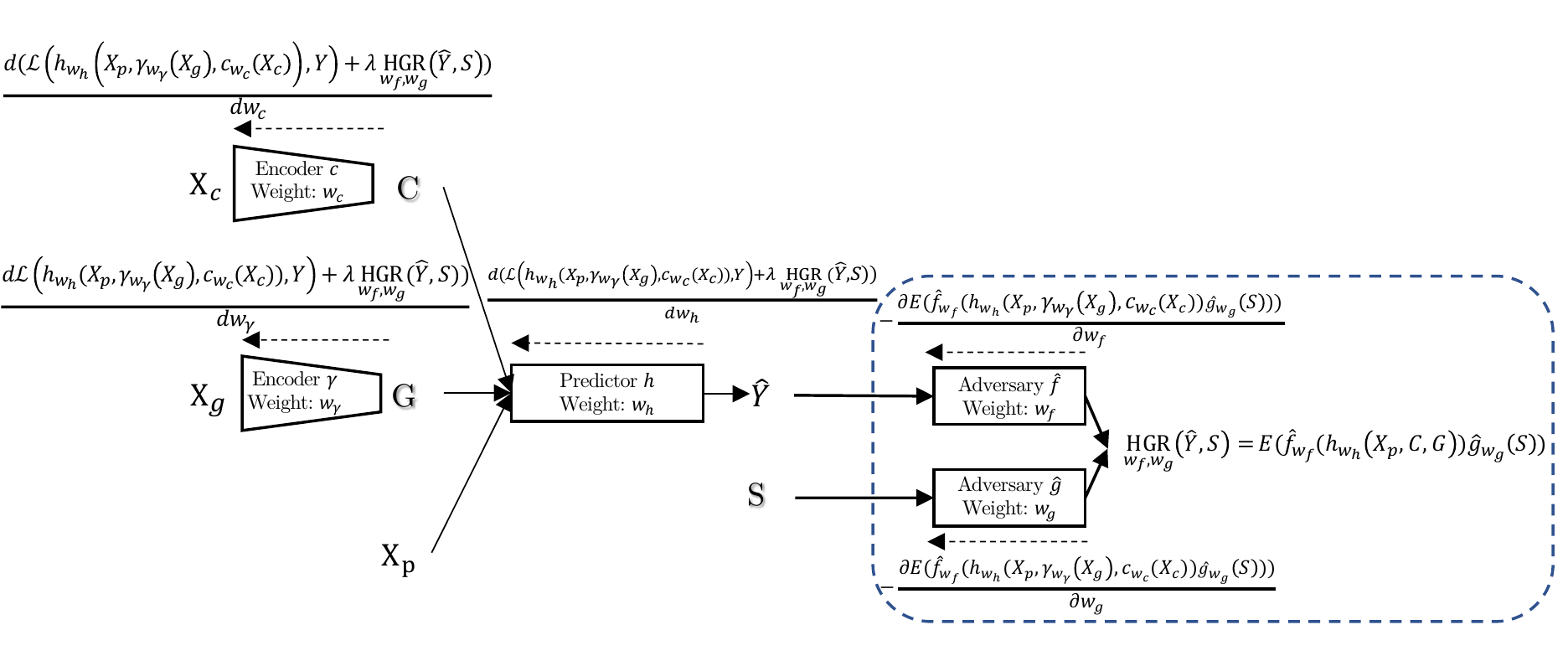}
\caption{\label{fig:OE} AE\_HGR - The Fair Demographic Parity Pricing Model via an $HGR$ autoencoder structure}
\label{fig:DP_Auto_Encoder}
\end{sidewaysfigure}

The predictor function is defined as $h_{w_{h}}(X_p,\gamma_{w_\gamma}(X_g),c_{w_c}(X_c))$ where $h$ is a predictor model that takes as input $X_p$, the geographical risk $\gamma_{w_\gamma}(X_g)$ and the car risk $c_{w_c}(X_c))$.   


The mitigation procedure follows the optimization problem: 
\begin{eqnarray}
\begin{aligned}
    \argmin_{c_{w_{c}},\gamma_{w_{\gamma}},h_{w_{h}} }\Big\lbrace\max_{{w_f,w_g}} \Big\lbrace& \ \mathcal{L}(h_{w_{h}}(X_{p},\gamma_{w_{\gamma}}(X_{g}),c_{w_{c}}(X_{c})),Y) \\ &+
    \lambda \mathbb{E}(\widehat{f}_{w_f}(h_{w_{h}}(X_{p},\gamma_{w_{\gamma}}(X_{g}),c_{w_{c}}(X_{c})))
    *\widehat{g}_{w_g}(X_{p}))\Big\rbrace\Big\rbrace
    \label{DP_Auto_Encoder}
\end{aligned}
\end{eqnarray}
where  $\mathcal{L}$ is the predictor loss function between the output prediction and the corresponding target $Y$. 
We add in this objective optimization a second term representing our $HGR$ estimation between the output prediction and the sensitive attribute $S$. It corresponds to the expectation of the products of standardized outputs of both networks ($\hat{f}_{w_{f}}$ and $\hat{g}_{w_{g}}$). The hyperparameter $\lambda$ controls the impact of the correlation loss in the optimization.

Figure \ref{fig:DP_Auto_Encoder} gives the full architecture of our adversarial learning algorithm using the neural $HGR$ estimator between the output predictions and the sensitive attribute $S$. It depicts the encoders functions $\gamma_{w_{\gamma}}$ and $c_{w_{c}}$, which respectively outputs a latent variable $G$ from $X_g$ and $C$ from $X_c$. The two neural networks $f_{w_f}$ and $g_{w_g}$, which seek at defining the most strongly correlated transformations of the output predictions $h_{\omega_h}(X_p,\gamma_{w_\gamma}(X_g),c_{w_c}(X_c))$ and $S$. The model $h_{\omega_h}$ 
outputs the prediction $\widehat{Y}$ from the information $X_p$, $G$ and $C$. The encoders aggregate the information of $G$ and $C$ from the information of $X_g$ and $X_c$ in order to maximize the performance accuracy for $h_{w_h}$ and simultaneously minimize the $HGR$ estimation finding with the adversary $f_{w_f}$ and $g_{w_g}$.
 Left arrows represent gradient back-propagation. The learning is done via stochastic gradient, alternating steps of adversarial maximization, and global loss minimization. 

The algorithm 
takes as input a training set from which it samples batches of size $b$ at each iteration. At each iteration, it first standardizes the output scores of networks $f_{w_f}$ and $g_{w_g}$ to ensure 0 mean and a variance of 1 on the batch. Then it computes 
the $HGR$ neural estimate and the prediction loss for the batch. 
At the end of each iteration, the algorithm updates 
the prediction parameters $\omega_h$ as well as encoder parameters $\omega_\gamma$ and $\omega_c$ by one step of gradient descent. 

\paragraph{Equalized Odds}
For equalized odds, we also extend the main idea of mitigation seen in section~\ref{sec:notations:method}.
The mitigation procedure follows the optimization problem: 
{\small
\begin{eqnarray}
\begin{aligned}
    \argmin_{c_{w_{c}},\gamma_{w_{\gamma}},h_{w_{h}} } \Big\lbrace\max_{w_{f_0},w_{g_0},..,w_{f_K},w_{g_K}}\Big\lbrace& \ \mathcal{L}(h_{w_{h}}(X_{p},\gamma_{w_{\gamma}}(X_{g}),c_{w_{c}}(X_{c})),Y)  \\
    &+
    \frac{1}{K+1}\sum_{y \in \Omega_{Y}}
    \lambda_{y}\mathbb{E}_{(X,S)\sim\mathcal{D}_{y}}(\widehat{f}_{w_{f_y}}(h_{w_{h}}(X_{p},\gamma_{w_{\gamma}}(X_{g}),c_{w_{c}}(X_{c}))*\widehat{g}_{w_{g_y}}(S))\Big\rbrace\Big\rbrace
    \label{genericfunctionCHAP7}
\end{aligned}
\end{eqnarray}
}
\noindent
where $\mathcal{D}_{y}$ corresponds to the distribution of pair $(X,S)$ conditional on ${Y=y}$ and $K=\#\Omega_{Y}-1$. 
The hyperparameters $\lambda_{y}$ control the impact of the dependence loss for the different number event objective. 
The penalization enforces the independence between the output prediction 
and the sensitive $S$ only for the cases where $Y=y$. It enforces naturally the mitigation of equalized odds since it enforces the mitigation of biases for demographic parity for each number of events.



We evaluate the performance of these fair algorithms empirically with respect to performance accuracy and fairness. We conduct the experiments on a synthetic scenario and 
real-world datasets. 

\newpage
\section{Synthetic Scenario for Target and Sensitive Binary Features}
\label{sec:toyscenario}
We illustrate the fundamental functionality of our proposal with a simple synthetic scenario that was inspired by the Red Car example~\cite{Kusner2017,grari2019fairness}. The subject is a pricing algorithm for a fictional car insurance policy. The purpose of this exercise is to train a fair classifier that estimates the claim likelihood without incorporating any gender bias in terms of \emph{Demographic Parity}. We compare for this objective, the fair traditional pricing structure with an $HGR$ penalization denoted as TP\_HGR shown in the subsection~\ref{TP_FAIR} and the fair pricing autoencoder structure as shown in the subsection~\ref{subsec:fairpriauto} denoted as AE\_HGR. 
We focus on the general claim likelihood and ignore the severity or cost of the claim. Further, we only consider the binary case of claim or not (as opposed to a frequency). We assume that the claim likelihood only depends on the aggressiveness and the inattention of the policyholder. To make the training more complex, these two properties are not directly represented in the input data but are only indirectly available through correlations with other input features. We create a binary label $Y$ with no dependence on the sensitive attribute $S$. Concretely, we use as protected attribute the \textit{gender} of the policyholder. The unprotected attributes \textit{color}, the maximum \textit{speed} of the cars, and the average \textit{salary} of the policyholder's area are all caused by the sensitive attribute. 
In our data distribution, the \textit{color} and the maximum \textit{speed} of the car are strongly correlated with both \textit{gender} and aggressiveness. \textit{Age} is not correlated with \textit{gender}. 
However, \textit{age} is correlated with the inattention of the policyholder. Thus, the latter input feature is actually linked to the claim likelihood. 

\noindent%
\begin{minipage}[t]{0.5\textwidth}

First, we generate the training samples ${(x_{p_{i}},x_{g_{i}},x_{c_{i}},s_{i},y_{i})}_{i=1}^{n}$. The unprotected attribute $x_{p_{i}}=age_{i}$ represents the policy information with the \textit{age} 
of the policyholder, $x_{g_{i}}=Sal_{i}$ represents the geographical information with the average \textit{salary} of the area and $x_{c_{i}}=(Col_{i},Sp_{i})$ represent the \textit{colors} and the maximum \textit{speed} of the car.
\end{minipage}
\begin{minipage}[t]{0.4\textwidth}
  \centering\raisebox{\dimexpr \topskip-\height}{%
  \includegraphics[width=\textwidth]{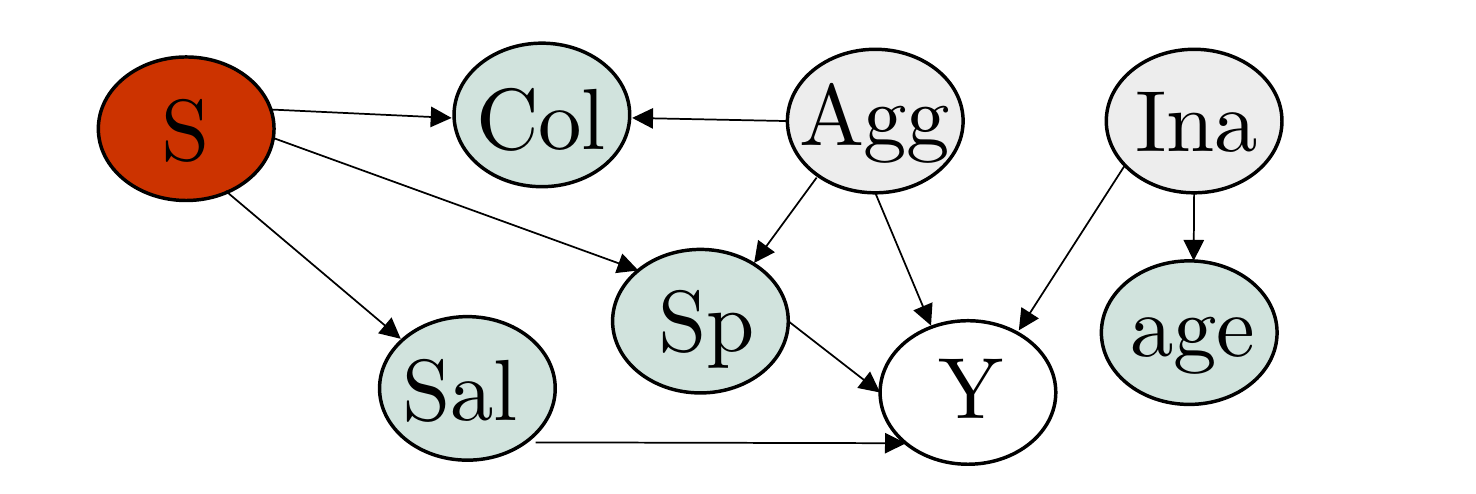}}
  \captionof{figure}{Causal Graph - Synthetic Scenario} 
  \label{fig:correlation_bounds}
\end{minipage}\hfill

\begin{figure}[h]
  \centering
  \includegraphics[scale=0.30,valign=t]{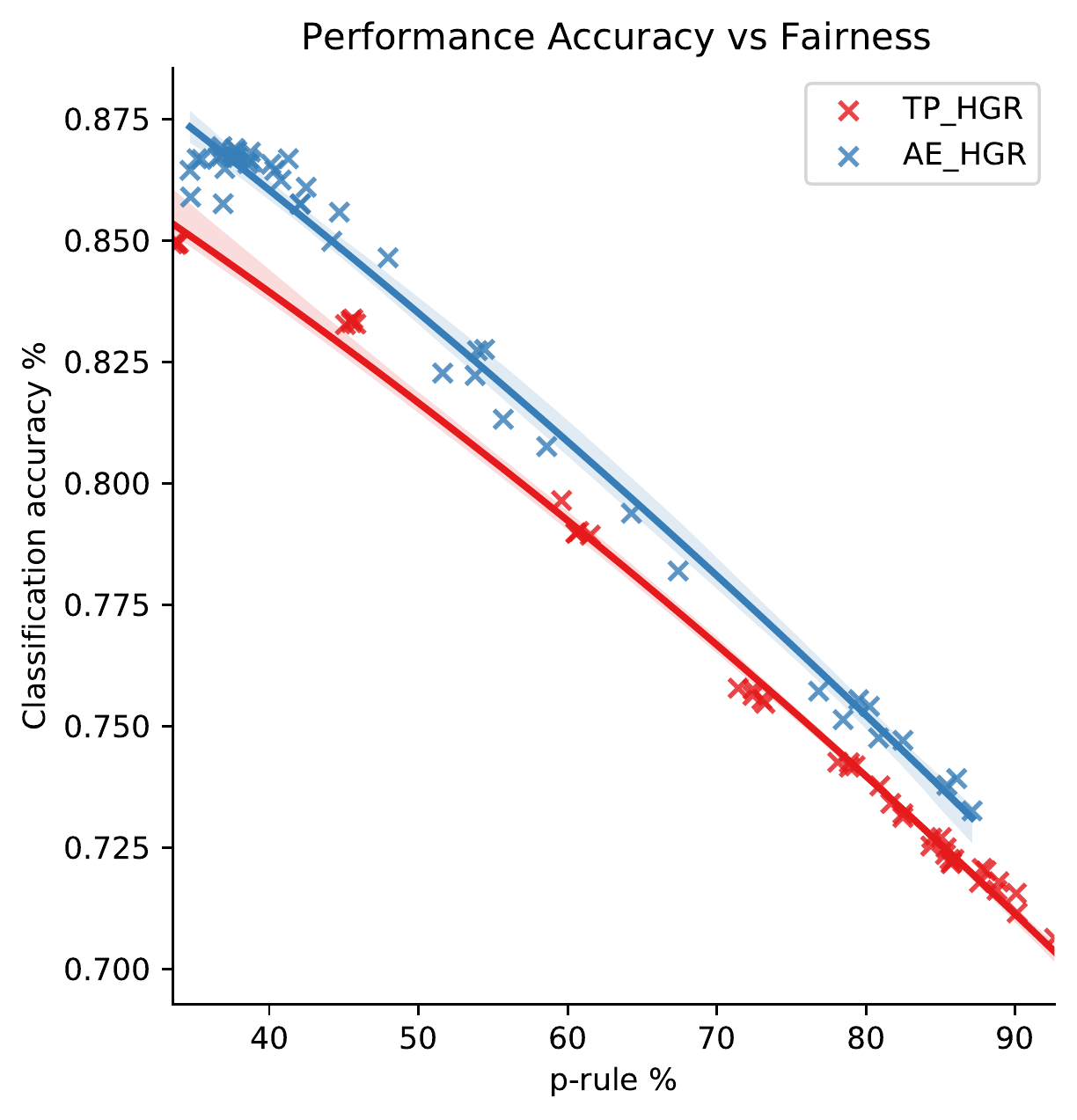}
  \includegraphics[scale=0.30,valign=t]{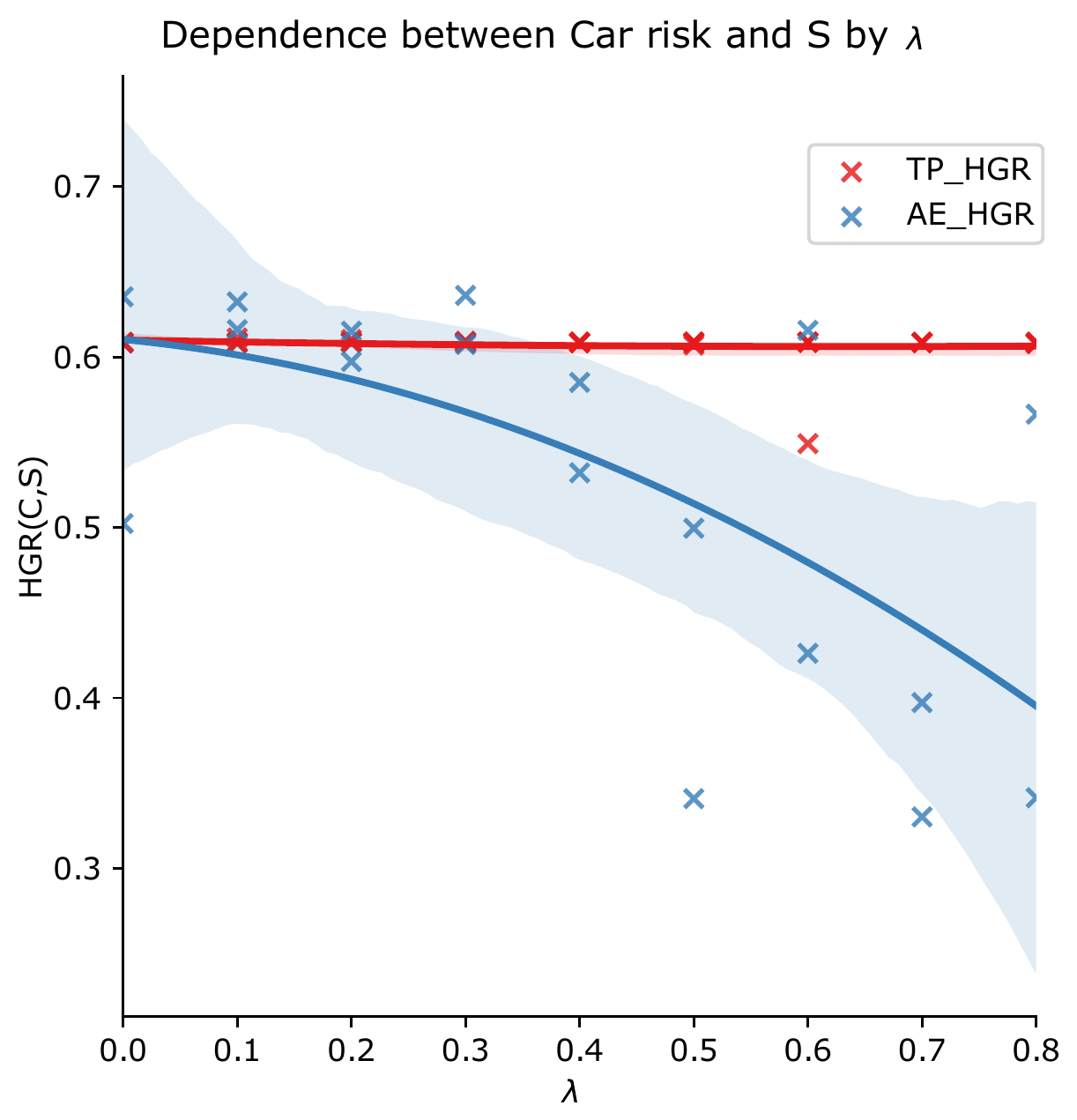} 
  \includegraphics[scale=0.30,valign=t]{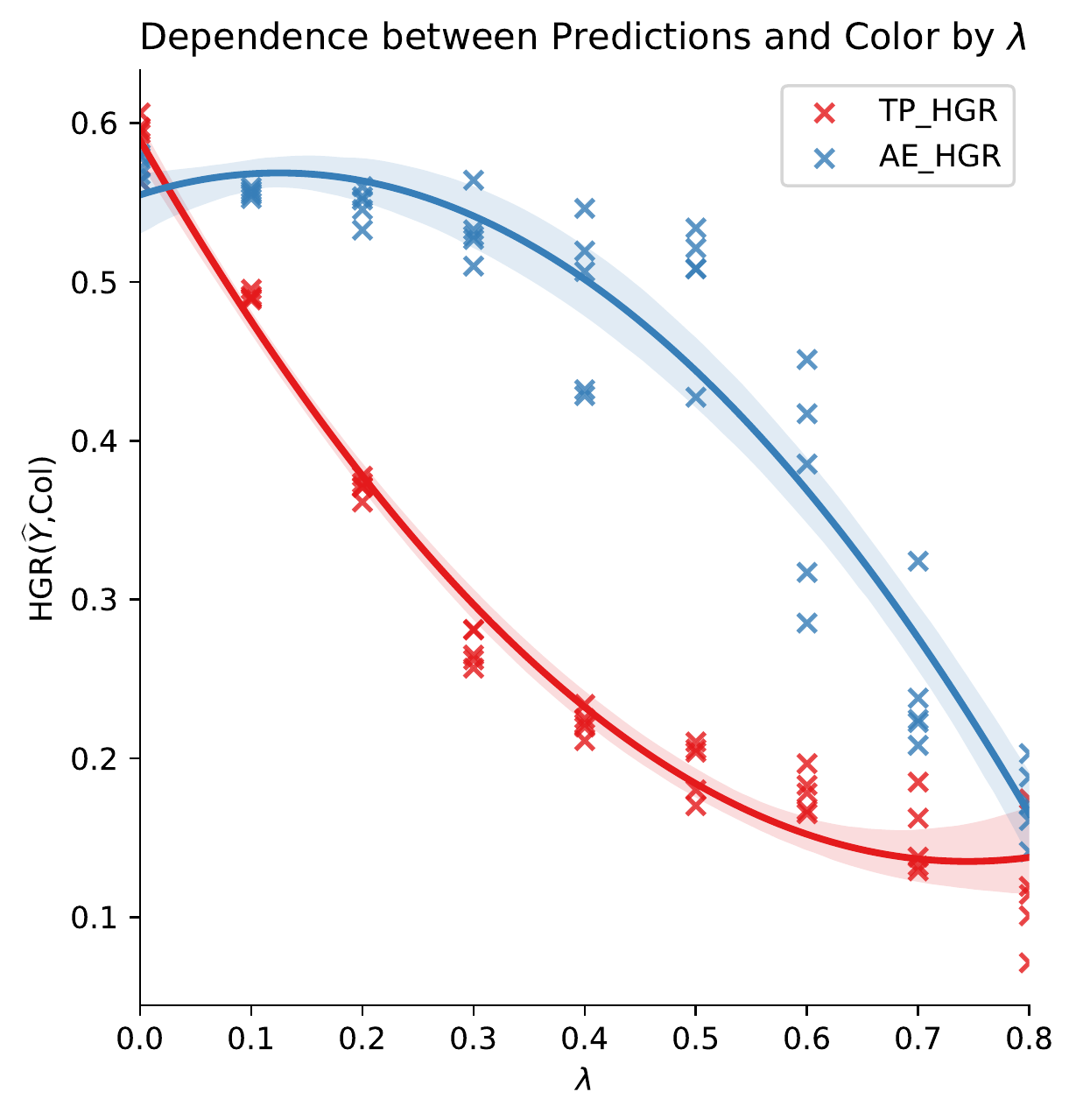}
  \includegraphics[scale=0.30,valign=t]{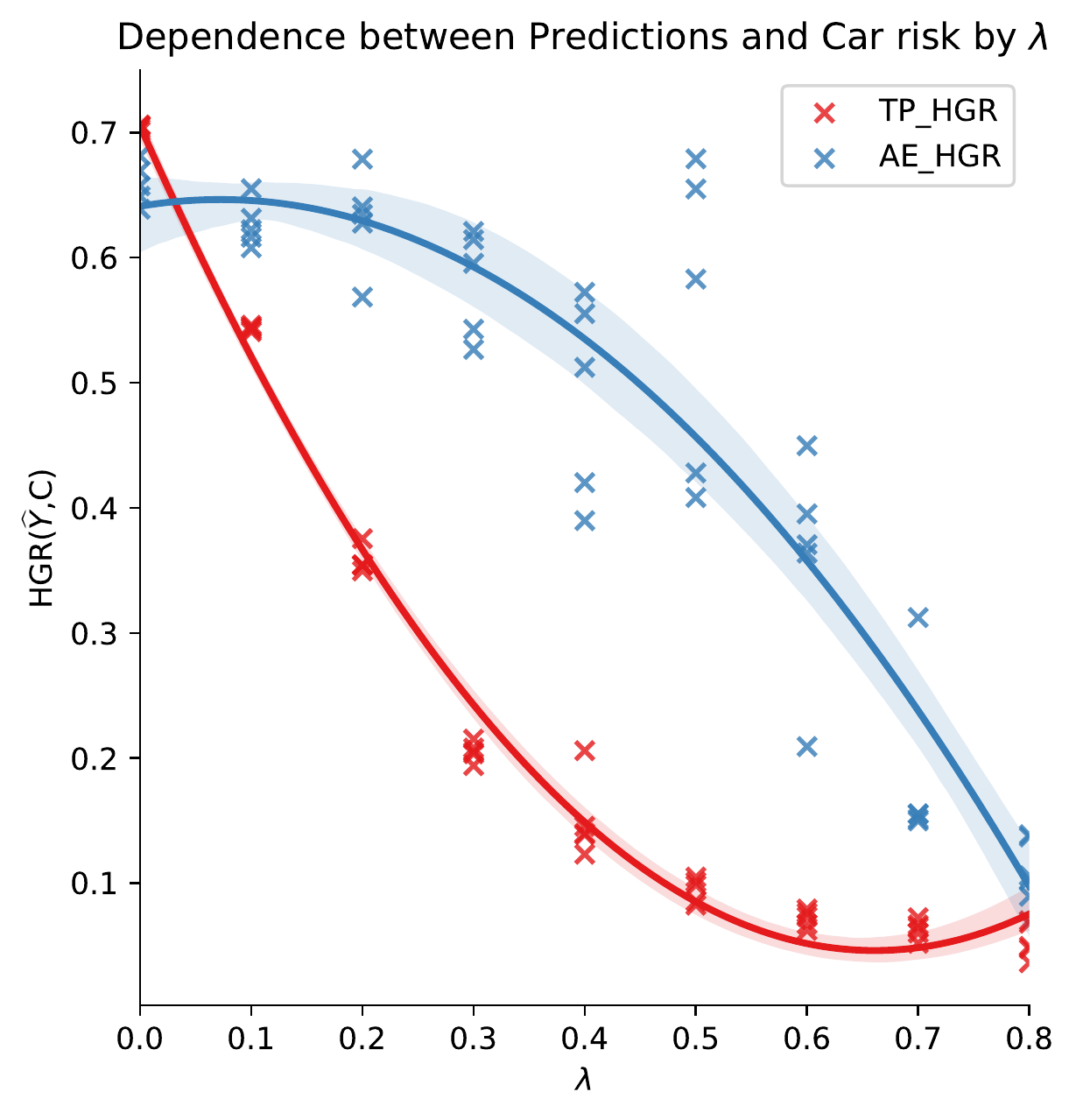}
  \caption{
 Impact of hyperparameter $\lambda$ (Synthetic Scenario): (Left) 
 For all levels of predictive performance accuracy, our proposed method outperforms the traditional fair pricing model. (Middle left) Higher $\lambda$ values produce a fairer car risk to our approach than traditional car risk where biases mitigation is not performed. (Middle right and Right) Higher $\lambda$ values significantly reduce the dependence between output predictions and the car risk, as well as for the geographical risk in comparison to our proposed model. 
  \label{fig:Toy_scenario_DP_results}}
\end{figure}

We plot the performance of these different approaches in 
Figure~\ref{fig:Toy_scenario_DP_results}.
In the 
first graph, we can see the 
curves of accuracy against the fairness metric $p$-rule during the training phase. Note that on the top of the left of these curves, the $\lambda$ hyper-parameters are fixed to $0$ for the two models, therefore, only the performance accuracy is optimized during the training for this case. We observe that, for all levels of fairness (controlled by the mitigation weight $\lambda$ in the two approaches), the model via autoencoder outperforms the traditional two-stage model. 
We attribute this to the ability of the autoencoder to extract a useful car and geographic risk. The traditional pricing structure has significant limitations in achieving fairness. In the middle left graph, as expected for the autoencoder model, we observe that the dependence between car risk and the sensitive $S$ is lower for higher $\lambda$. However, this is not the case for the traditional model, where we observe stagnation since the minimax fairness optimization is performed only on the final predictor model. Furthermore, we observe on the two most right graphs that the bias reduction is at the expense of the essential unbiased information 
for the traditional one. This information is too strongly reduced for the latter, in contrast to the autoencoder model where the dependence between the car risk and the prediction is higher while being less biased. We attribute this to fact that the traditional pricing approach misses the relevant information
about $G$ and $C$ for predicting $Y$. Since the car and the geographical risks for the traditional pricing are not re-trained in the fairness optimization, there is a risk to lose information about $Y$ (see the discussion in the subsection \ref{TP_FAIR}). Whereas our autoencoder approach generates fair geographical and car components, 
and keeps the most information on $Y$ as possible.

\section{Binary Task on a Real-World Dataset}
\label{sec:res_bin}
Our experiments on the binary task are performed on the pricingame 2015 dataset \cite{pricinggame15,charpentier2014computational} that contains $100,000$ TPL policies for private a motor insurance. We perform the fairness objective of \emph{Demographic Parity}. 
We compare for this objective, the fair traditional pricing structure with the different fairness penalization seen in subsection~\ref{TP_FAIR} denoted as {TP\_Corr} (i.e., w.r.t. the Pearson coefficient),
{TP\_Simple} for the simple adversarial and {TP\_HGR} for the $HGR$ adversarial. We also compare our proposal approach, the fair pricing autoencoder structure as shown in the subsection~\ref{subsec:fairpriauto} denoted as {AE\_Corr} for the Pearson coefficient,  {AE\_Simple} for the Simple adversarial and the {AE\_HGR} for the $HGR$ adversarial.

\begin{figure}[h]
  \centering
  \includegraphics[scale=0.42,valign=t]{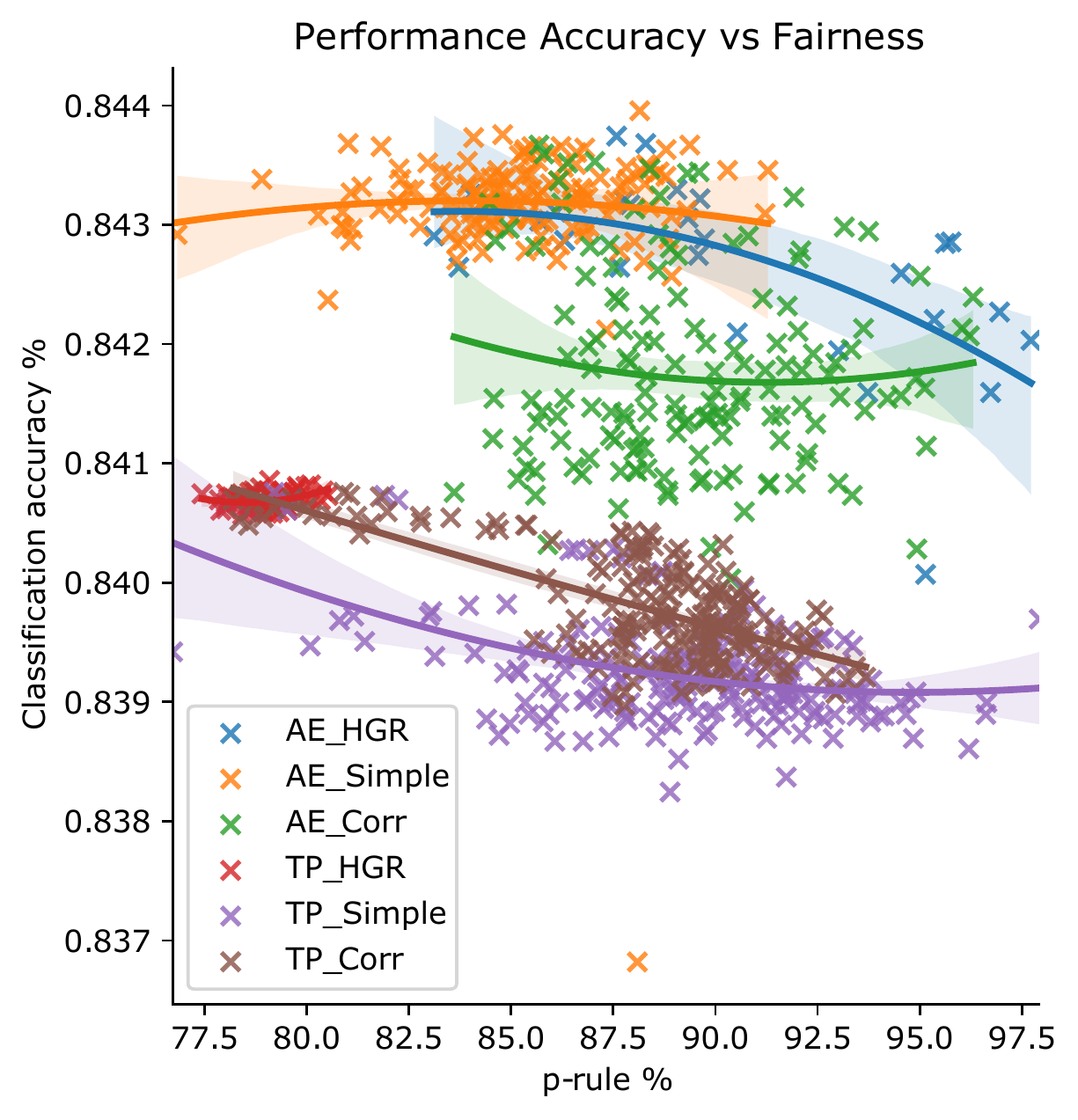}
  \includegraphics[scale=0.42,valign=t]{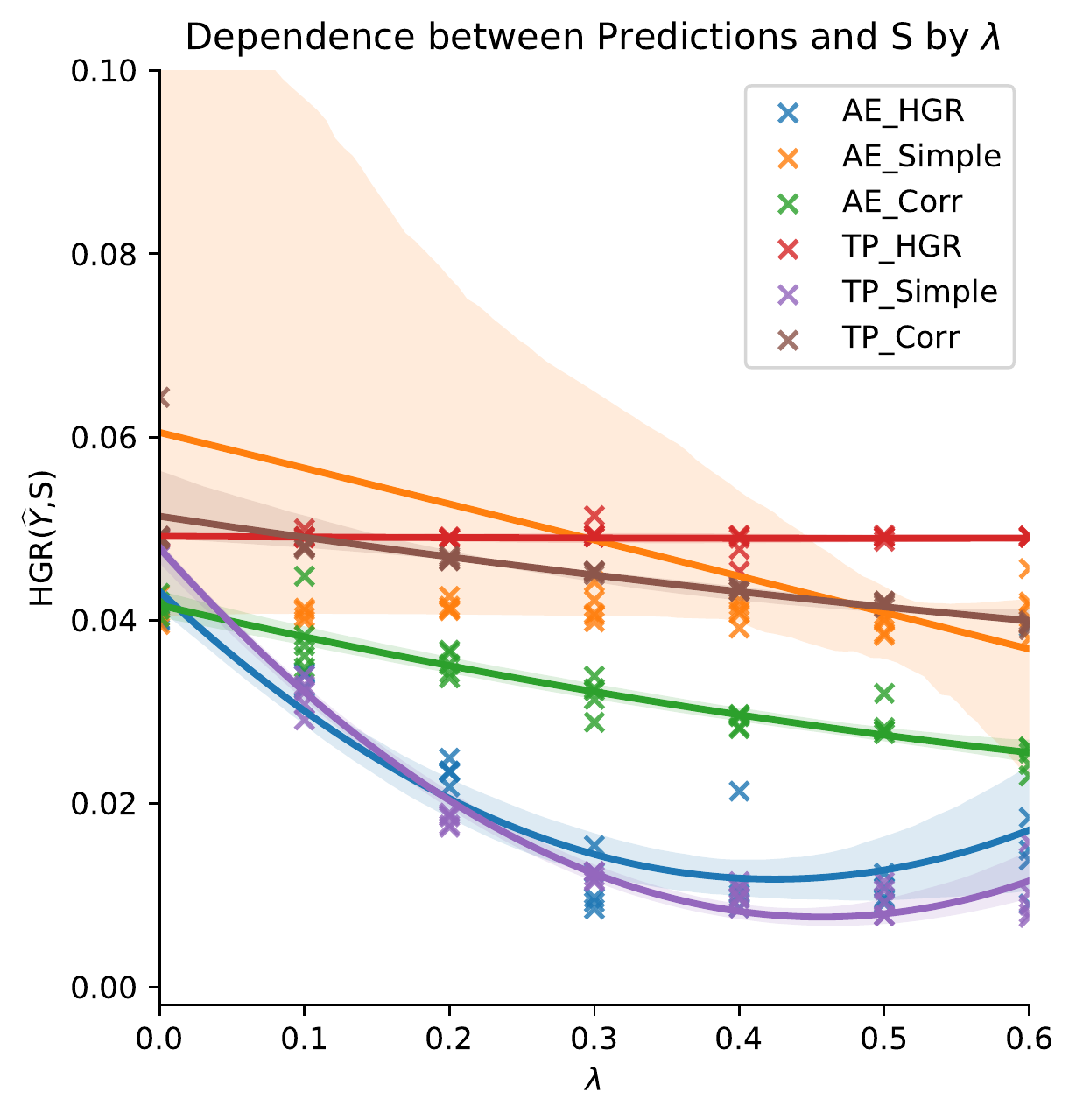}
  \\ 
  \includegraphics[scale=0.42,valign=t]{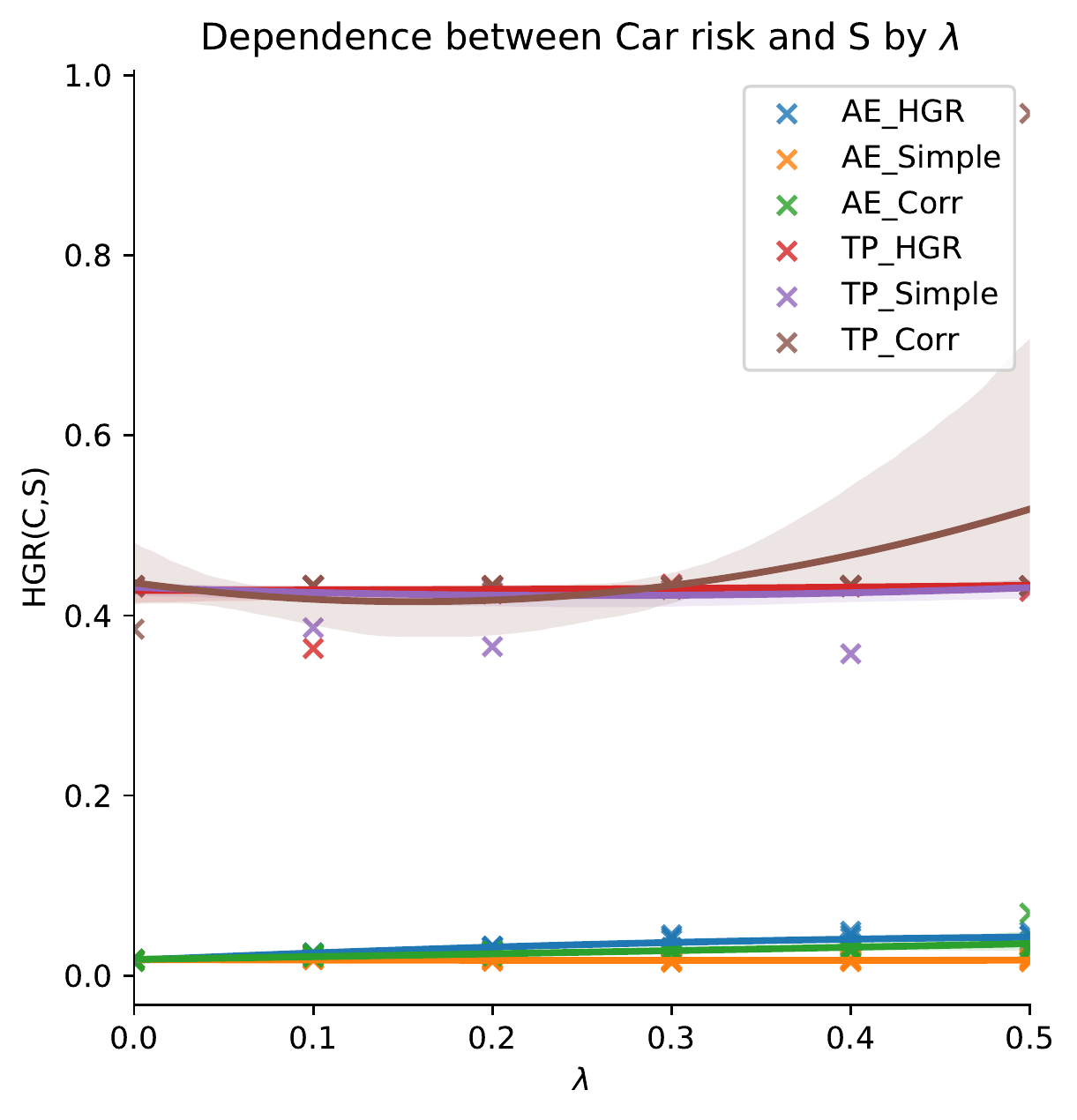}
  \includegraphics[scale=0.42,valign=t]{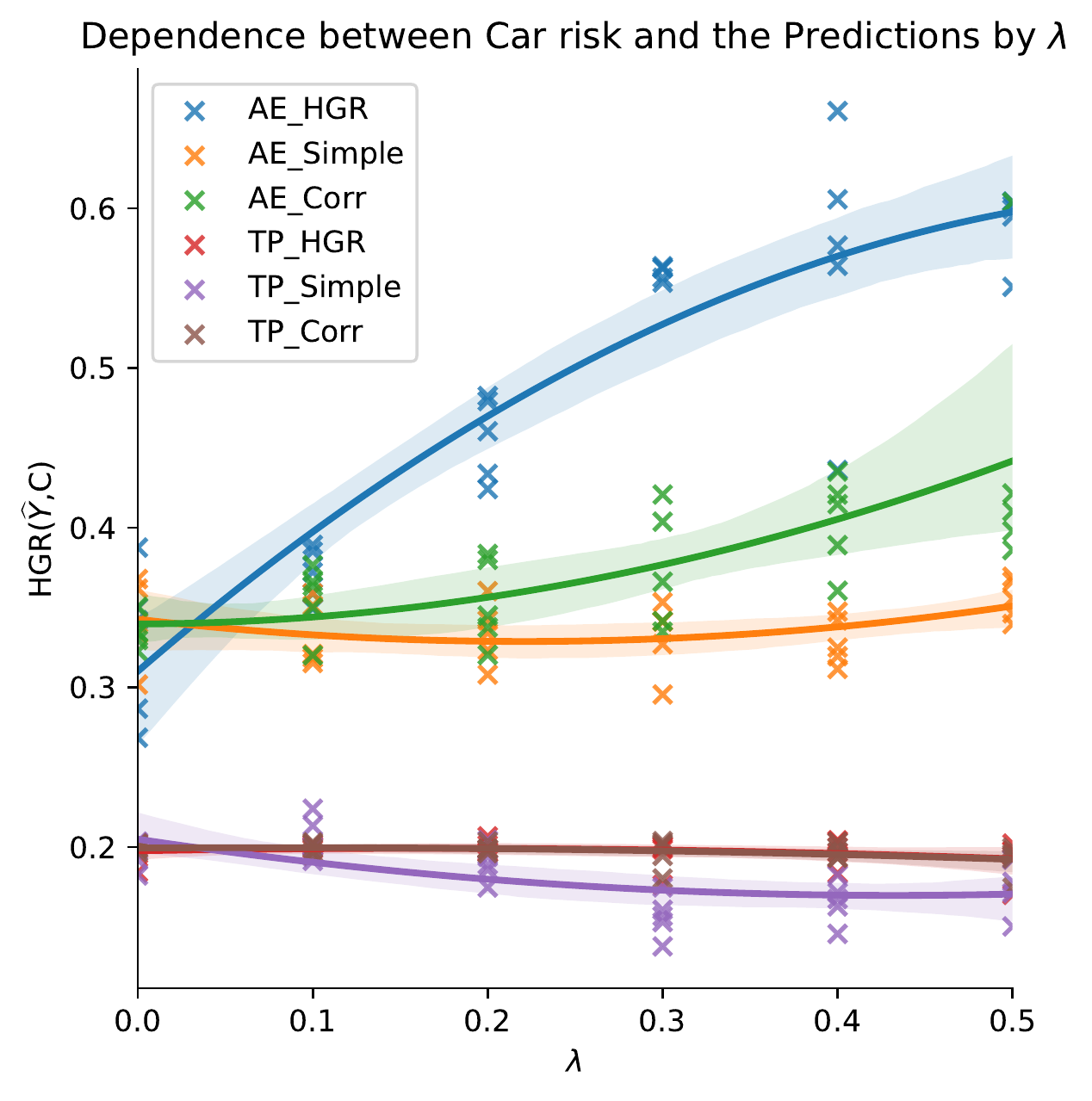}
  \\
  \includegraphics[scale=0.42,valign=t]{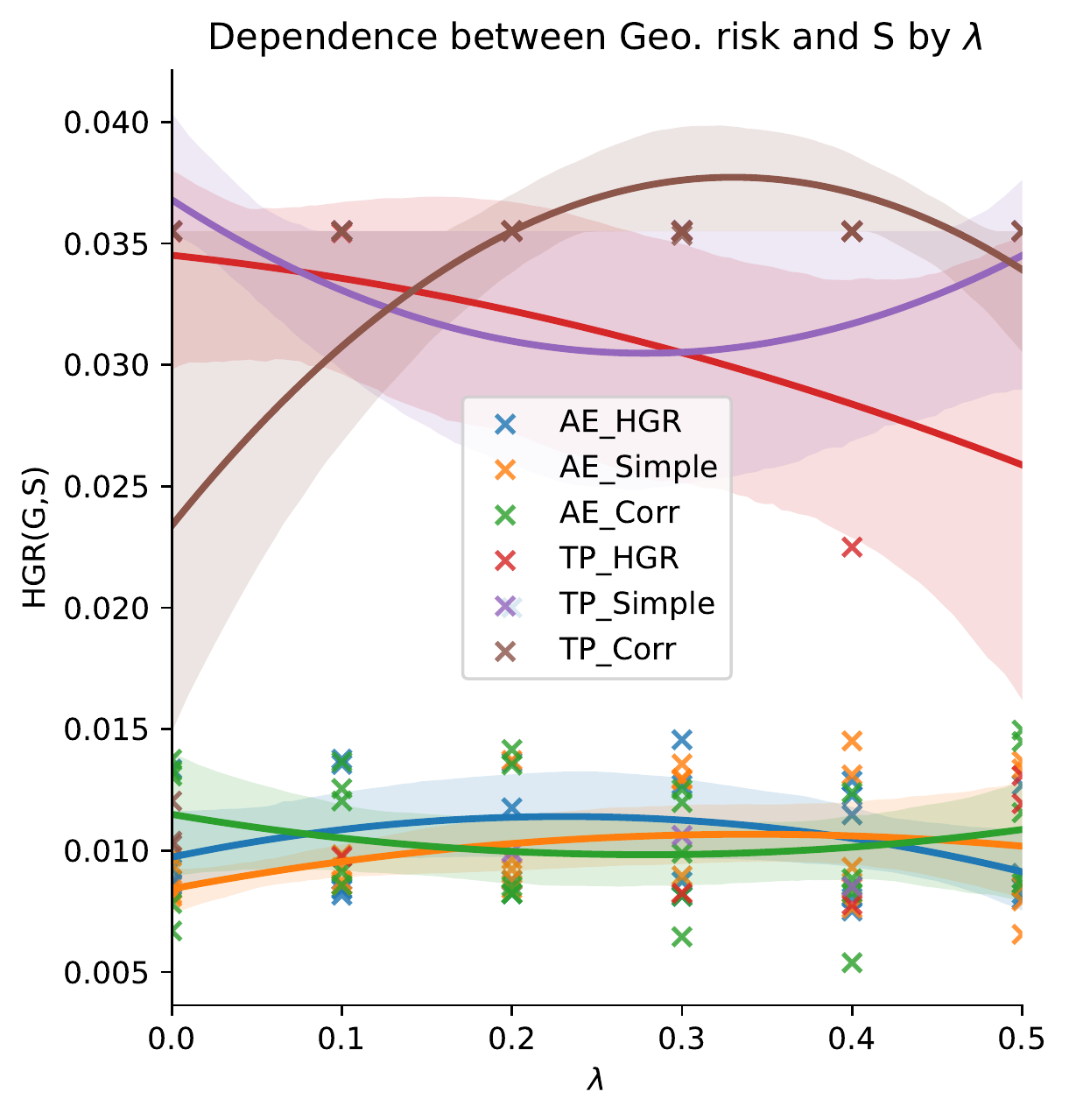}
  \includegraphics[scale=0.42,valign=t]{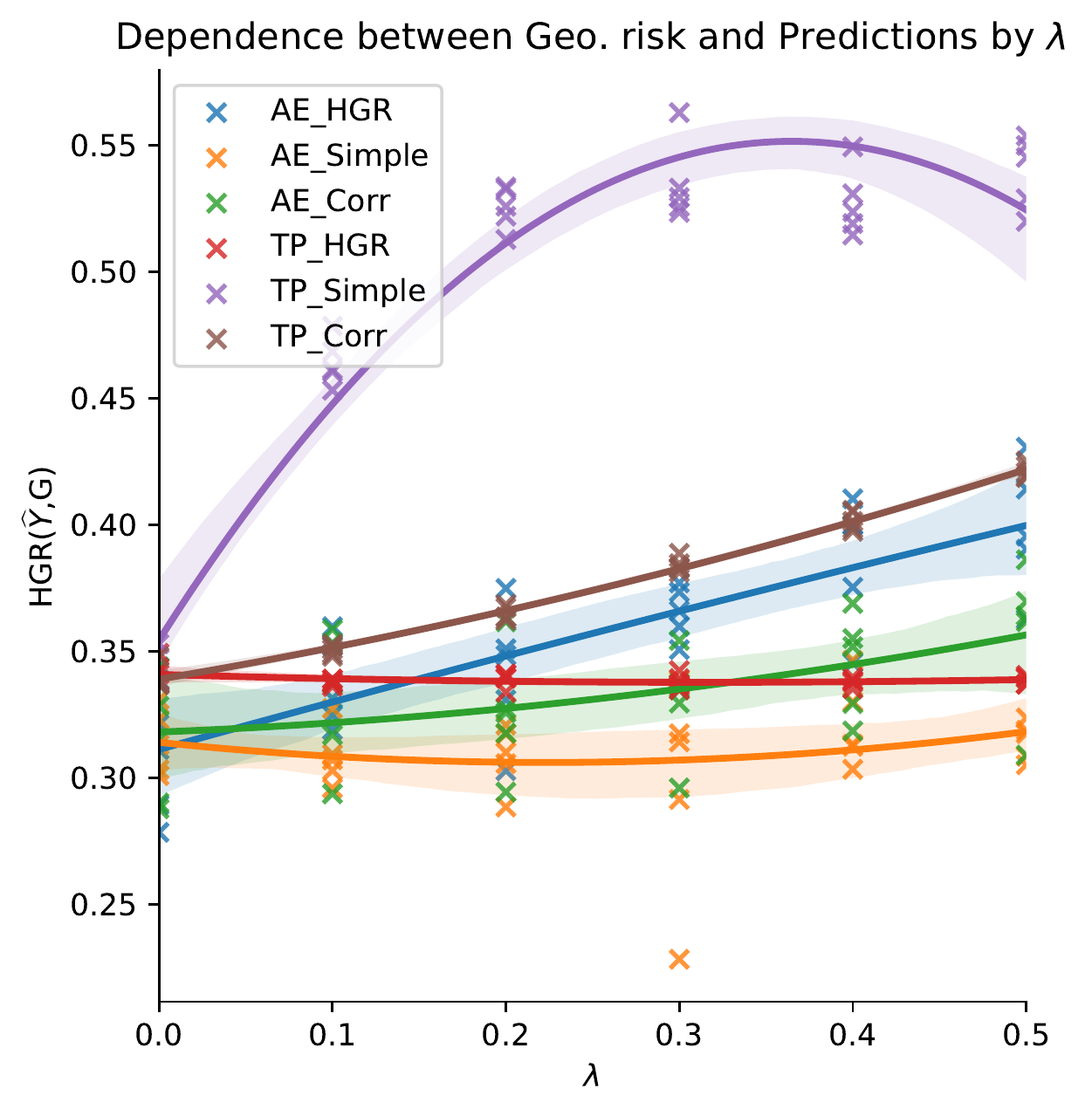}

  \caption{Scenario 1 - Binary objective - Pricingame 2015 dataset
   \label{fig:S1_BO}}
\end{figure}

For all data sets, we repeat five experiments by randomly sampling two subsets, 80\% for the training set and 20\% for the test set. 
For this binary objective, we propose to predict the TPL claim of policyholders from \emph{Pricingame 2015} dataset \cite{charpentier2014computational}. 
We considers the sensitive variable $S$ as the \emph{gender} of the policyholder (i.e., $1$ for male or $0$ for female),  $x_c$ as the set of variables \emph{Type}, \emph{Category}, \emph{Group1}, \emph{Value} of the car, $x_g$ as the \emph{Density} feature and all the remaining variables for $x_p$ (more information in \cite{dutang2020package}).
We report the average of the accuracy ($ACC$), 
the mean of the $HGR$ metric $HGR_{NN}$ \cite{grari2019fairness}, the P-rule in percentage representing the level of Demographic Parity (section~\ref{sec:Binary_Setting}), and the $FairQuant$ metric \cite{grari2019fairness} 
on the test set. 
We plot the performance of these different approaches by displaying the predictive performance ($ACC$) 
against the P-rule for Demographic Parity in the two upper left most graphs in Figure~\ref{fig:S1_BO}. 
For all algorithms, we clearly observe that the predictive performance decreases when fairness increases. We note that, for all levels of fairness (controlled by the mitigation weight $\lambda$ in every approach), our proposed methods via autoencoder outperform the traditional algorithm. The three approaches via autoencoder (i.e., AE\_HGR, AE\_Simple, AE\_Corr) obtain better accuracy for all level of P-rule. As expected, in the middle and  bottom graphs of Figure~\ref{fig:S1_BO}, we observe that the three traditional approaches obtain more biased car and geographic components (higher HGR dependence) for all levels of fairness. We attribute this result, as in the synthetic scenario, to the ability of our approach to extract useful components from the mitigated car and geographic risks. In contrast, the traditional approach suffers significantly from merely mitigating biases on the predictor model only. Also, we note that the approach via $AE\_HGR$ allows to obtain the best performance (except some points for very low levels of fairness, at the left of the curves) and allows to obtain stronger fairness (P-rule at $97.5\%$). As already observed in another context in \cite{grari2019fairness}, mitigating biases 
with an HGR penalization allow better generalisation abilities at test time.  

\begin{figure}[h]
  \centering
  \includegraphics[scale=0.32  ,valign=t]{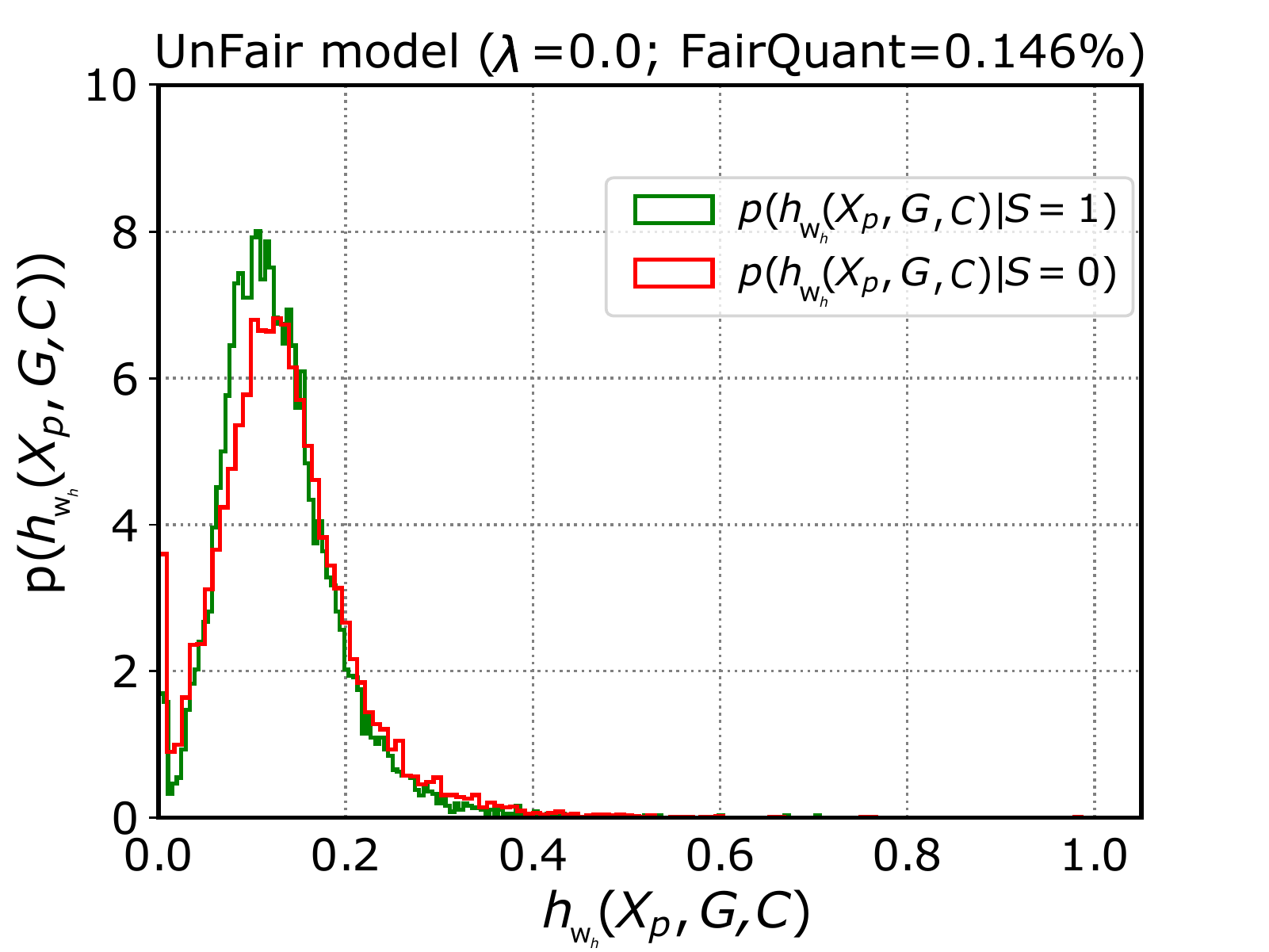}
  \includegraphics[scale=0.32,valign=t]{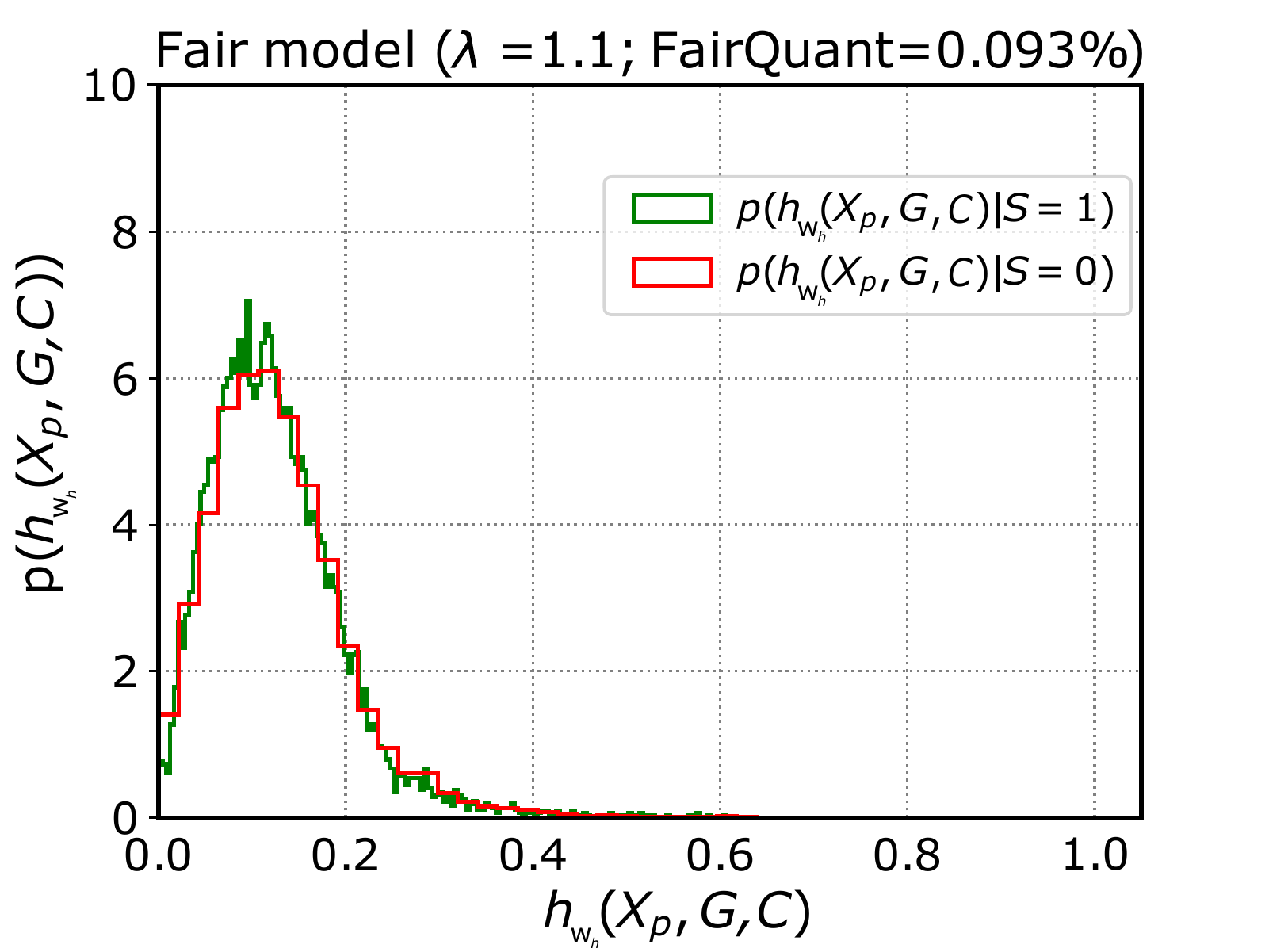}
   \includegraphics[scale=0.32,valign=t]{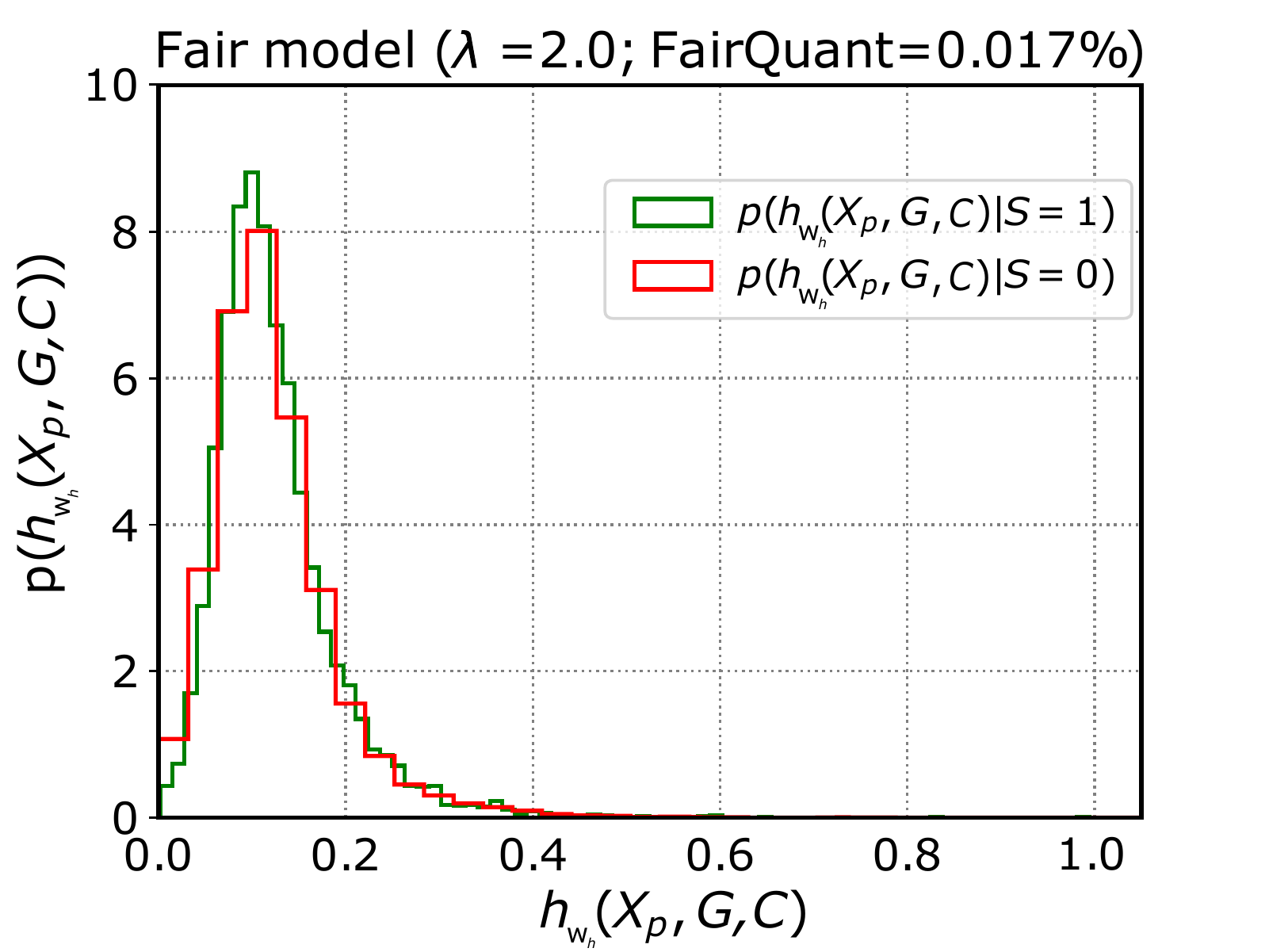}
  \caption{Distributions of the predicted probabilities given the sensitive attribute $S$ (Pricingame 2015 dataset). Higher $\lambda$ values produce more aligned distributions between men and women.
   \label{fig:distributions}}
\end{figure}

In Figure~\ref{fig:distributions}, we plot the distribution of the predicted probabilities for each demographic group of the sensitive attribute $S$ for 3 different AE\_HGR models: An unfair model ($\lambda$ = 0),  a mildly fair model ($\lambda$ = 1.1) and a strongly fair one ($\lambda$ = 2.0).
For the unfair model, the distribution differs most for the lower probabilities. As expected, we observe that the distributions are more aligned with a higher $\lambda$. We also observe better results in term of the $\emph{FairQuant}$ fairness metric from $0.146\%$ for $\lambda=0$ to $0.017\%$ for $\lambda=1.1$.


\newpage


\noindent


\section{Frequency Task on Real-World Datasets}
\label{sec:res_freq}
Our experiments on the frequency task 
are performed on two real-world datasets: the pricingame 2015 \cite{pricinggame15,charpentier2014computational} and the pricingame 2017 \cite{pricinggame17}. Each of them contains $100,000$ TPL policies
for private motor insurance. In these data sets, we perform the two objectives of fairness \emph{Demographic Parity} and \emph{Equalized Odds}
We propose on the frequency task to predict the number of third-party claims from \emph{Pricingame 2015} and \emph{Pricingame 2017}. For the \emph{Pricingame 2017} dataset,  we considers $x_c$ as the set of variables \emph{vh\_age}, \emph{vh\_cyl},	\emph{vh\_din},	\emph{vh\_speed}, \emph{vh\_value},	\emph{vh\_weight}, \emph{vh\_make}, \emph{vh\_model}, \emph{vh\_sale\_begin},\emph{sale\_end}, \emph{vh\_type}, $x_g$ as the set of variables representing more than $100$ hundreds external features collected from French INSEE organism via the $INSEE\_code$ (more details in the code\footnote{1}) and all the remaining variables for $x_p$. For the pricingame 2015 dataset, the set of variables $x_c$ and $x_d$ are similar to the binary scenario above. 

\begin{figure}[h]
  \centering
  \includegraphics[scale=0.30,valign=t]{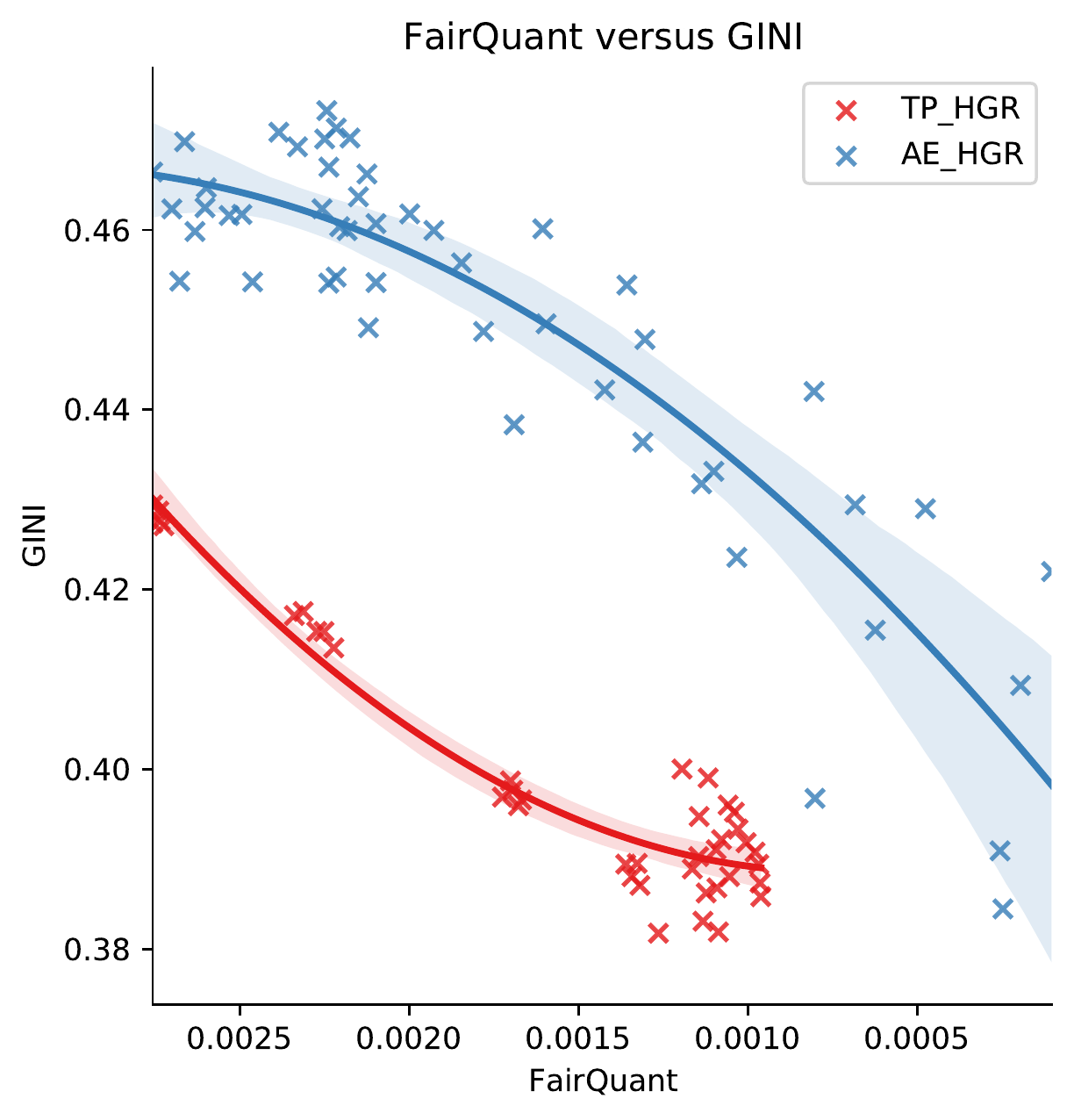}
  \includegraphics[scale=0.30,valign=t]{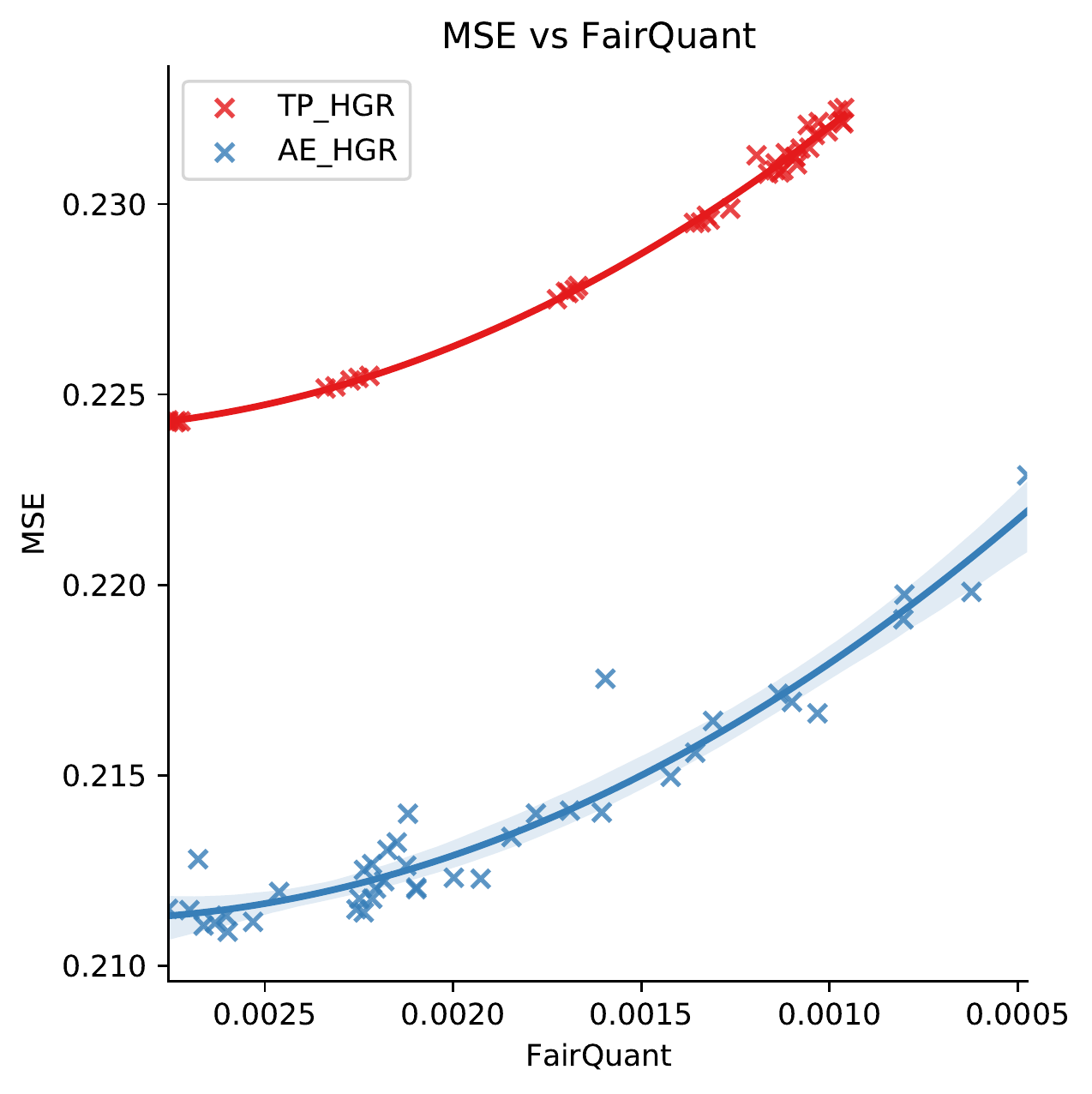}
  \includegraphics[scale=0.30,valign=t]{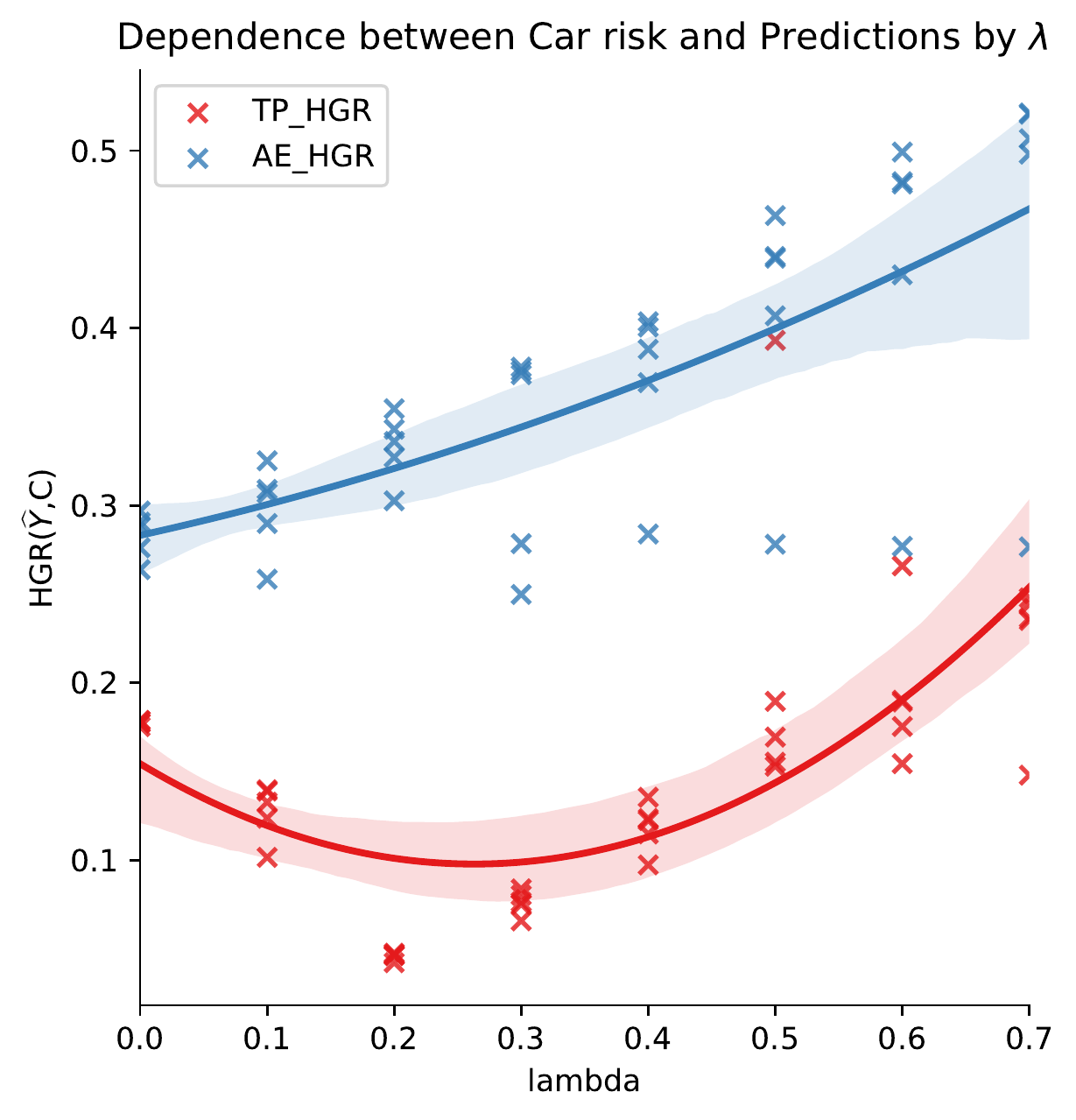}
  \includegraphics[scale=0.30,valign=t]{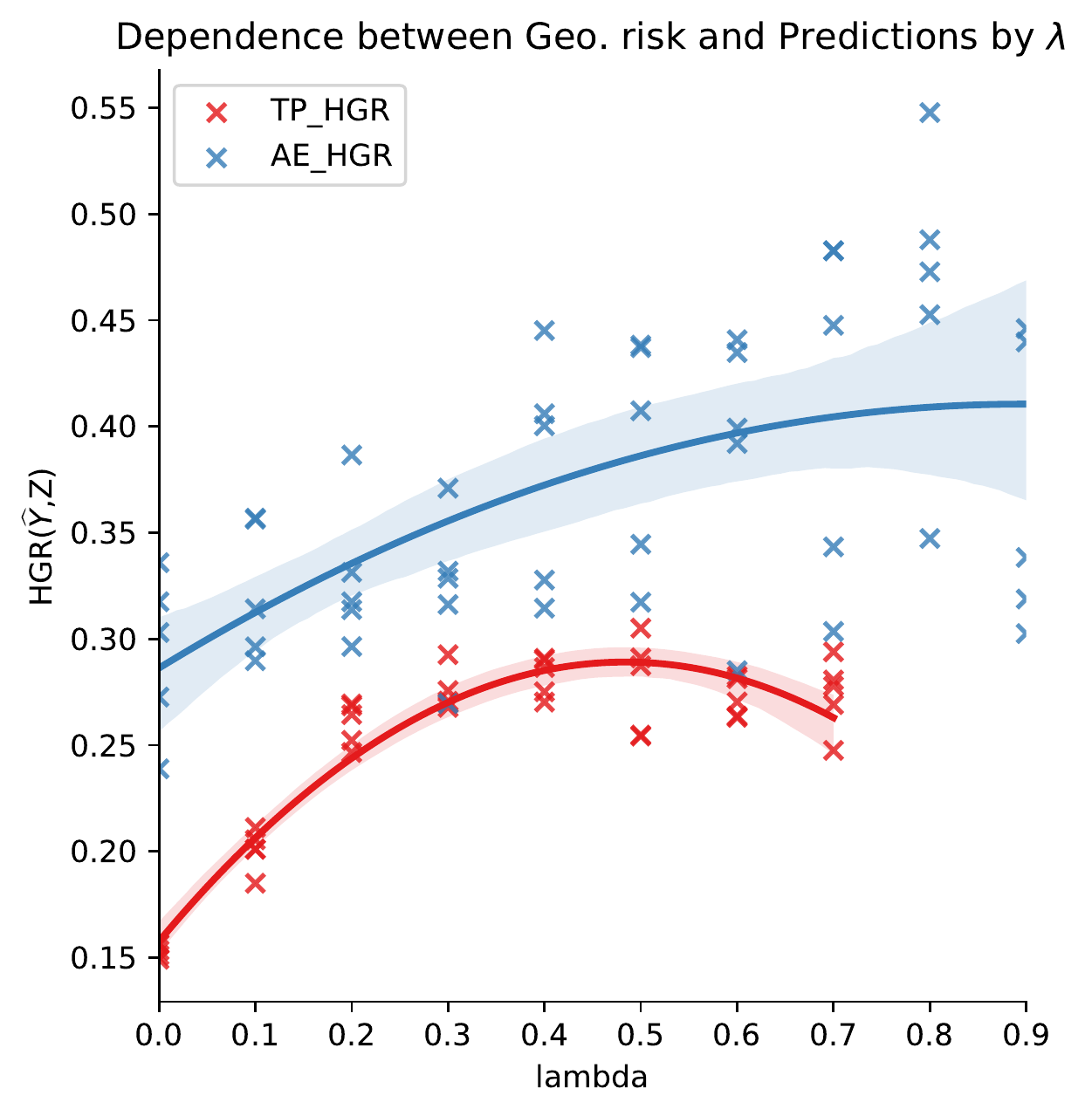}
  \caption{Scenario 2 - Frequency objective 
   (Left and Middle left) For all levels of predictive performance, \emph{GINI} or \emph{MSE}, our proposed method outperforms the traditional fair pricing model. (Middle right and right) Higher $\lambda$ values significantly reduce the dependence between output predictions and the car risk, as well as for the geographical risk in comparison to our proposed model.
   \label{fig:S2_FO}}
\end{figure}

We observe for the \emph{Demographic parity} task in Figure~\ref{fig:S2_FO}, as the binary task, that the method with autoencoder outperforms the traditional one. In the two left graphs, we observe that that the autoencoder approach (i.e., blue curves) obtain better $\emph{GINI}$ and  $\emph{MSE}$ for all level of $Fair\_Quant$. In the two right graphs in Figure~\ref{fig:S2_FO}, we observe that the autoencoder  prediction dependencies with the car risk and the geographical risk
are both more important than the traditional one, and this is valid for all levels of fairness.  We attribute this result, to the ability of our approach to extract useful components from the mitigated car and geographic risks. Since the car risk and the geographical risk for the traditional pricing are not re-trained in fair fairness optimization, there is a risk to miss the relevant information
about $G$ and $C$ for predicting $Y$.


\begin{figure}[H]
\centering
\subfloat[Demographic Parity task]{
\includegraphics[scale=0.35,valign=t]{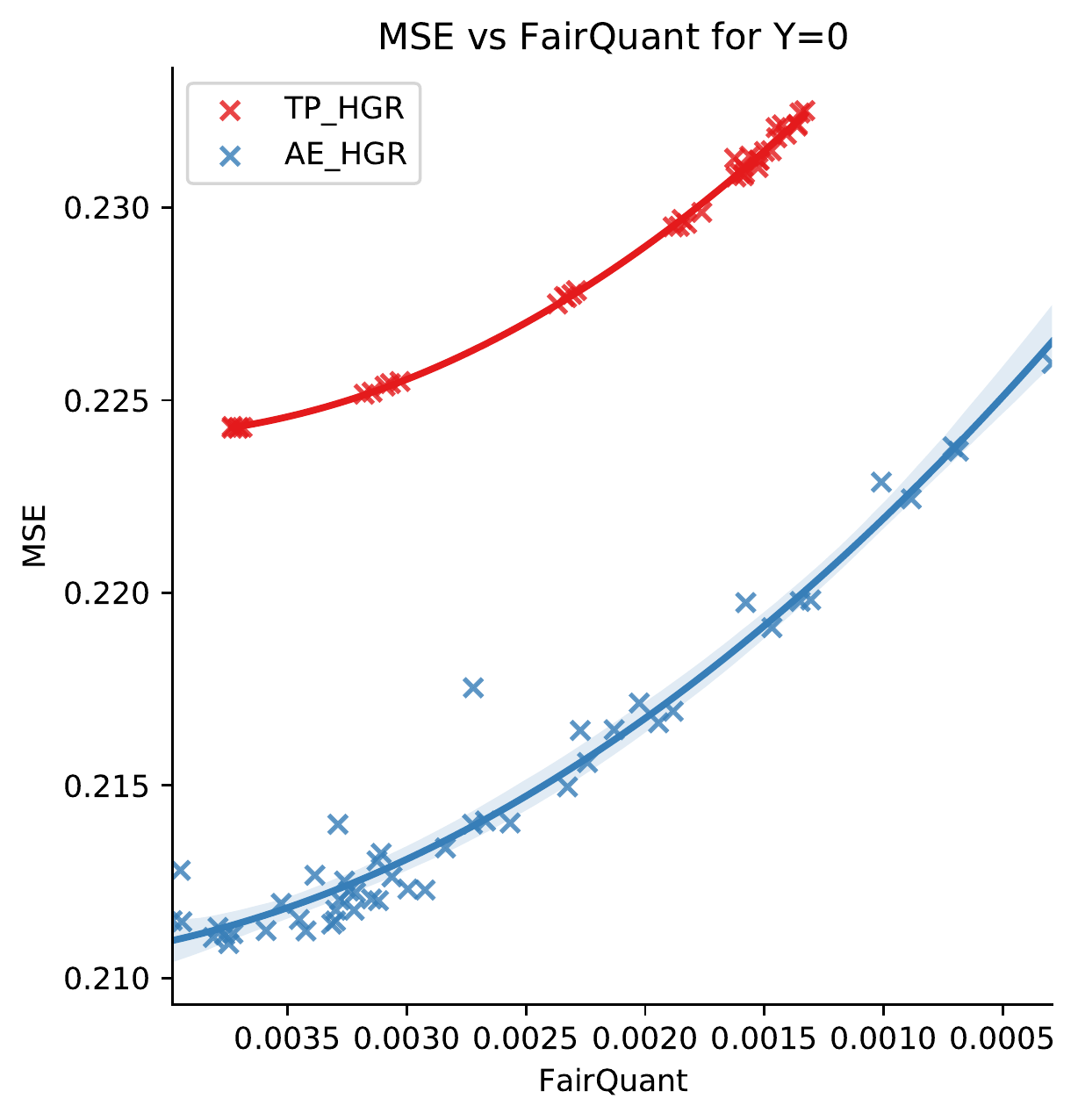}
\includegraphics[scale=0.35,valign=t]{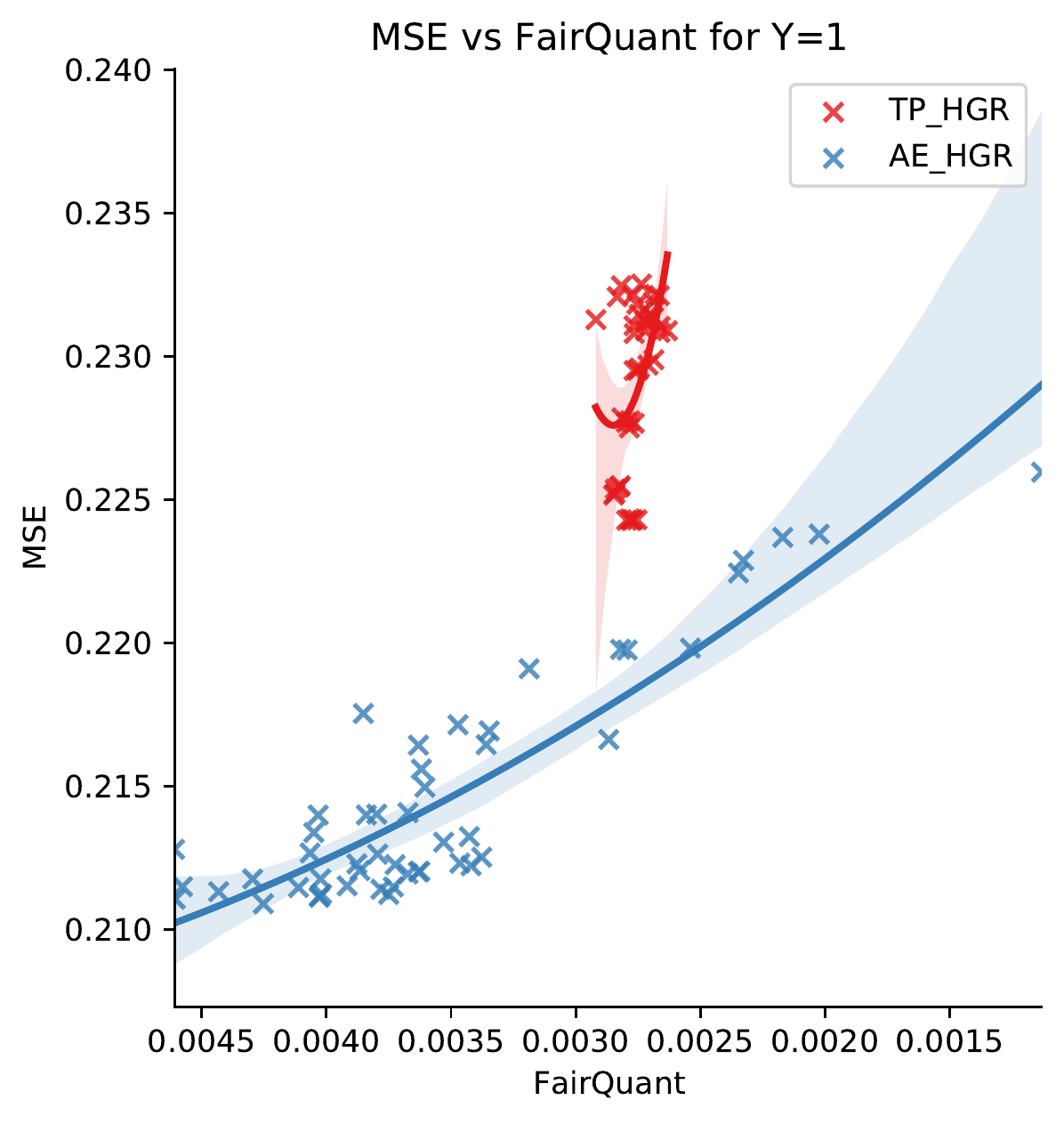} 
\includegraphics[scale=0.35,valign=t]{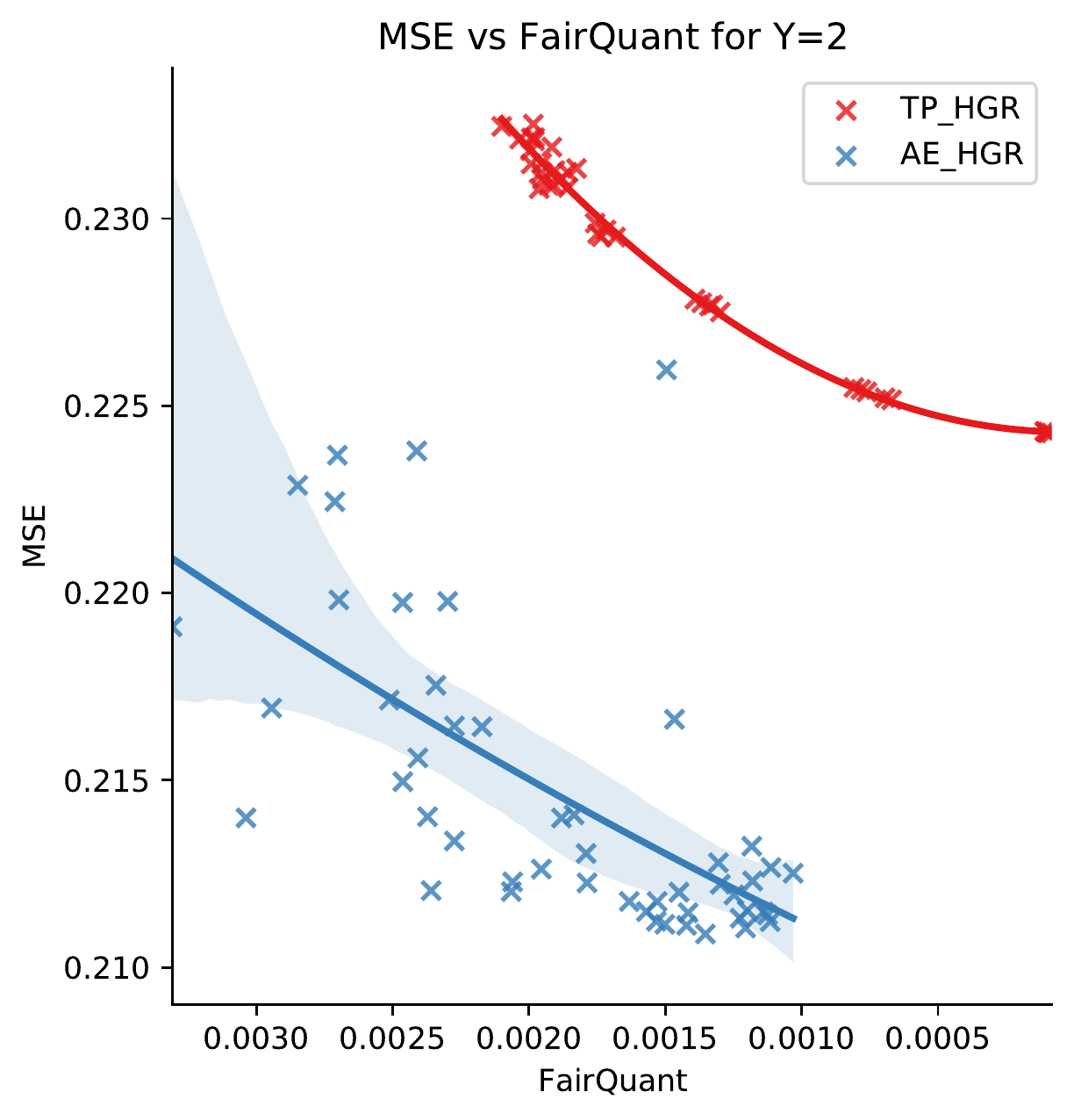} 
} \\
\subfloat[Equalized odds task with $\lambda_0=\lambda_1=\lambda_{2+}$]{
\includegraphics[scale=0.35,valign=t]{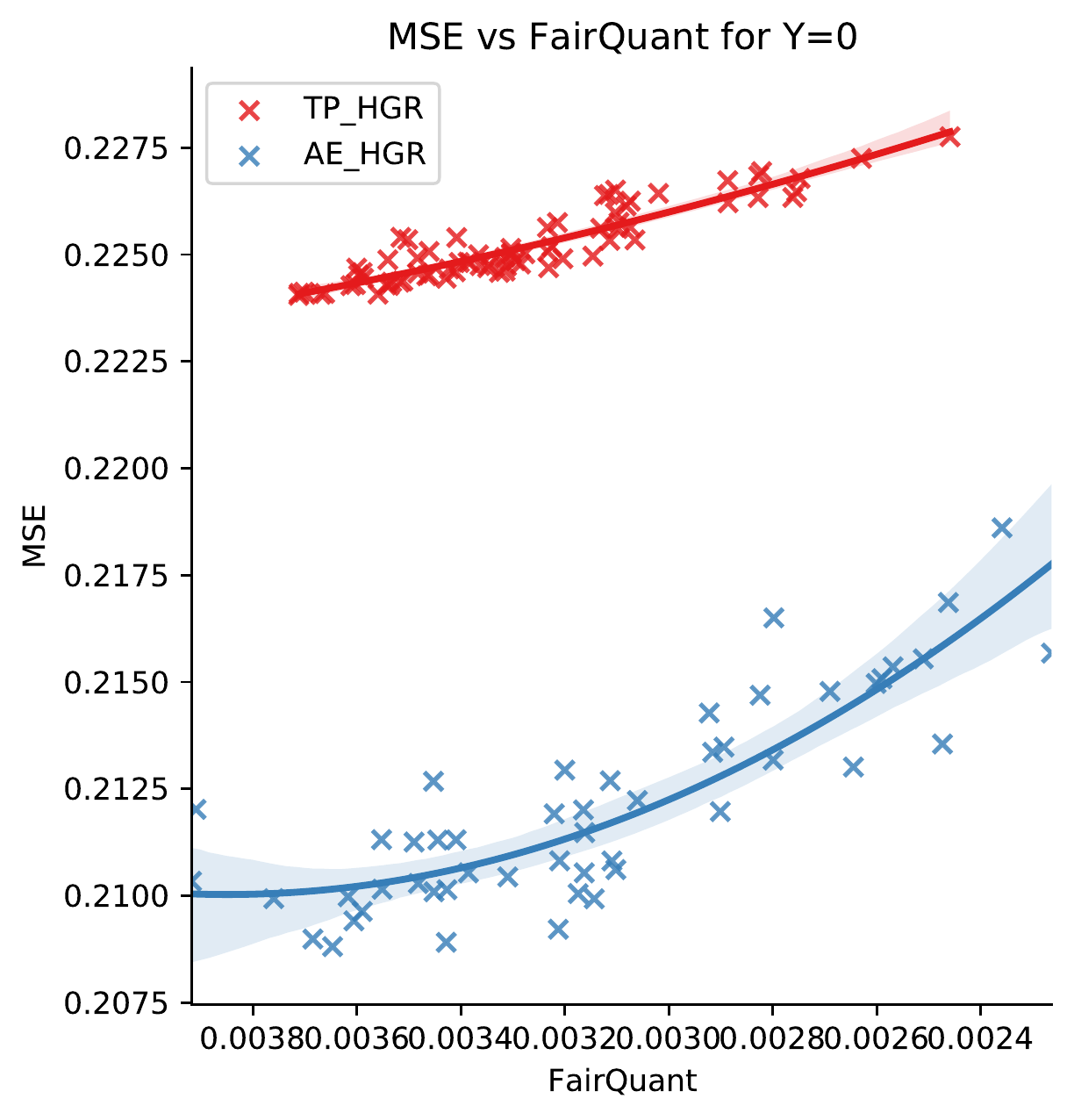} 
\includegraphics[scale=0.35,valign=t]{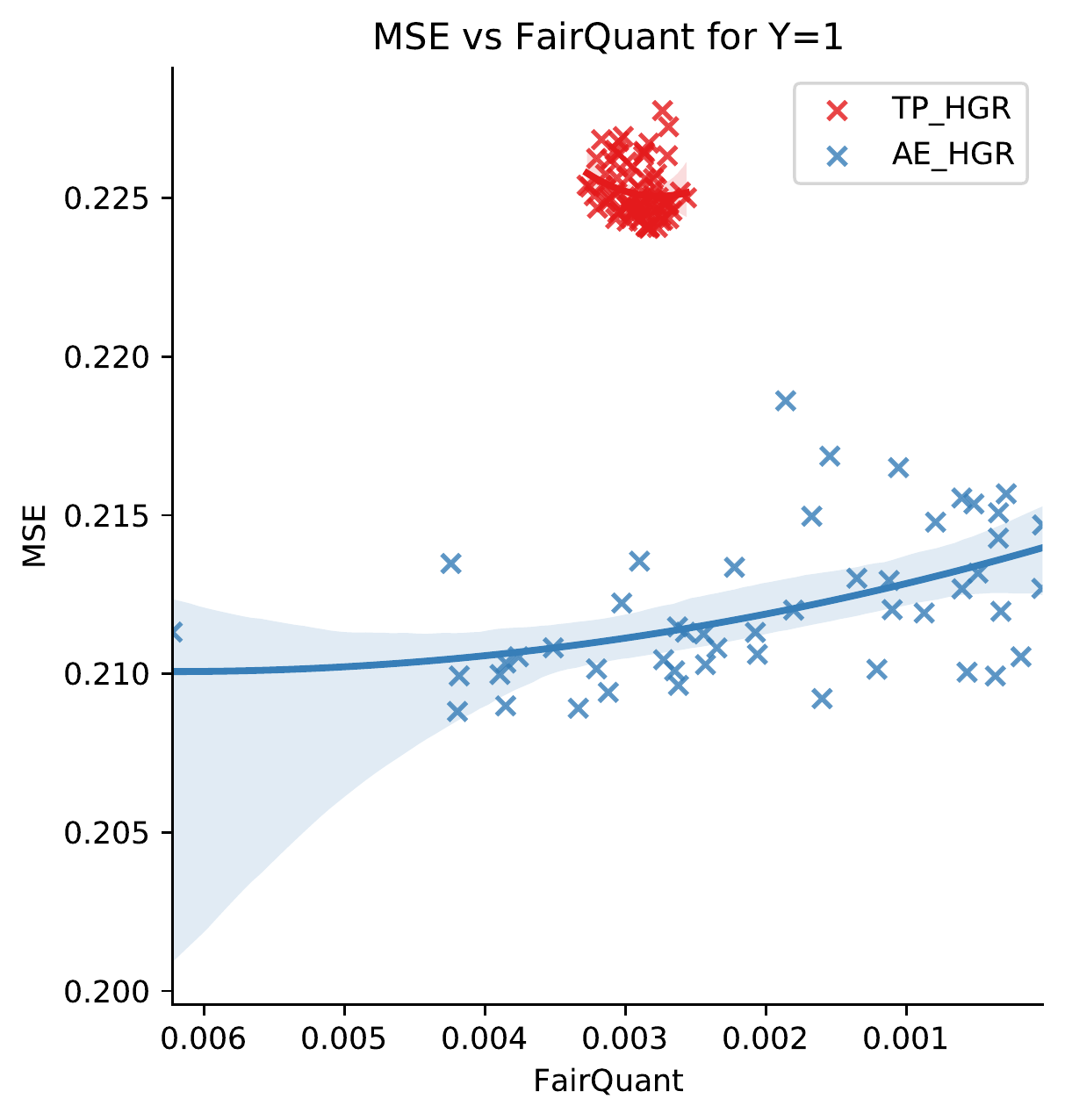} 
\includegraphics[scale=0.35,valign=t]{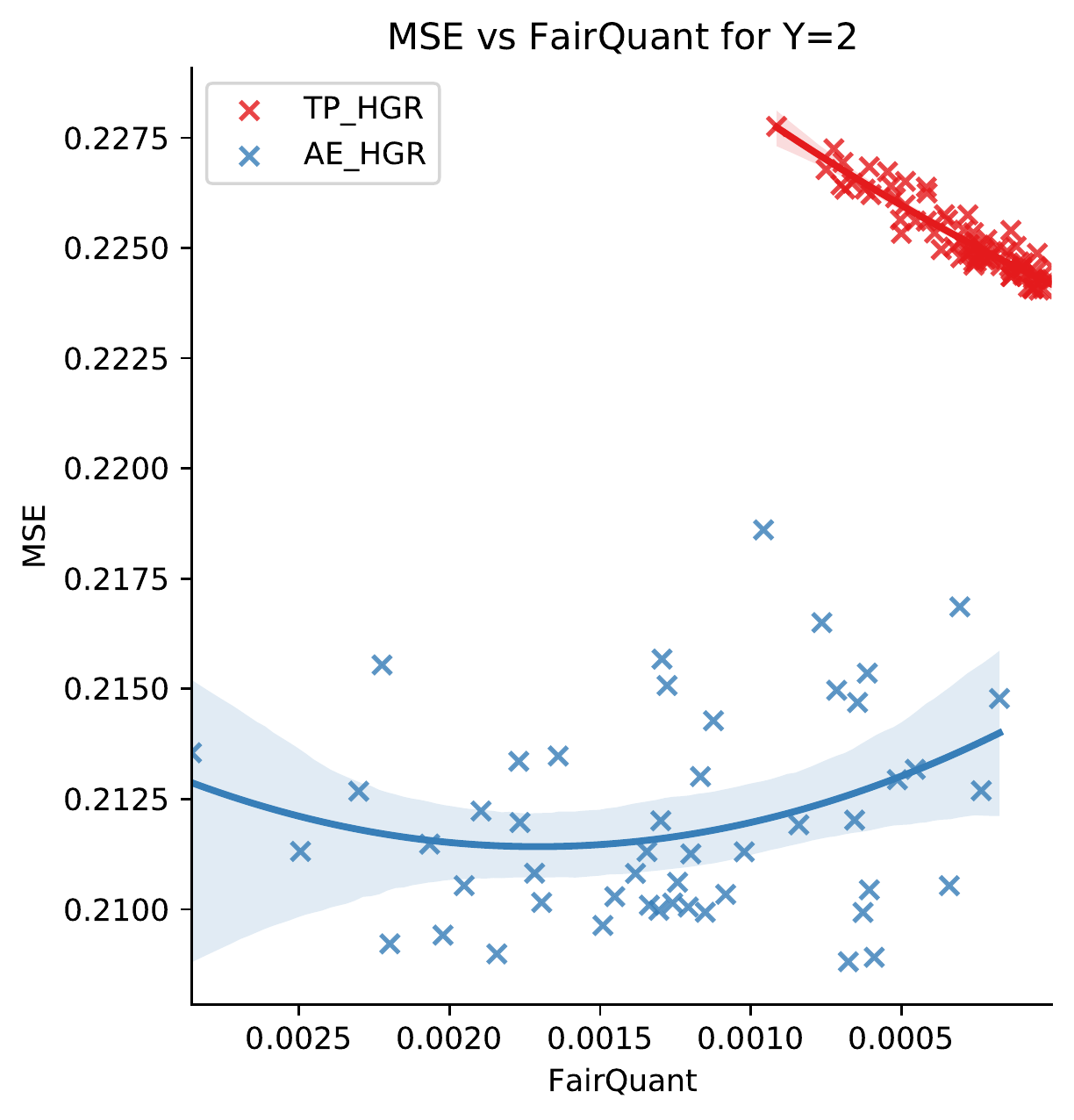}
} 
\caption{Frequency task (Pricingame 2015 dataset) - 
The figures show the results between the prediction performance (\emph{MSE}) against the fairness metric (\emph{FairQuant)} for the demographic parity task (a) and for the equalized odds task (b). (Left) represents the results only for cases where $Y=0$ ($87.7\%$ of the observations), (Middle) only for one-claim cases where $Y=1$ ($10.4\%$ of the observations) and (Right) for high-claim cases where $Y\geq2$ ($1.9\%$ of the observations). For all cases, our method outperforms the traditional one. We also observe that the demographic parity objective gives more weight to mitigation where there are more observations, especially for unclaimed ($Y=0$). While for observations with claims, it seems to be more discarded. In contrast, our proposed method for equalized odds objective seems more consistent even for cases with fewer observations. We note that for the $3$ cases, the fairness is very close to the optimum ($FairQuant$ very close to $0$).
}
\label{fig:S4_EO}
\end{figure}
Figure~\ref{fig:S4_EO} compares the results of (a) the demographic parity task and (b) the equalized odds task. We plot for this purpose the predictive performance (measured by \emph{MSE}) against the fairness metric (measured by \emph{FairQuant}) with separate mistreatment for each cases of the target $Y=\{0,1,2^{+}\}$. For all cases, our method outperforms the traditional one. We also observe that for the demographic parity task and in particular where $Y=0$, the bias mitigation is more pronounced ($FairQuant$ closer to $0$) than the other cases ($Y=\{1,2^{+}\}$). Most observations have no claims ($87.7\%$), therefore, the weight for the $0$ case should be higher. 
In contrast, our proposed equalized odds strategy by separating the cases for different lambdas shows an apparent improvement for the cases with claims ($Y=\{1,2^{+}\}$). The $FairQuant$ is closer to zero in these cases, except for the traditional one where the $FairQuant$ stabilizes at $0.003$ for $Y=1$. Note that this is not at the sacrifice of the non-claimed observations where the results are close to the demographic parity results.

\section{Continuous Task on Real-World Datasets}
\label{sec:res_cont}

In this section, we will discuss two continuous cases. First, we will evaluate the strength of the HGR non-linearity dependence and secondly we apply a multidimensional sensitive scenario. 

\subsection{Continuous Task on Binary Sensitive Attribute}

\begin{figure}[ht!]
\subfloat[Pricingame 2015 Dataset]{
\includegraphics[scale=0.30,valign=t]{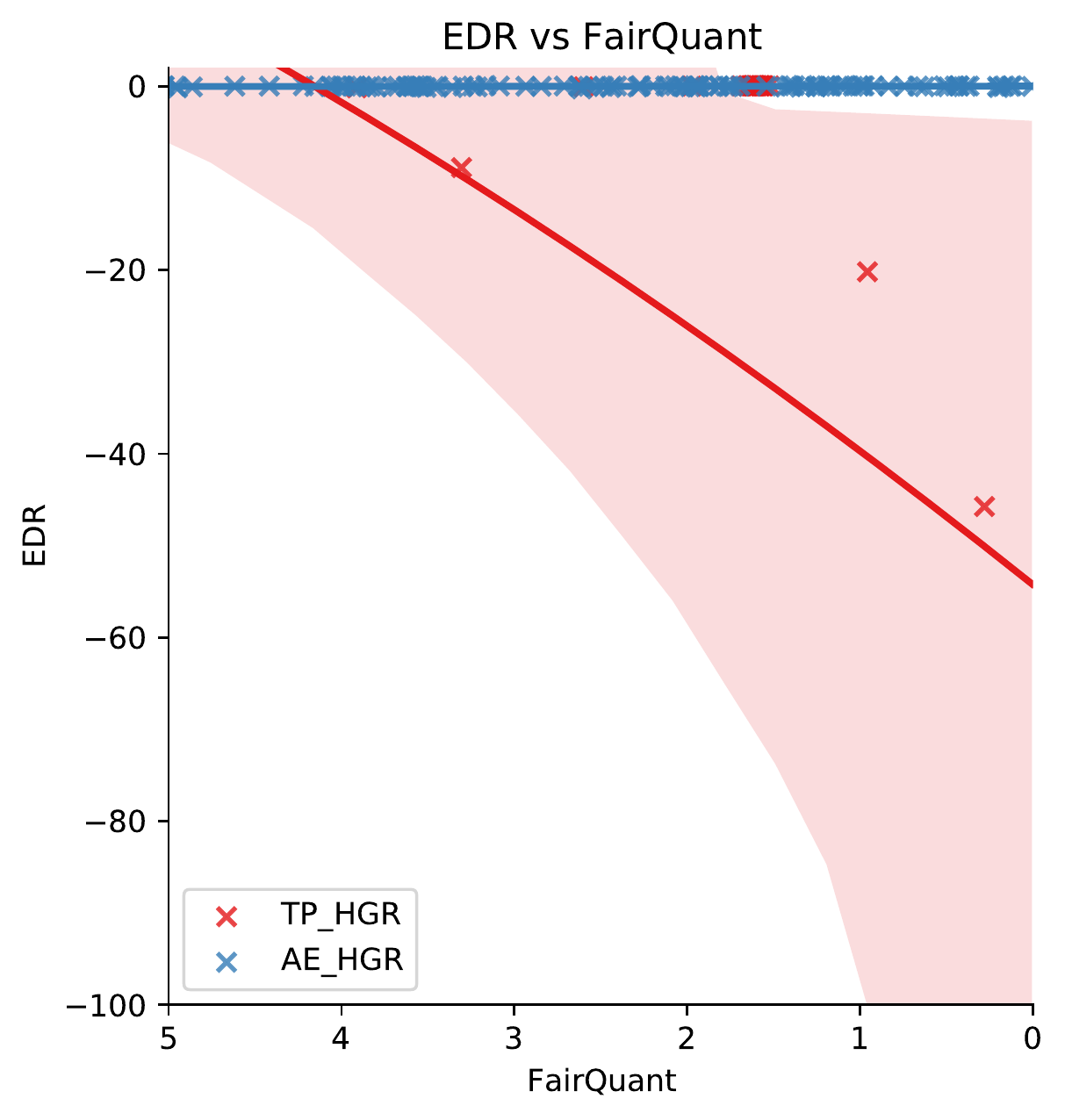}\\
\includegraphics[scale=0.30,valign=t]{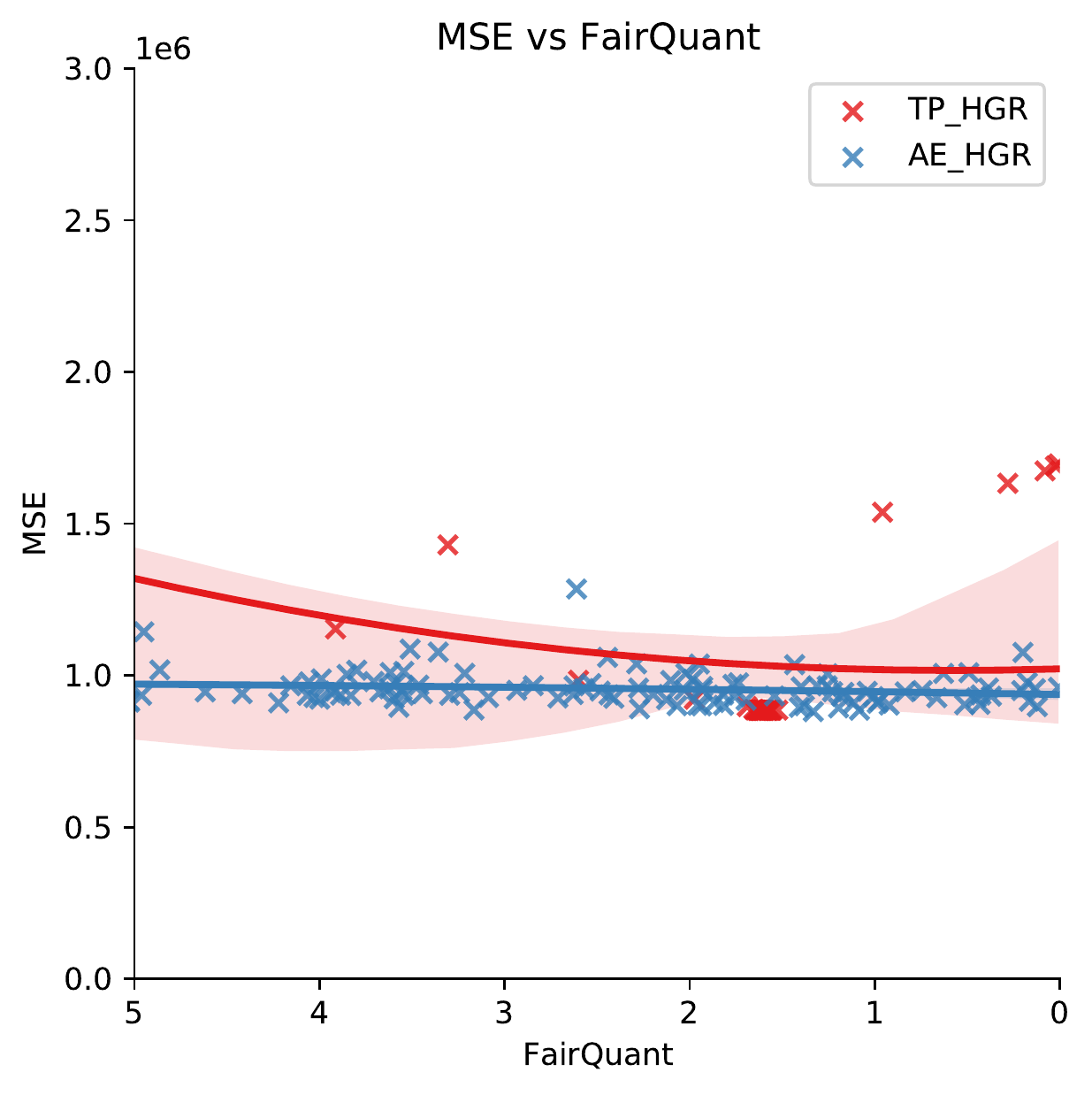} \\
\includegraphics[scale=0.30,valign=t]{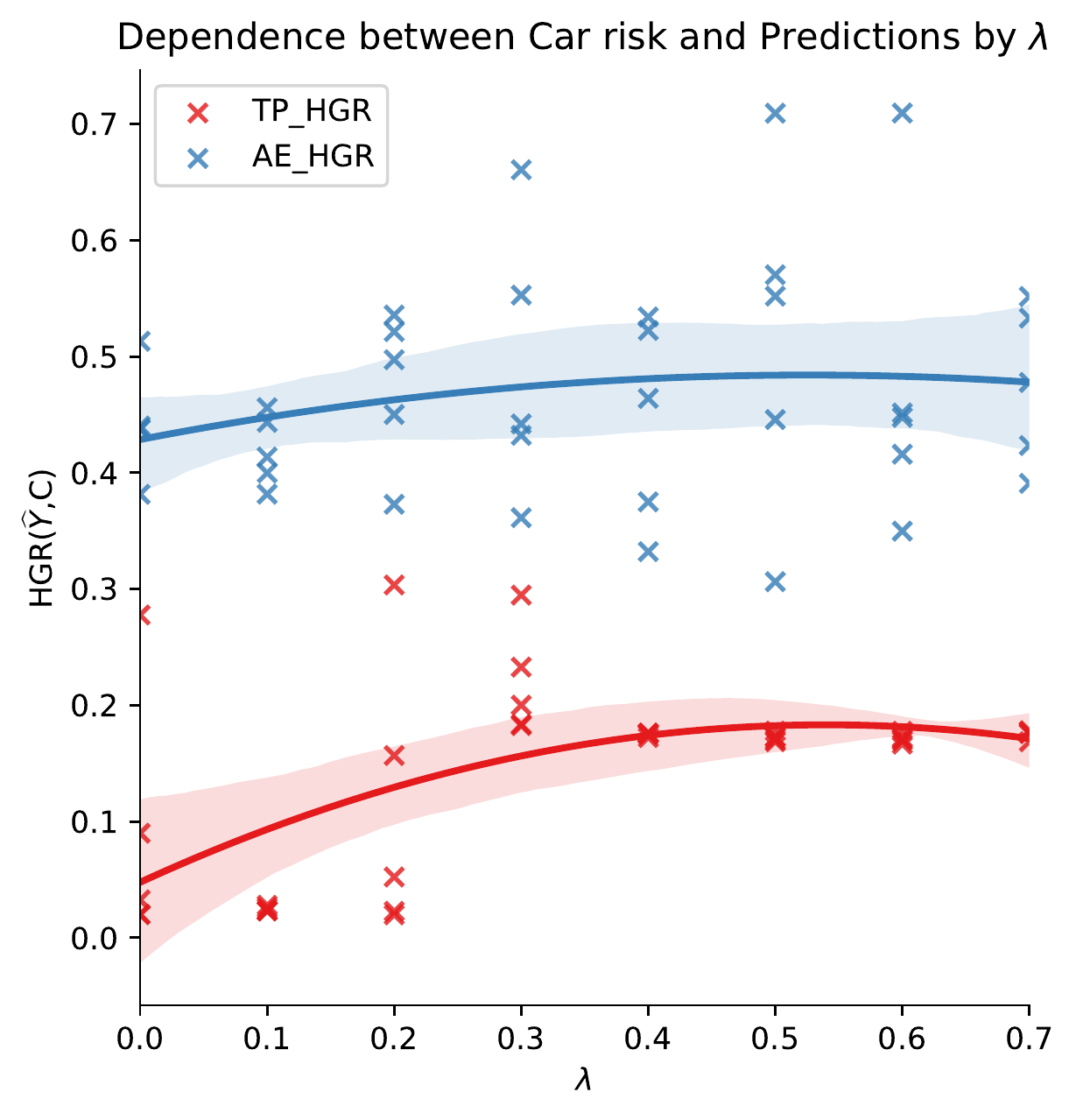} \\
\includegraphics[scale=0.30,valign=t]{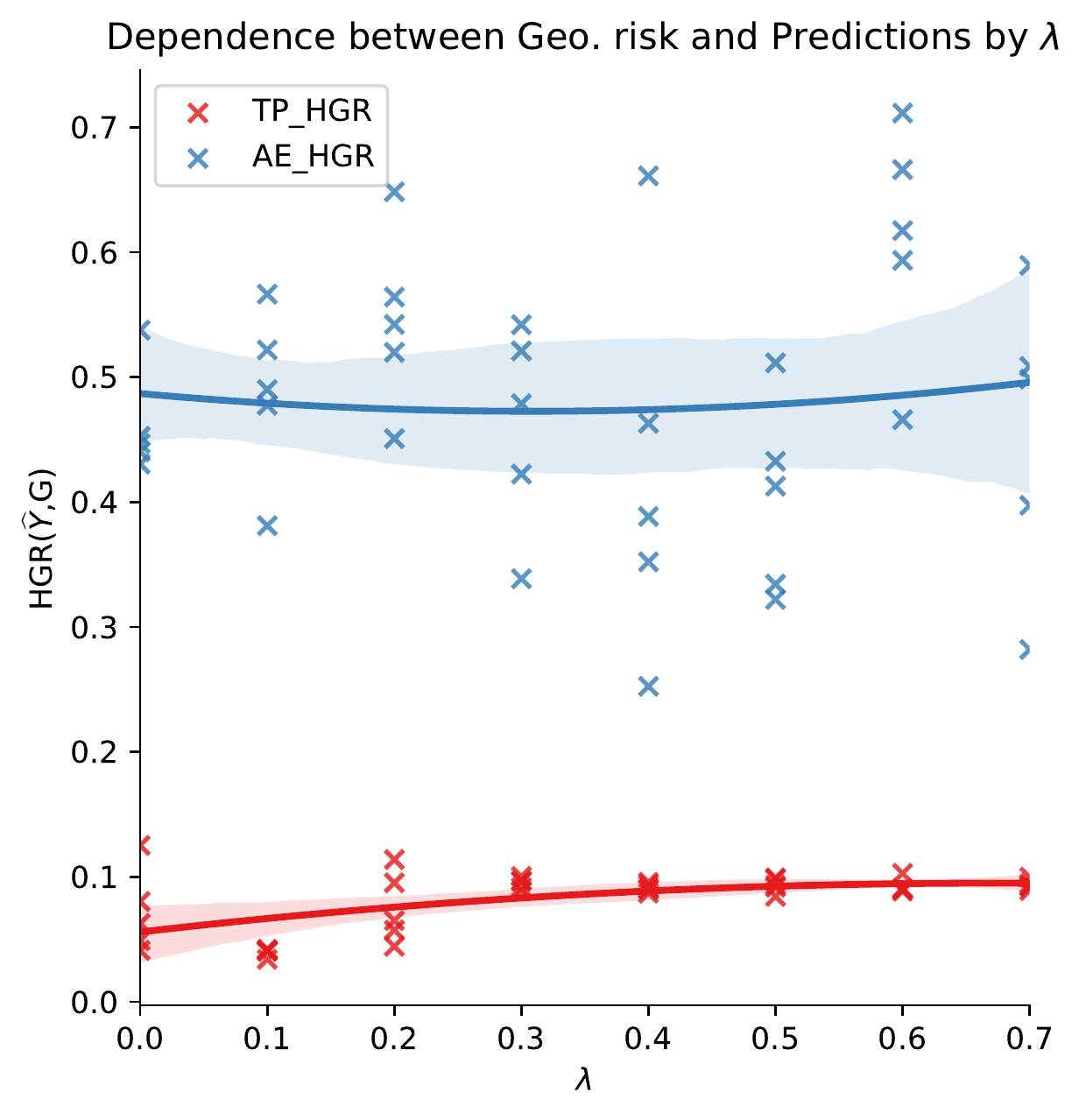}
} \\
\subfloat[Pricingame 2017 Dataset]{
\includegraphics[scale=0.30,valign=t]{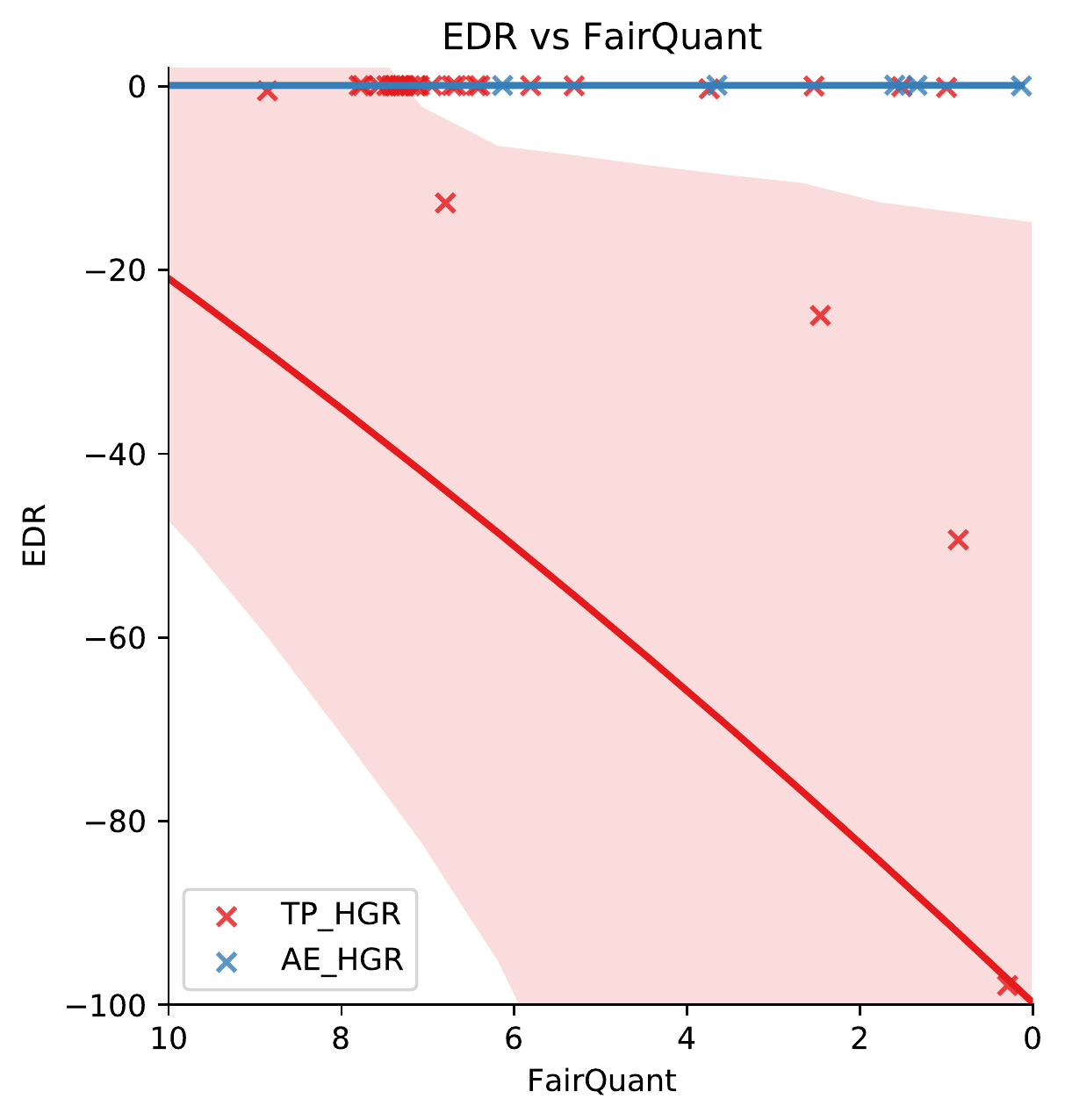} \\
\includegraphics[scale=0.30,valign=t]{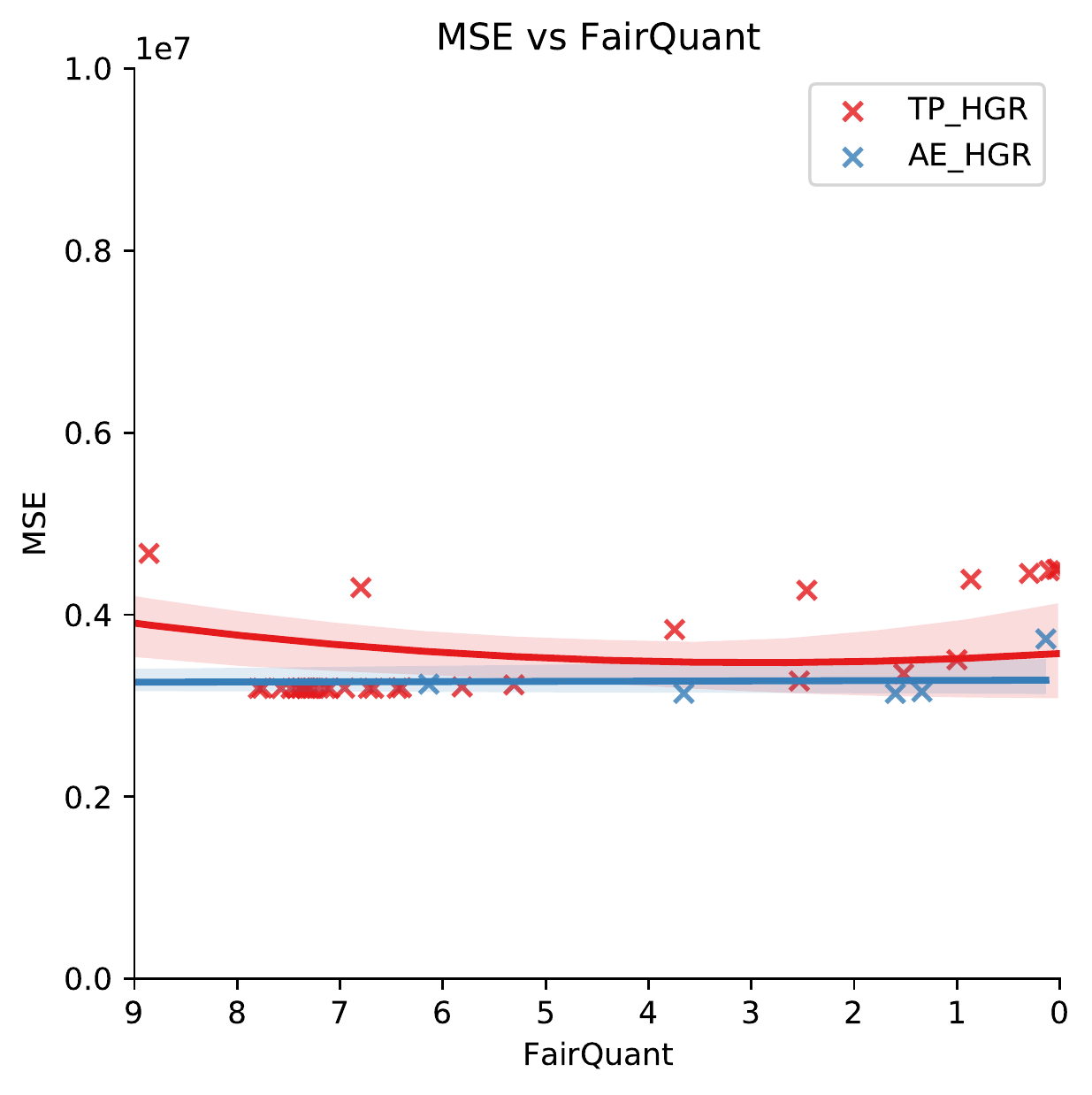} \\
\includegraphics[scale=0.30,valign=t]{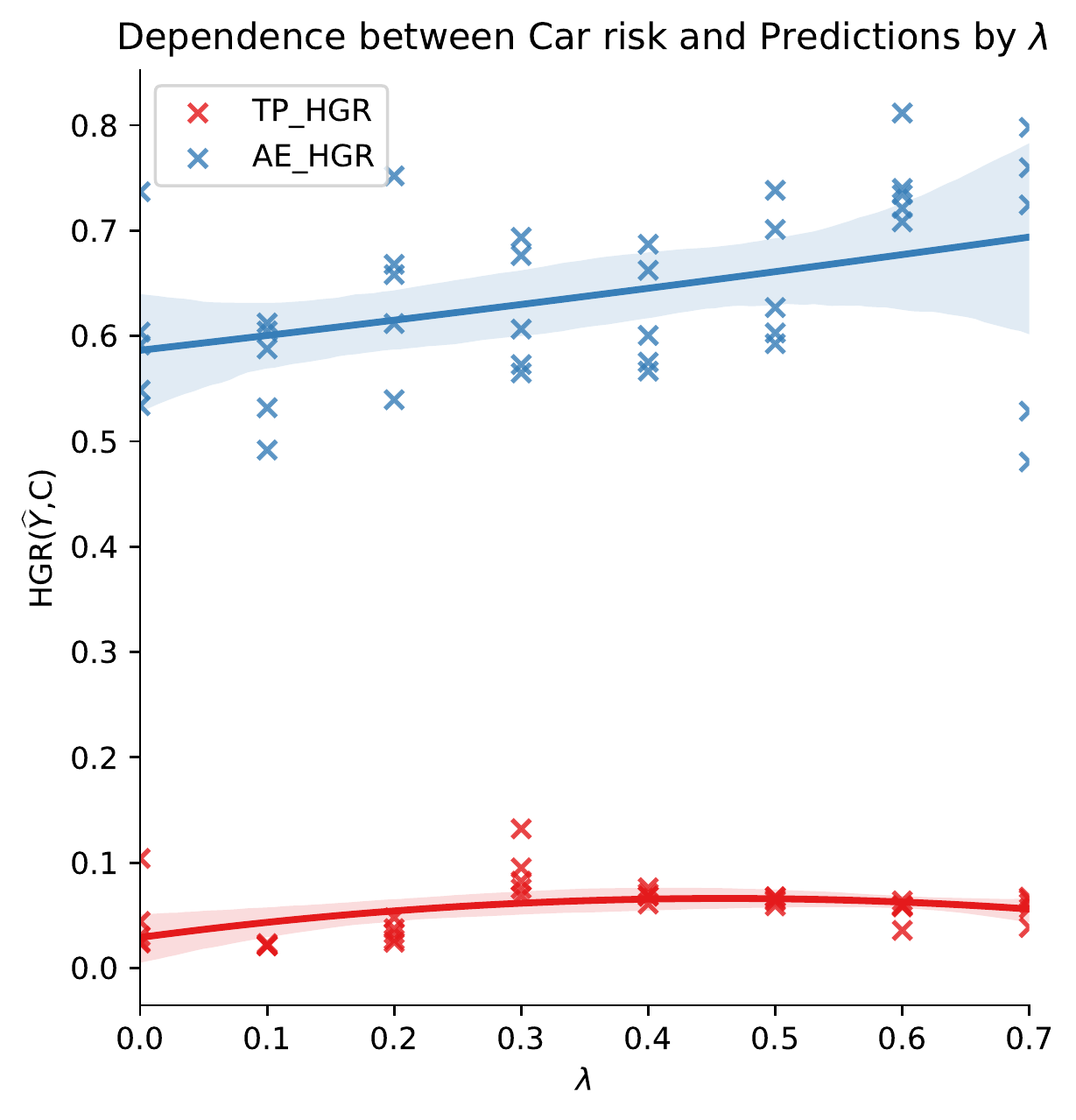} \\ \includegraphics[scale=0.30,valign=t]{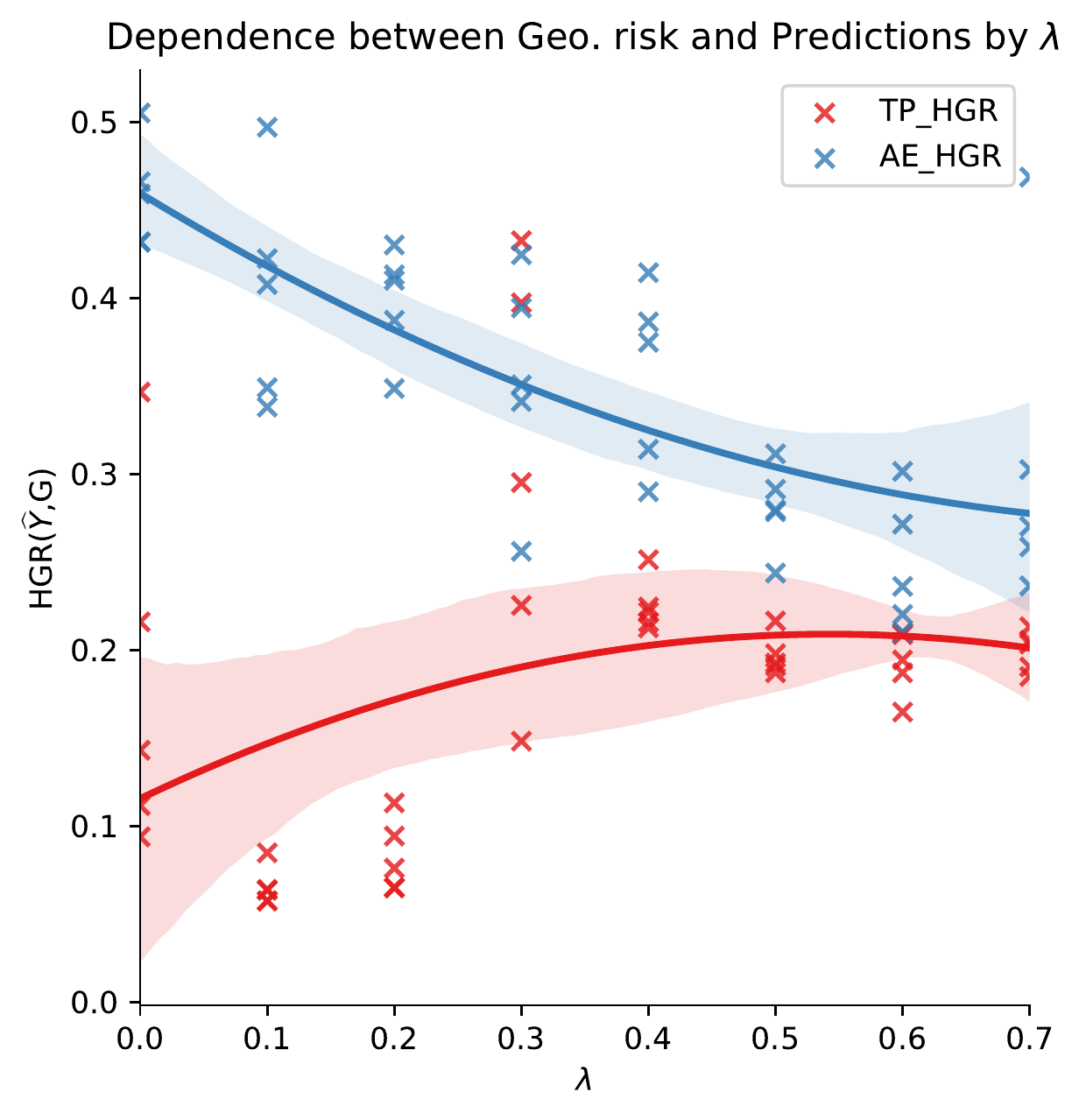}
}
\caption{Scenario 3 - Continuous Objective (Average Cost) 
(Left and Middle left) For all levels of predictive performance, \emph{EDR} or \emph{MSE}, our proposed method outperforms the traditional fair pricing model. (Middle right and Right) Higher $\lambda$ values significantly reduce the dependence between output predictions and the car risk, as well as for the geographical risk in comparison to our proposed model.
}
\label{fig:S3_CO}
\end{figure}

We propose on the continuous task to predict the total cost of Third-Party Material claims (euro) from 
\emph{Pricingame 2017} dataset. 
Depending on the objective, we report the average of 
the Expected Deviance Ratio (EDR), the average of the Mean Squared Error ($MSE$), 
the mean of the $HGR$ metric $HGR_{NN}$ \cite{grari2019fairness} and the $FairQuant$ metric \cite{grari2019fairness} 
on the test set. 
We plot the performance of these different approaches by displaying the predictive performance (
$EDR$, $MSE$) 
against the $FairQuant$ for Demographic Parity in the two left most graphs in Figure~\ref{fig:S3_CO}. 
We note that, for all levels of fairness (controlled by the mitigation weight $\lambda$ in every approach), our method outperforms the traditional algorithm. In the left most graph, we observe that that the TP\_HGR algorithm, represented with the red curve, obtains degenerated results in mitigating biases. The \emph{EDR} has decreased in negative values with more than $-40$ for the most important $lambda$ where the FairQuant in close to $0$. Our approach seems to be more stable and allows to keep values above or close to $0$.  As in the frequency task, the autoencoder prediction dependencies with the car risk and the geographical risk are both more
important than the traditional one, and this is valid for all levels of fairness.

\noindent

\noindent

\newpage

\subsection{Continuous Task on Continuous Sensitive Attribute (Spatial)}

\begin{figure}[H]
  \centering
  \subfloat[$\lambda$=0]{
  \includegraphics[scale=0.11,valign=t]{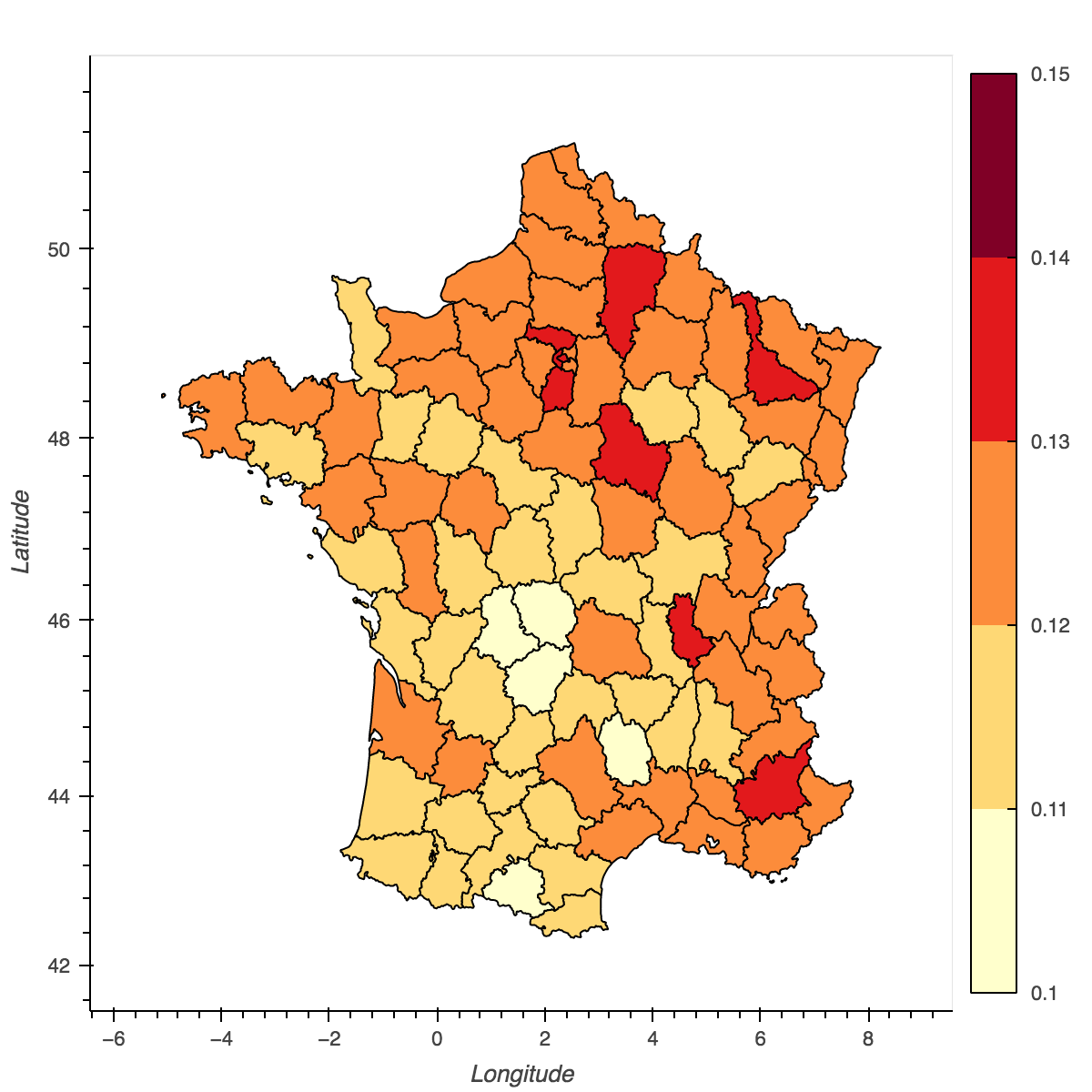}}
 \subfloat[$\lambda$=0.4]{
 \includegraphics[scale=0.11,valign=t]{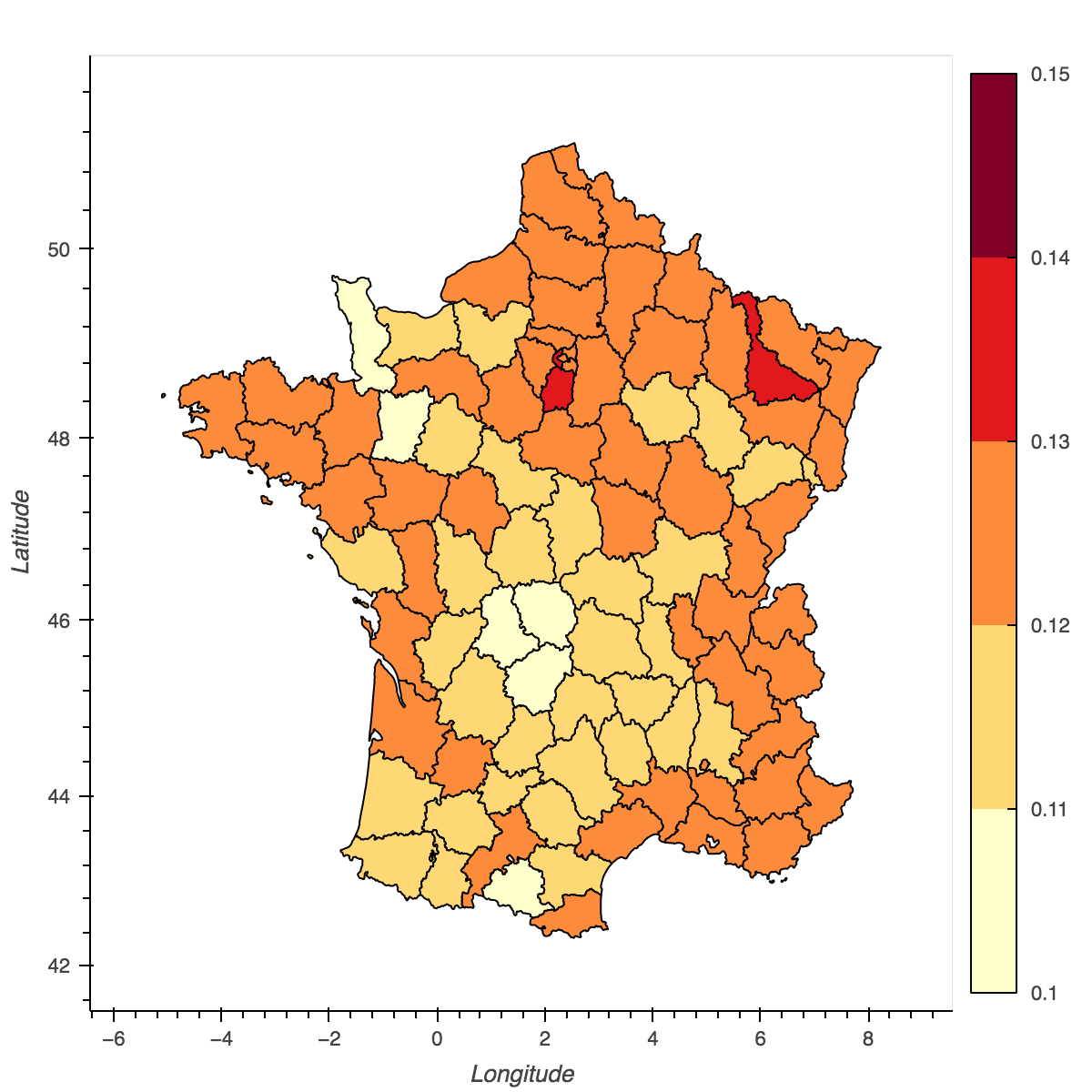}}
\subfloat[$\lambda$=0.6]{
 \includegraphics[scale=0.11,valign=t]{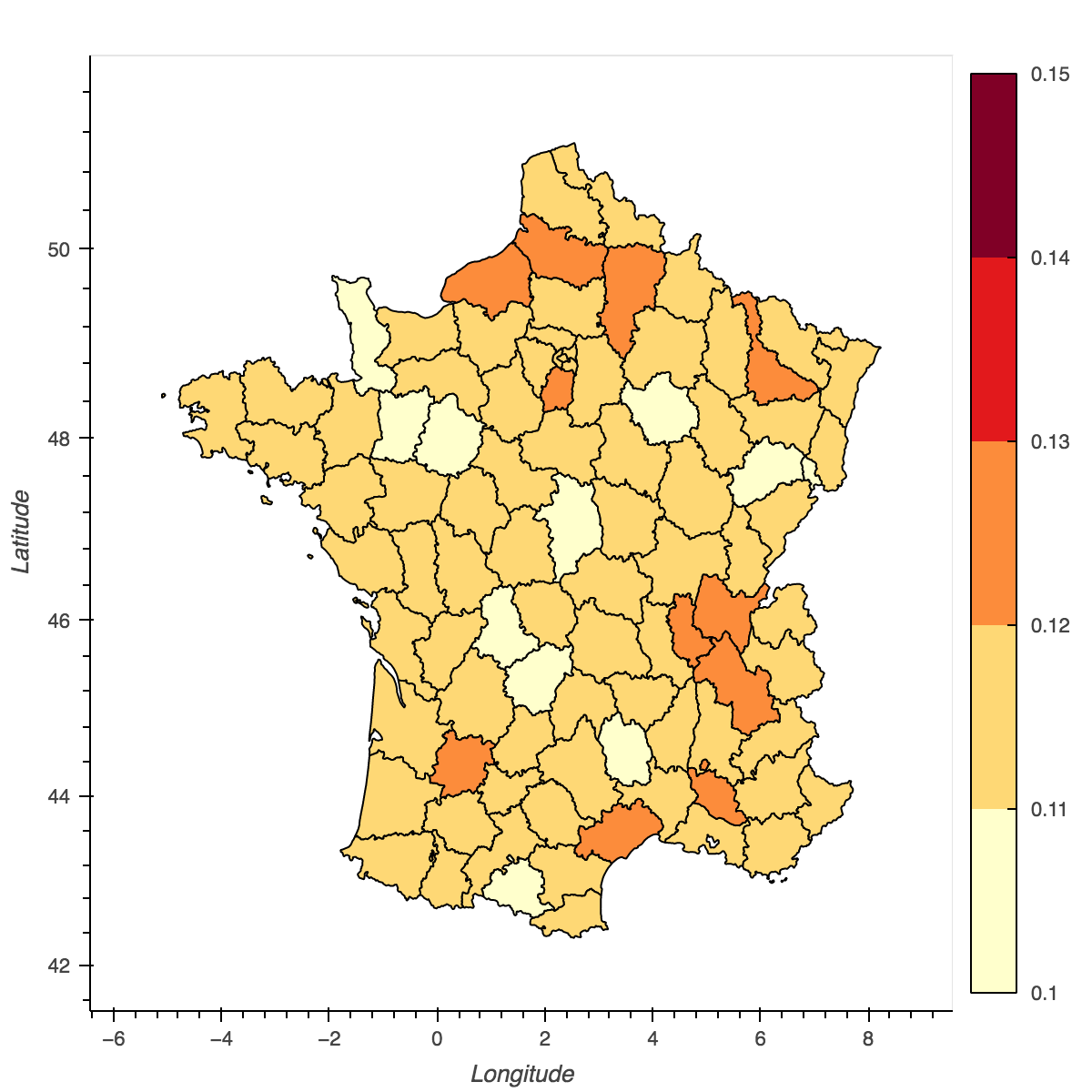}}
 \caption{Mitigating Geographical Biases - Frequency Objective. Higher $\lambda$ values allows to smooth the output prediction along the spatial information.
 \label{fig:S3_GEO}}
\end{figure}

\begin{figure}[H]
  \centering
  \subfloat[Department 44]{
  \includegraphics[scale=0.10,valign=t]{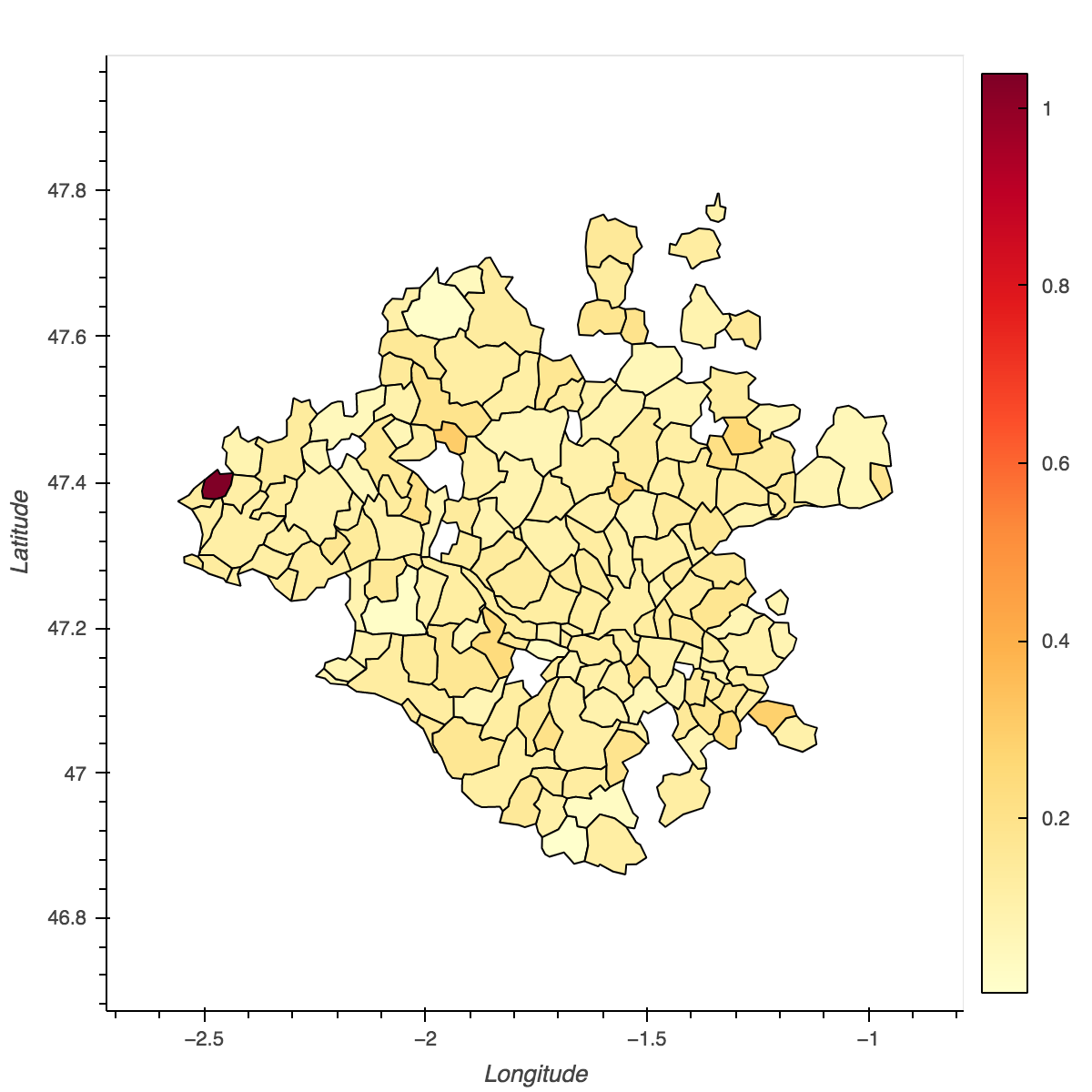}
 \includegraphics[scale=0.10,valign=t]{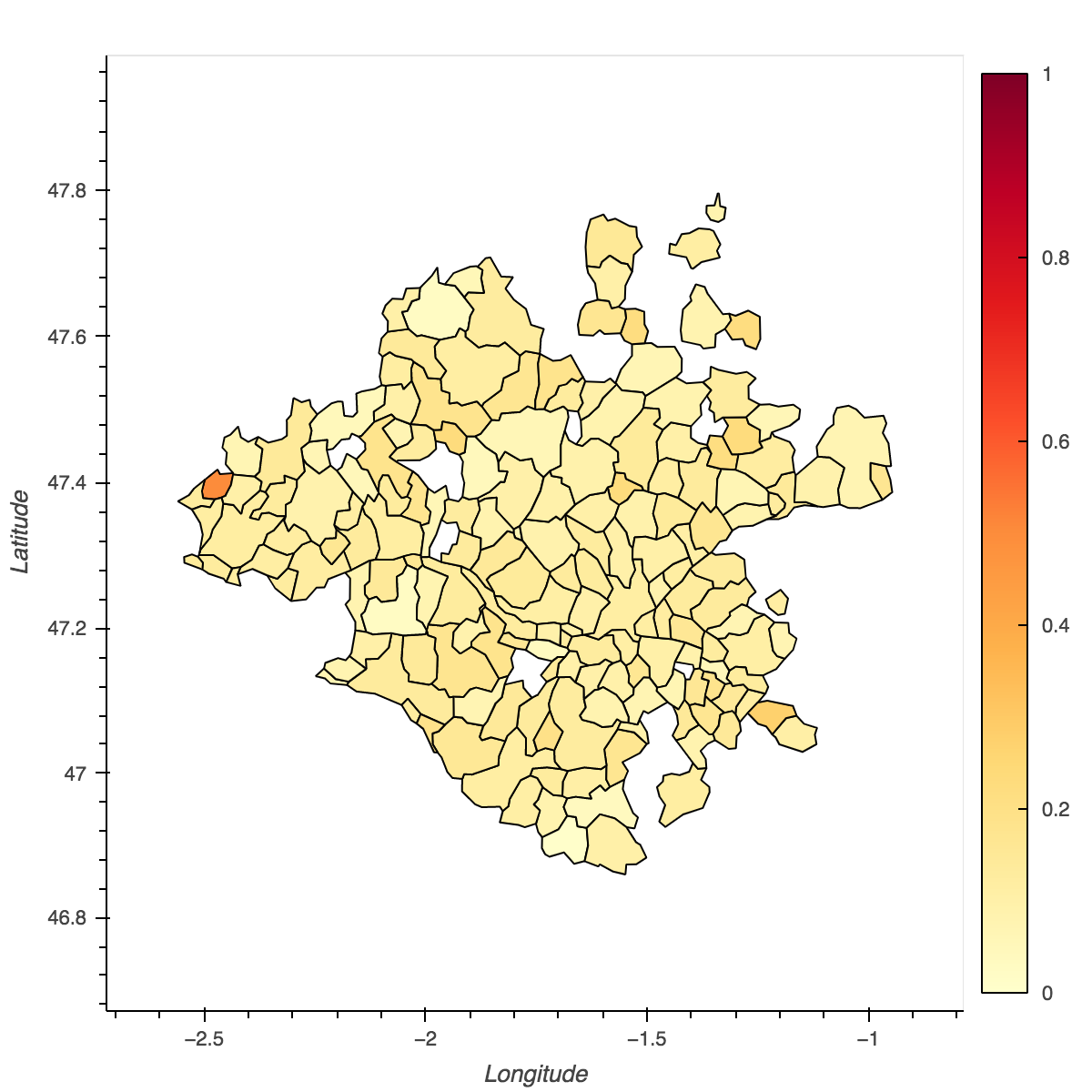}}
\subfloat[Department 65]{
\includegraphics[scale=0.10,valign=t]{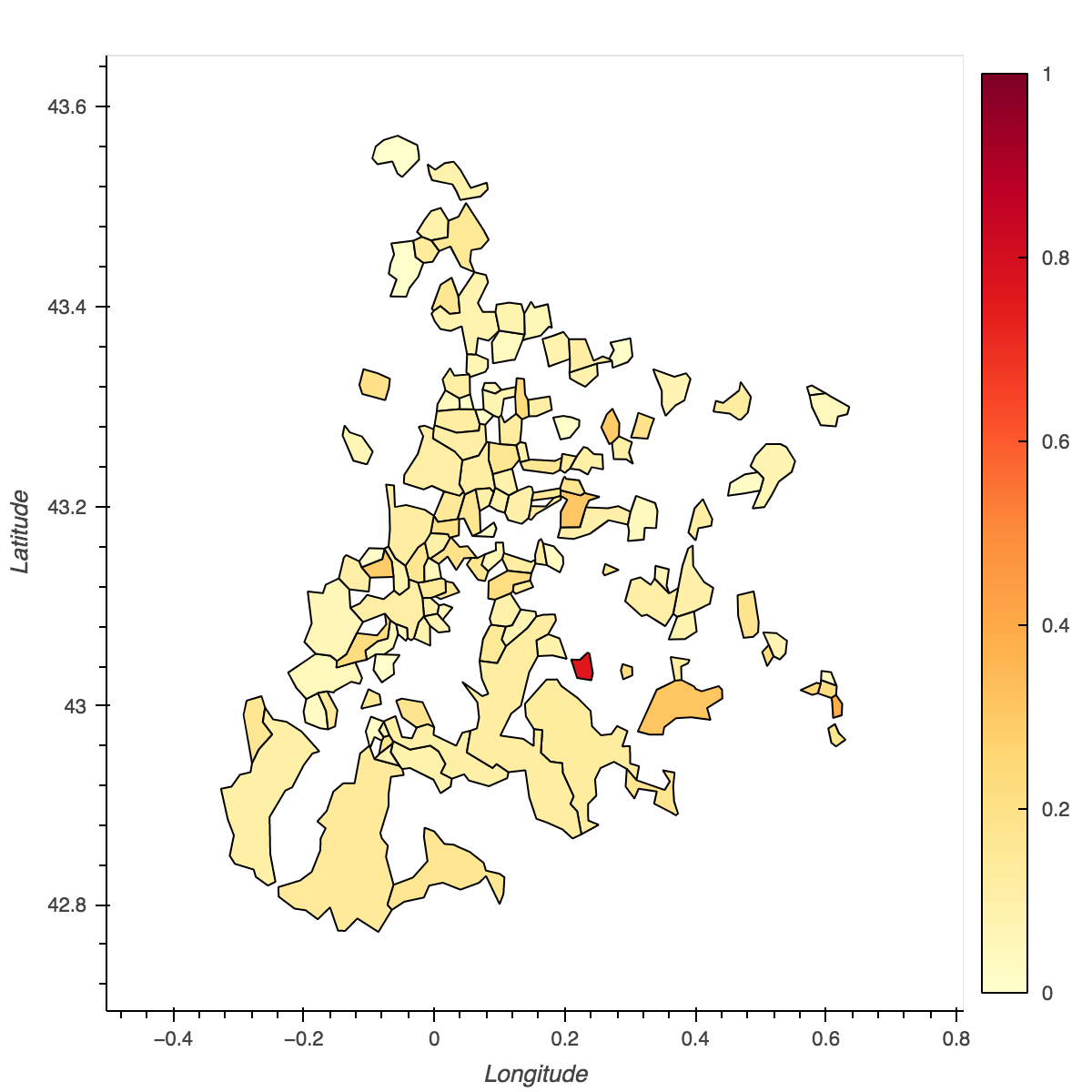}
\includegraphics[scale=0.10,valign=t]{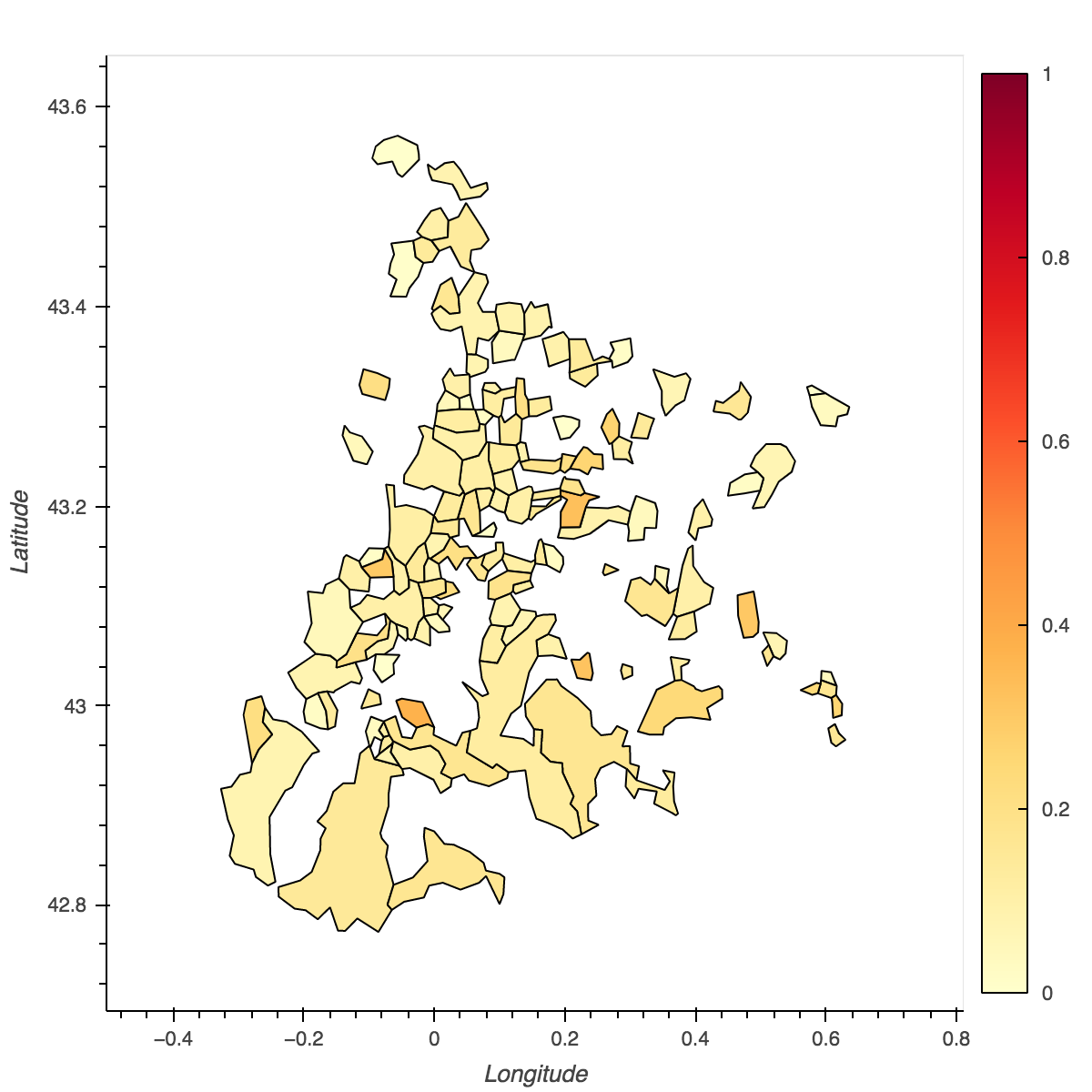}}
 \\
\subfloat[Department 67]{
\includegraphics[scale=0.10,valign=t]{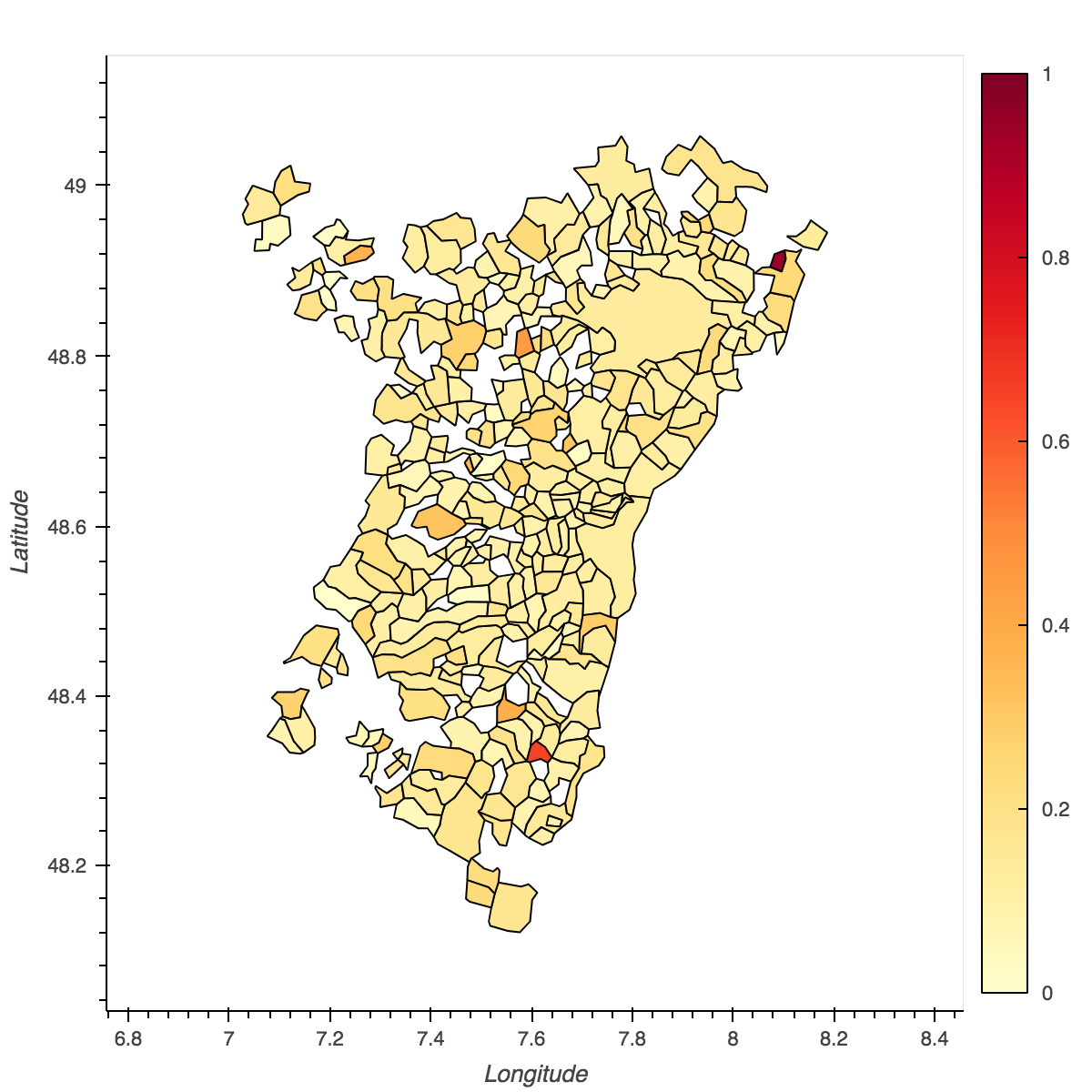}
\includegraphics[scale=0.10,valign=t]{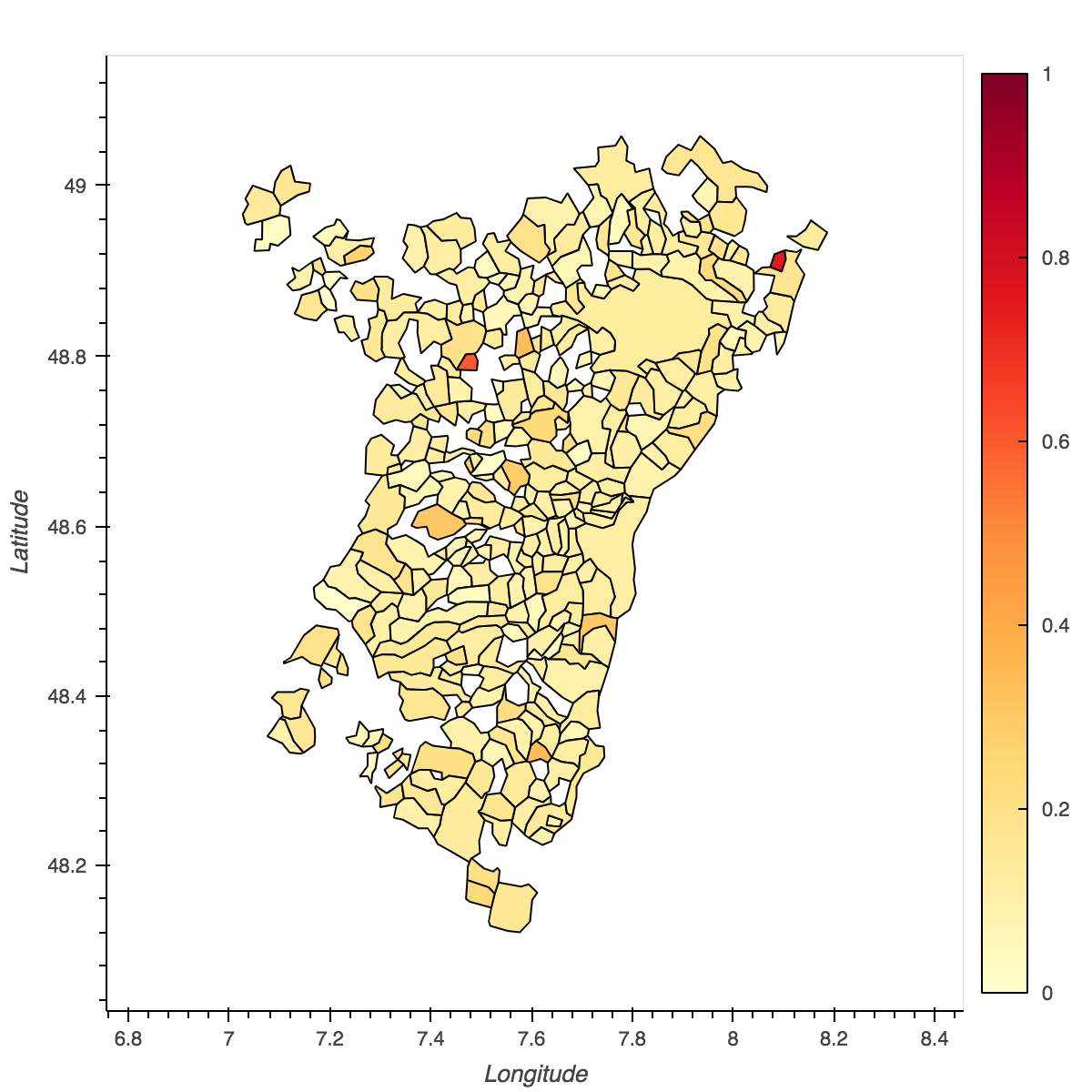}}
\subfloat[Department 91]{
\includegraphics[scale=0.10,valign=t]{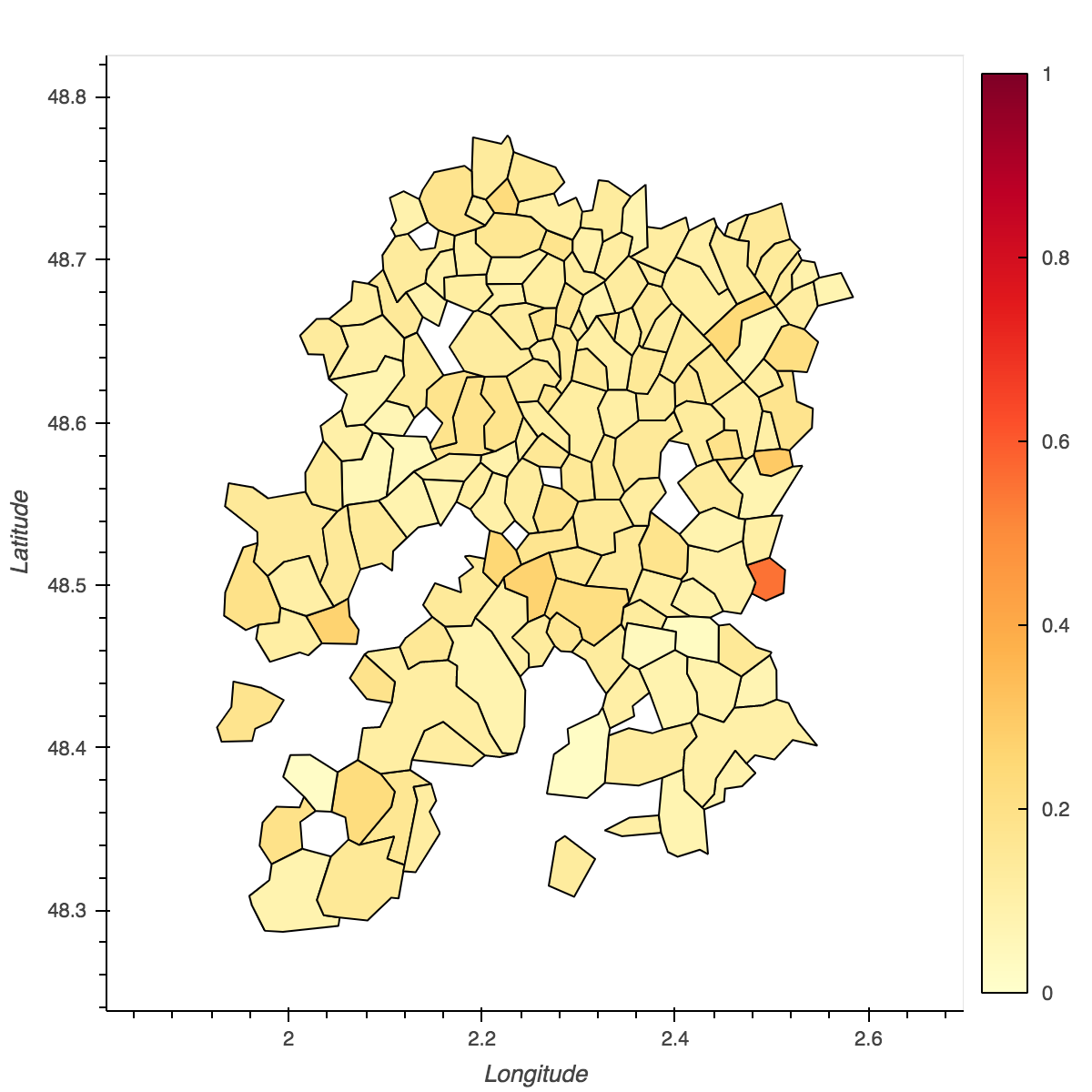}
\includegraphics[scale=0.10,valign=t]{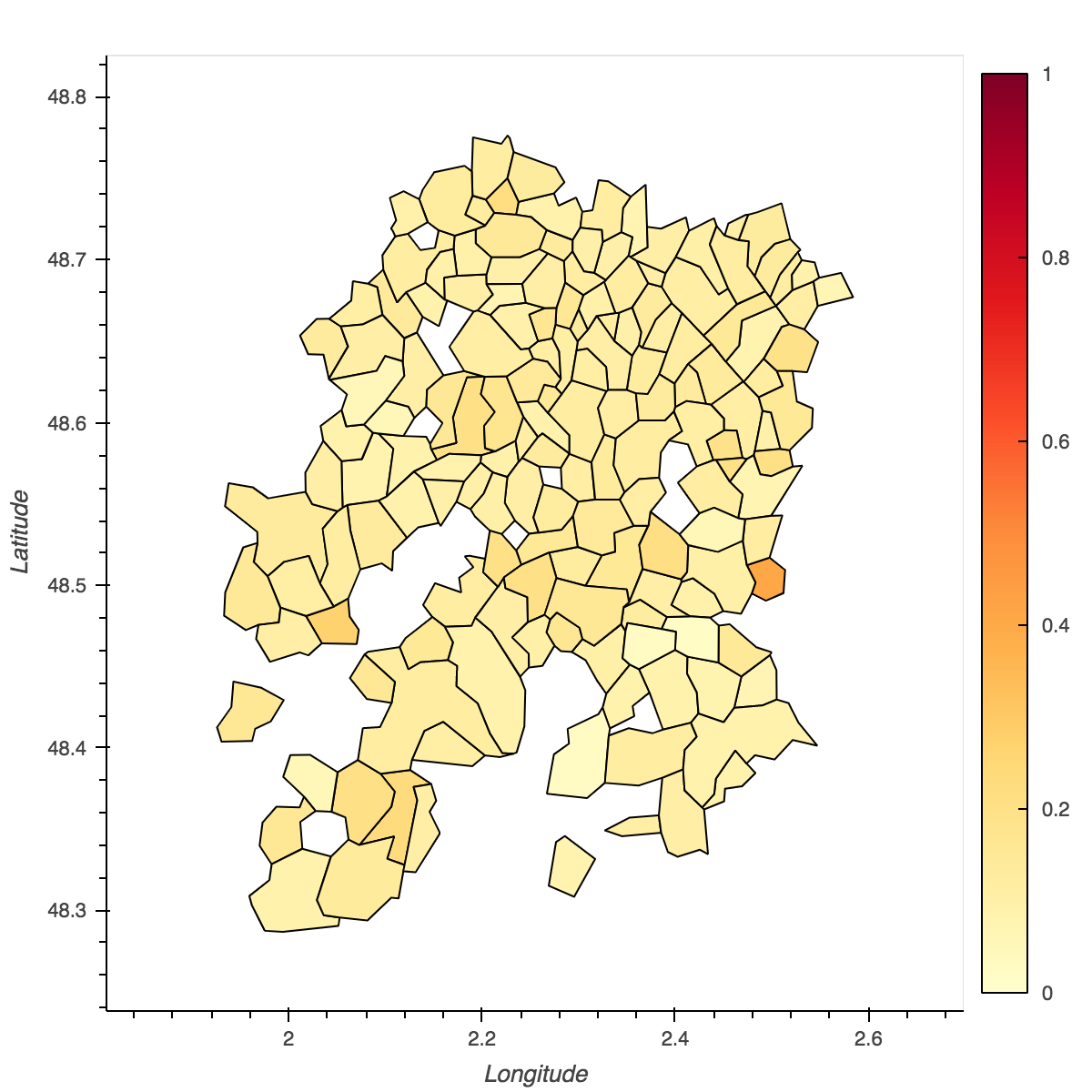}}
   \\ 
\subfloat[Department 69]{
\includegraphics[scale=0.10,valign=t]{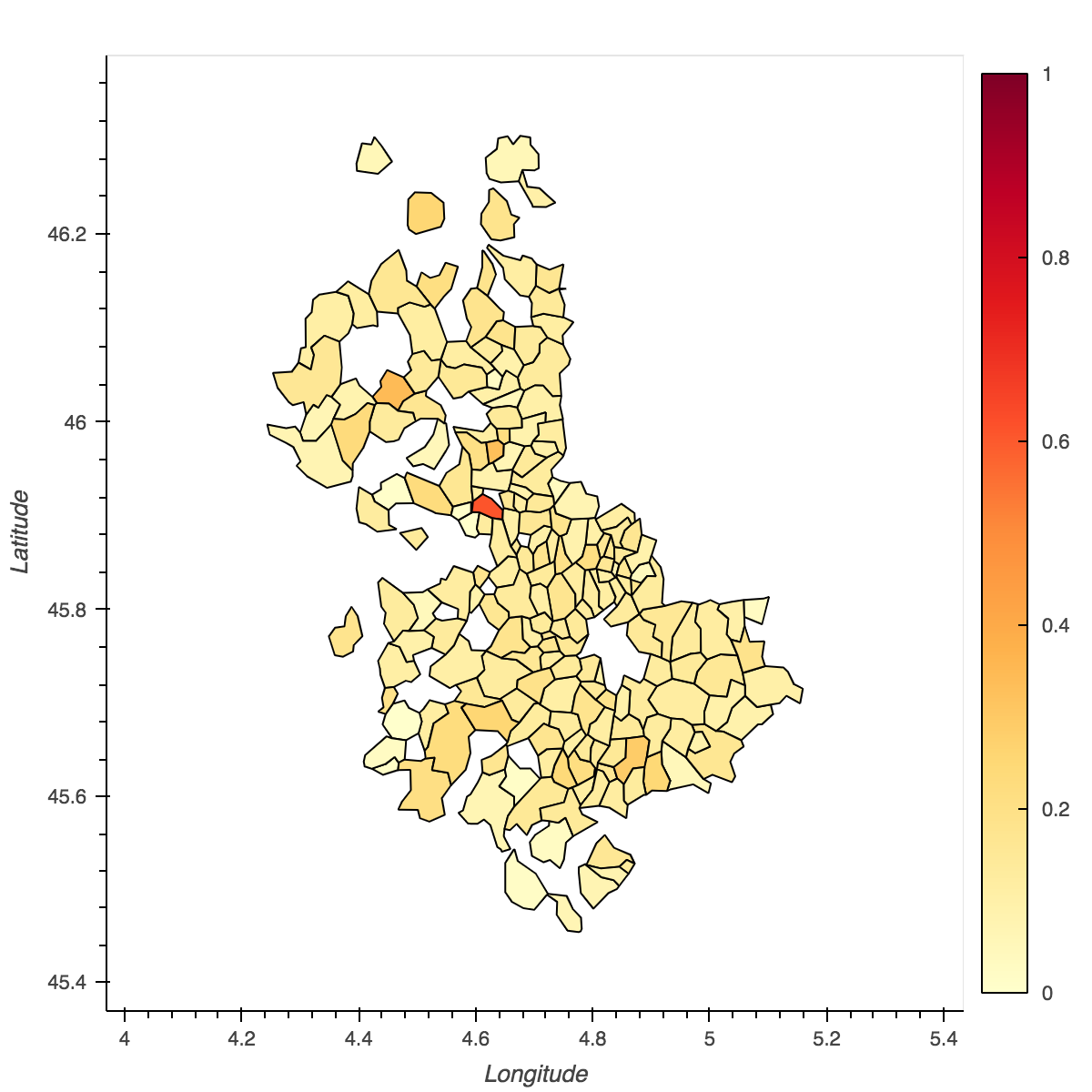}
\includegraphics[scale=0.10,valign=t]{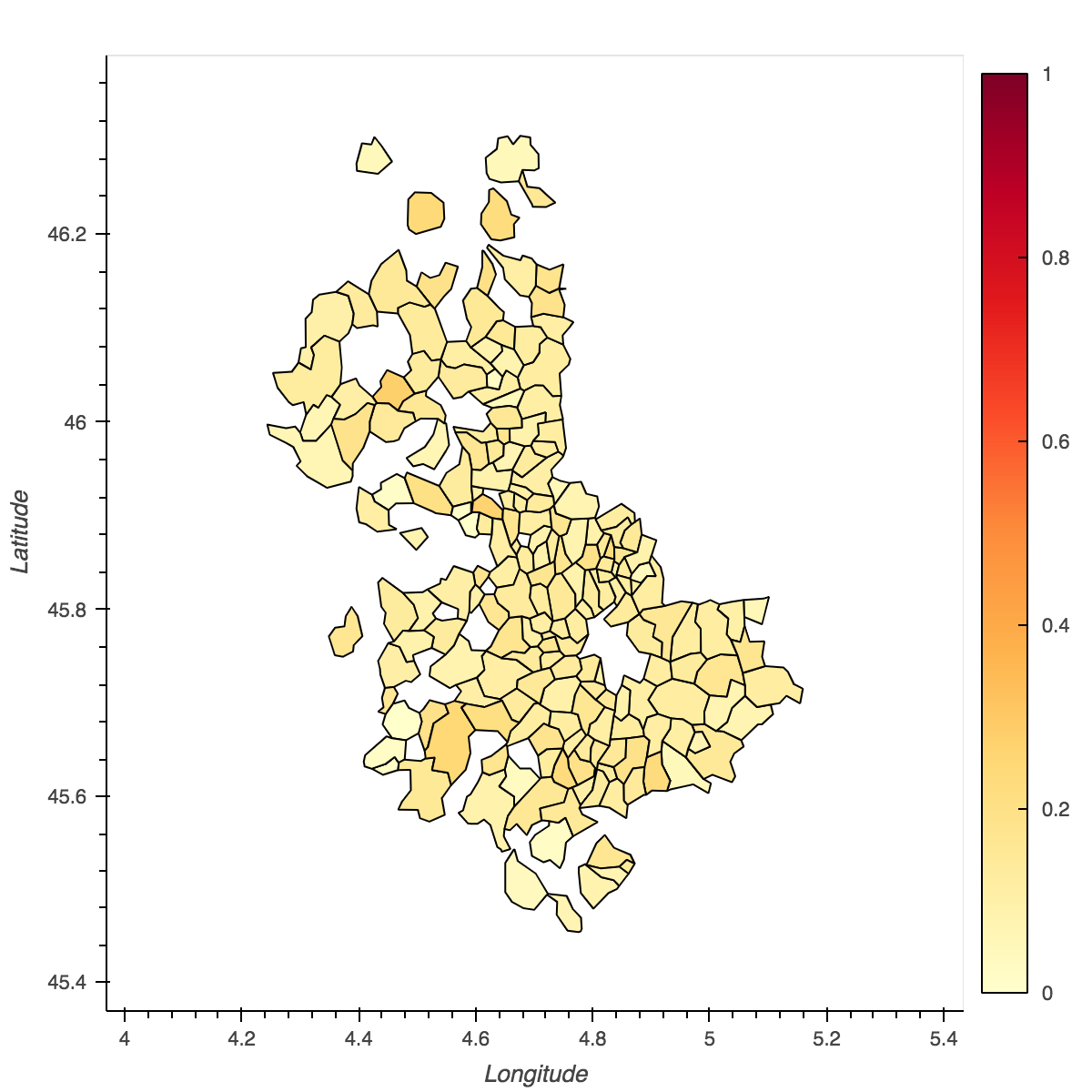}}
\subfloat[Department 86]{
\includegraphics[scale=0.10,valign=t]{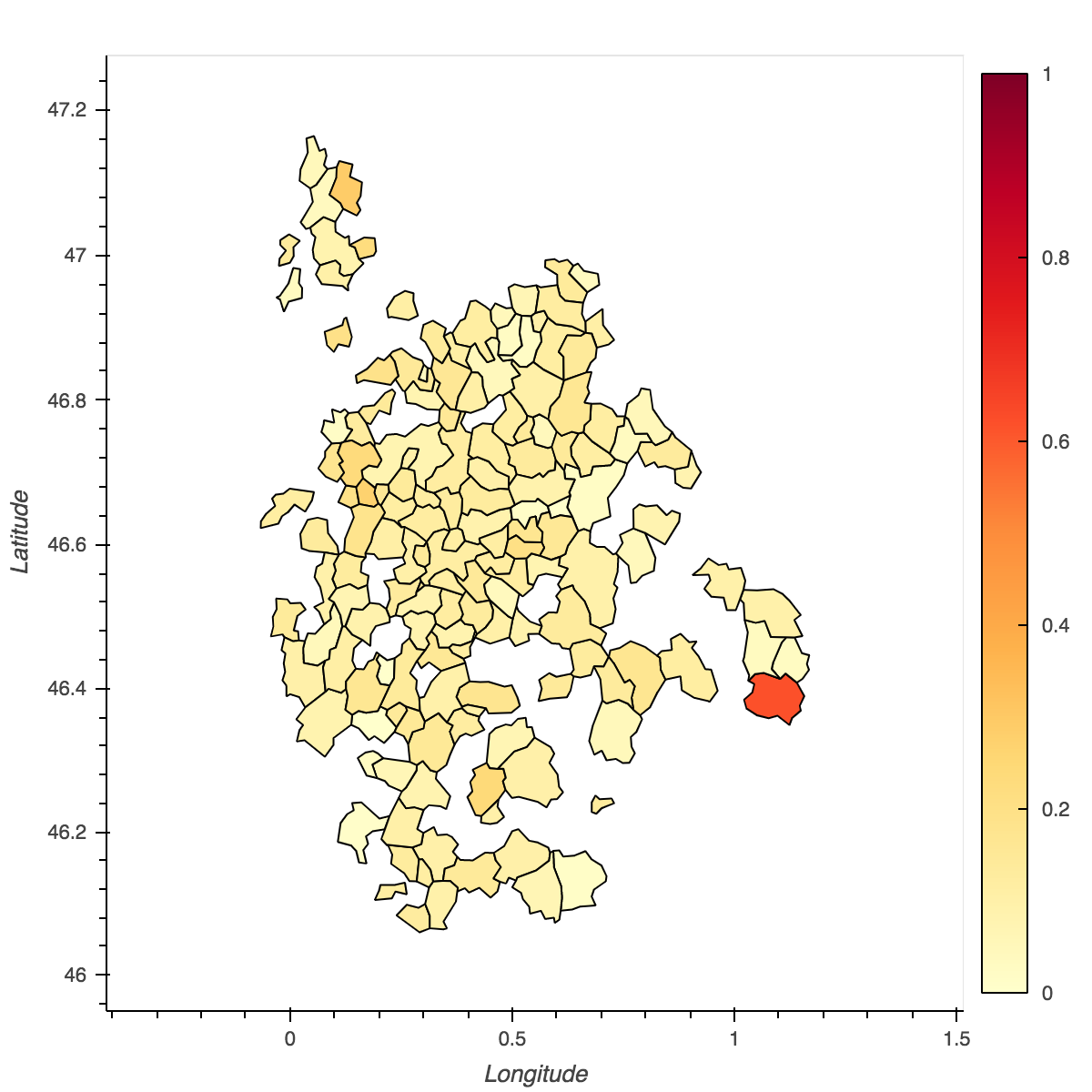}
\includegraphics[scale=0.10,valign=t]{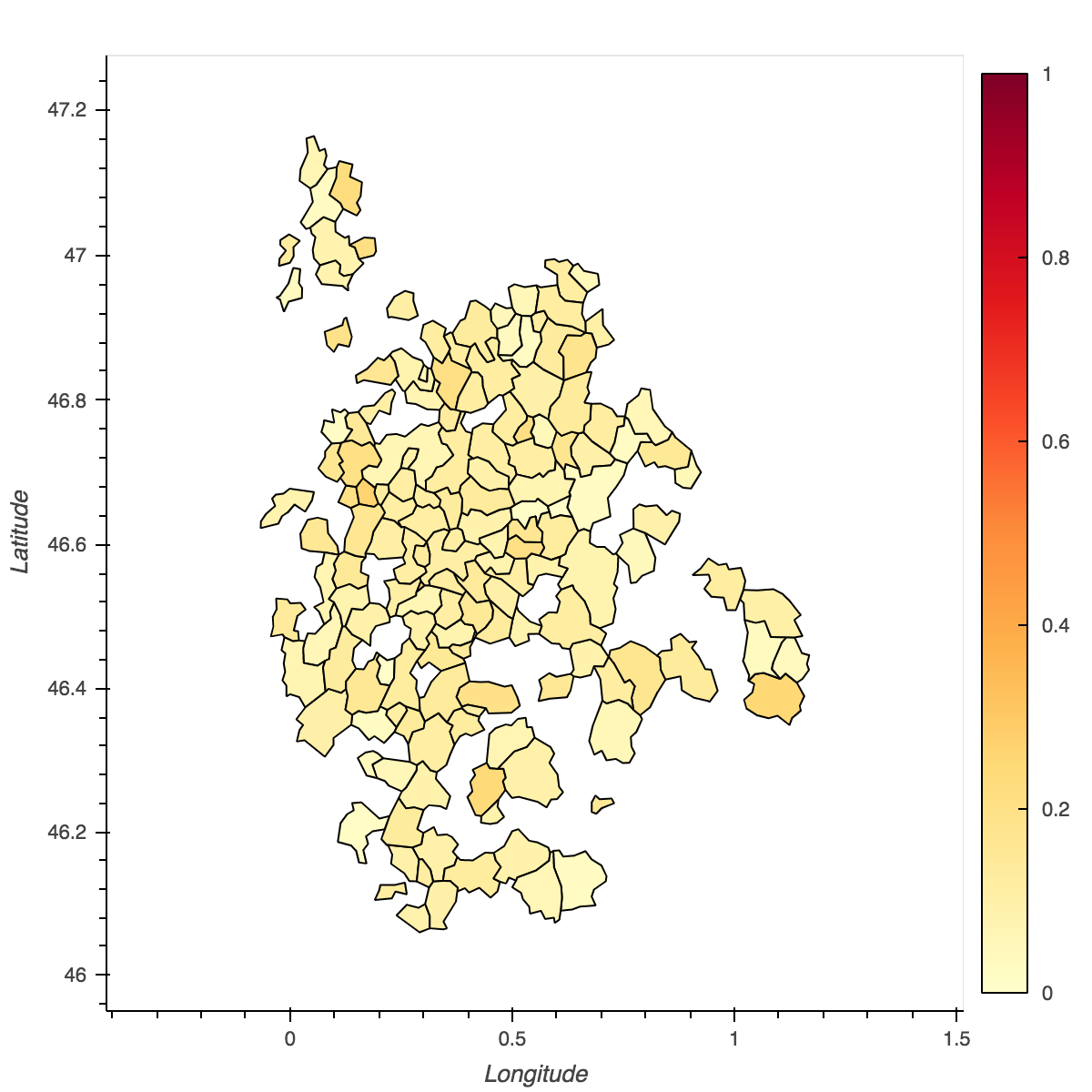}}
\\ 
\subfloat[Department 74]{   \includegraphics[scale=0.10,valign=t]{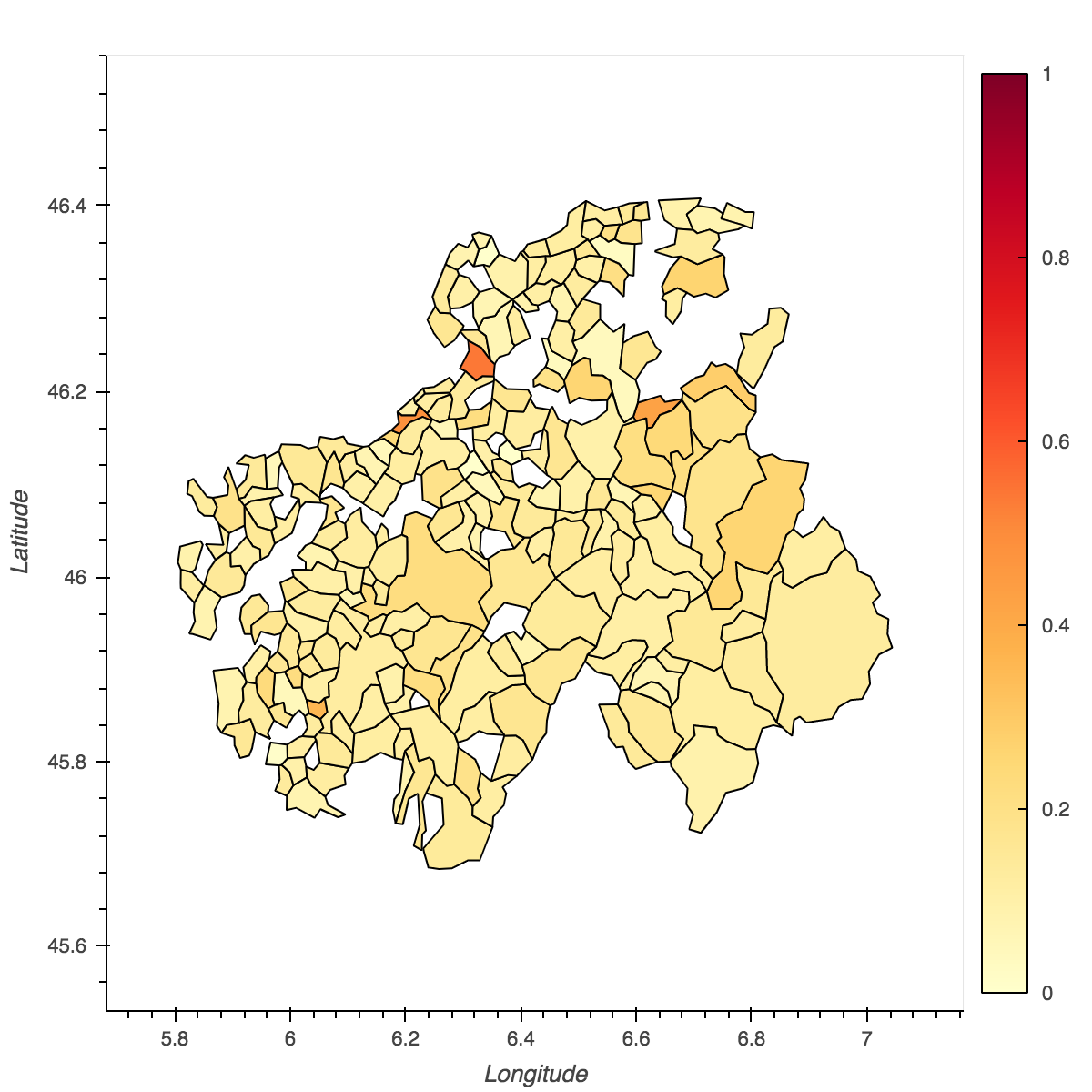}
\includegraphics[scale=0.10,valign=t]{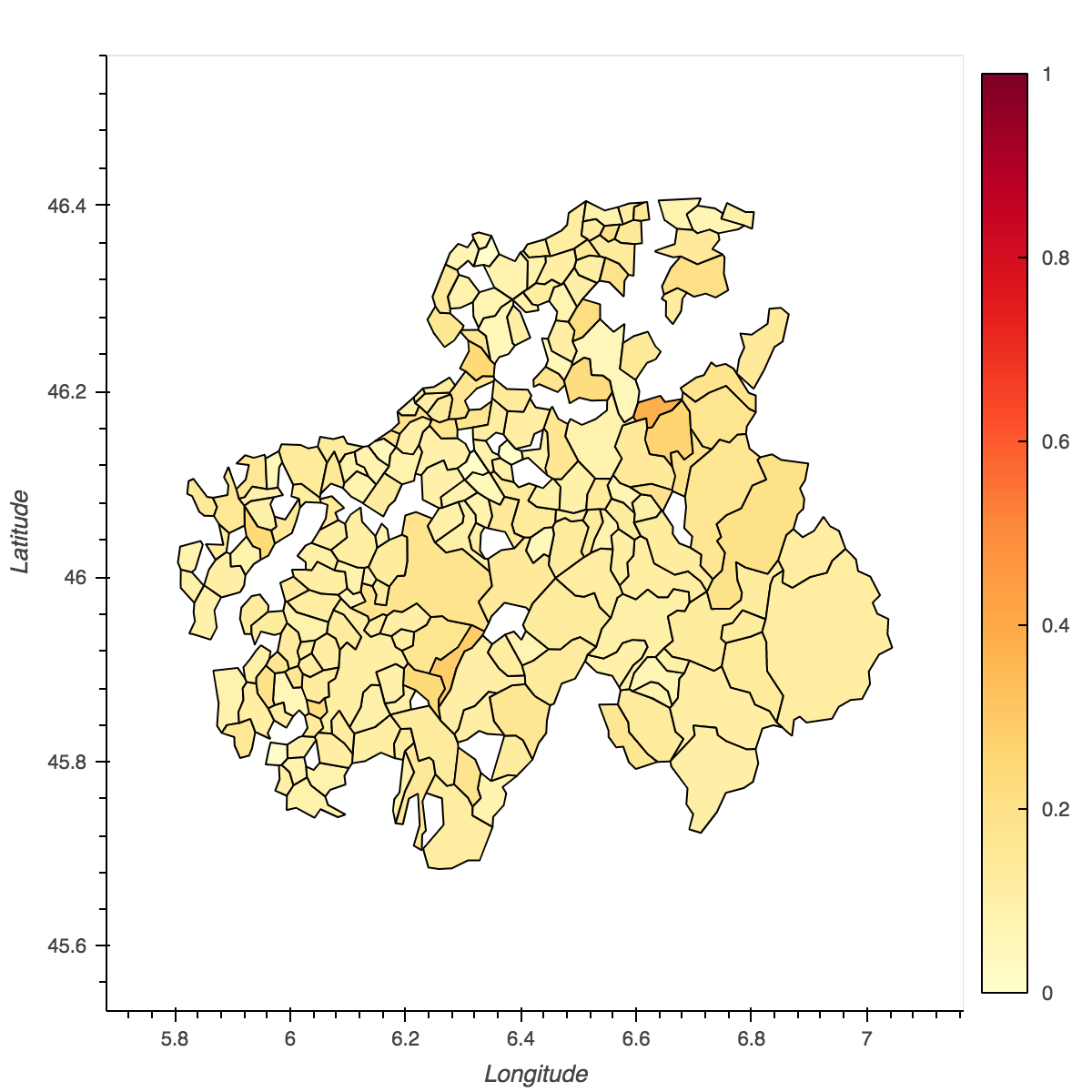}}
\subfloat[Department 81]{
\includegraphics[scale=0.10,valign=t]{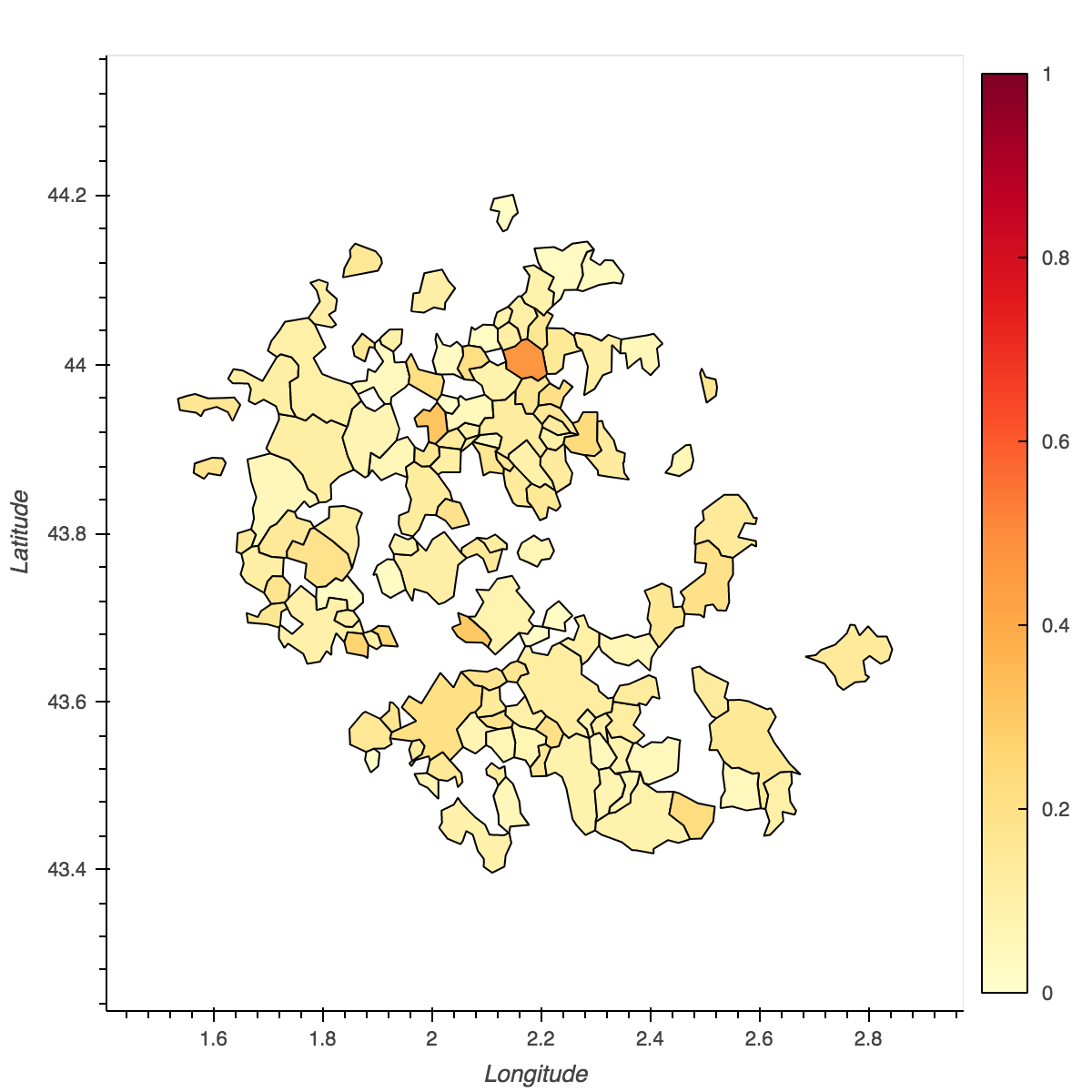}
\includegraphics[scale=0.10,valign=t]{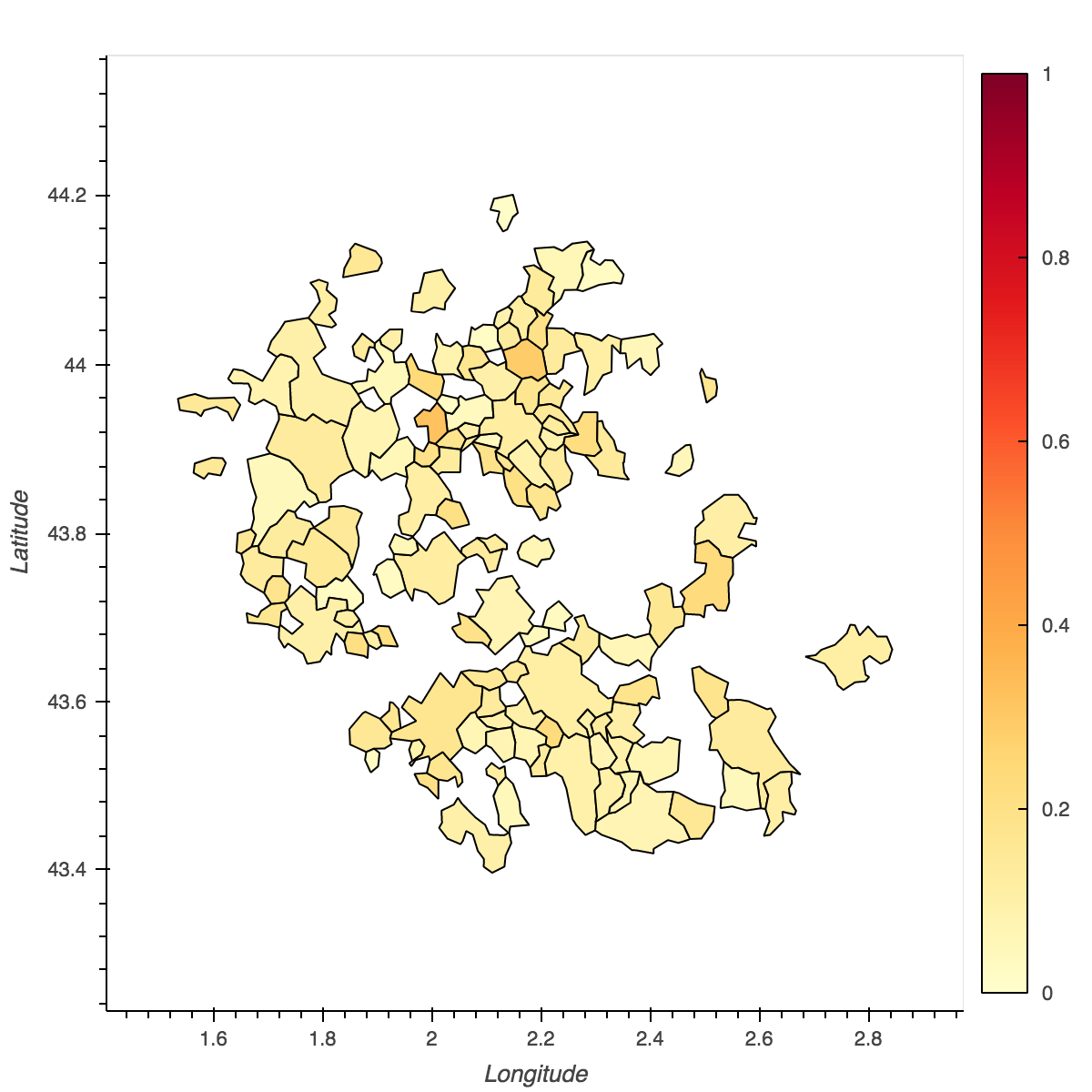}}
\\    
\caption{Mitigating Geographical Biases - Department Views. Left subfigure for each department represents a biased model ($\lambda$ = 0) and the right one represents an unbiased model ($\lambda = 0.6$)
\label{fig:S3_Dep}}
\end{figure}

In a context where geographic information can be a proxy of the race of the individuals, it is important to be able to remove any discrimination biases from the geographical position \cite{frees2021discriminating}. As in many cases, especially in Europe, sensitive information are not accessible, deriving it from the geographic position can be beneficial for many biases mitigation. Note that removing all variables that reflect the geographic information (i.e., place of residence, car parking position) is not enough to have independence with the output prediction of the models. 
In many cases, other correlated variables may be reflected in the data set (e.g., age, car brand). We perform this experiment on the pricingame 2017 dataset \cite{pricinggame17} on a binary task. The purpose of this exercise is to train a fair classifier that estimates
the TPL claim likelihood without incorporating any geographical bias in terms of Demographic Parity. 
For this scenario, we set the longitude and the latitude of the policyholder's residence as sensitive attributes and we remove the encoder $\gamma$ of the geographical component of our adversarial methodology (Figure~\ref{fig:DP_Auto_Encoder}). We run separate experiments for the demographic parity objective. More specifically, we apply our AE\_HGR algorithm with different $\lambda$ hyper-parameters. As a baseline, we use a classical, “unfair” GLM 
(i.e., $\lambda=0$). In order to be able to compare the results  we have represented in Figure~\ref{fig:S3_GEO} the average prediction by French department and in Figure~\ref{fig:S3_Dep} by French INSEE code. We observe that increasing $\lambda$ tend to smooth the predictions. By increasing the $\lambda$, the higher values observed in particular in \emph{Ile-de-France} tend to converge towards the average.


In Figure~\ref{fig:S3_Dep}, we observe the same pattern. Compared to the others, the French INSEE codes with significant average output prediction tend to be neutralized by increasing $\lambda$. For example, in Sub-Figure (a), department $44$, we observe that the INSEE code $29830$ with a claim likelihood
prediction of more than $0.8$ (represented in dark red in the left area) has decreased between a range of $0.4$ and $0.6$. This scenario is confirmed in other departments such as $86$, $69$, $81$ where the INSEE code obtain more similar output predictions in average in the fair model.
As expected increasing the $\lambda$ hyper-parameter seems to neutralize the spatial effect in the predictive model. 

\newpage
\section{Conclusion}
We developed a novel general framework designed specifically for insurance pricing. First, we extend autoencoders to generate the geographic and car rating components needed for pure premium prediction. To the best of our knowledge, this is the first such method to be applied in a single whole insurance pricing model, as it traditionally requires a two-step structure. Compared with the traditional pricing method, our method proves to be more efficient in terms of performance accuracy on various artificial and real data sets. We attribute this to the ability of the autoencoder to extract useful car and geographic risks, while the traditional pricing structure has significant limitations in correctly modeling the risks associated with these deep and complex components. 
Furthermore, this general framework has proven to be highly interesting for applying an adversarial learning approach specifically designed to improve the fairness of insurance pricing. We show empirically that the different components predicted by the model are debiased in contrast to traditional approaches that might remove 
the important information for predicting the pure premium. 
This approach shows to be more efficient in terms of accuracy for similar levels of fairness on various data sets.
As future work, it might be interesting to consider a generalization of our proposal for telematics insurance where some biases can be mitigated on different aggregate scores. 



\newpage
\bibliographystyle{apalike}
\bibliography{bib}

\end{document}